\documentclass{bmvc2k}

\usepackage{booktabs}
\usepackage[table]{xcolor}
\usepackage[capitalize]{cleveref}
\usepackage{svg}
\usepackage{amsfonts}
\usepackage{bbm}
\usepackage[nice]{nicefrac}

\usepackage{amsmath, amssymb}

\title{ReFineR: Single image to dense point cloud registration at pixel level with point splatting}

\addauthor{Ugo Insalaco}{ugo.insalaco@minesparis.psl.eu}{1,2}
\addauthor{Jean-Emmanuel Deschaud}{jean-emmanuel.deschaud@minesparis.psl.eu}{1}
\addauthor{Pierre Dellenbach}{pdellenbach@chapsvision.com}{2}

\addinstitution{
Mines Paris, PSL University\\
Centre for robotics (CAOR)\\
75006 Paris, France
}
\addinstitution{
ChapsVision\\
4 Rue du Port aux Vins, \\
92150 Suresnes, France
}
\runninghead{Insalaco, Deschaud, Dellenbach}{ReFineR}

\def\eg{\emph{e.g}\bmvaOneDot}

\def\etal{\emph{et al}\bmvaOneDot}

\newcommand{\smallpm}[2]{#1{\fontsize{4pt}{4.5pt}\selectfont$\pm$#2}}
\newcommand{\imgext}{.jpg}
\begin{document}

\maketitle

\begin{abstract}
Cameras and LiDARs (Light Detection and Ranging) provide complementary observations of the same scene: cameras capture rich appearance cues while LiDAR sensors provide accurate geometric measurements. In this work, we address the problem of fine image-to-point cloud registration from a single image and a roughly estimated camera pose. On robotic platforms, residual reprojection errors of several pixels are commonly observed even after calibration due to temporal desynchronization, mechanical vibrations, or inaccuracies in point cloud reconstruction. Such errors can lead to decimeter-level geometric discrepancies at moderate distances and significantly affect multiple downstream tasks, such as dense cross-modal supervision, texture mapping, and semantic label transfer. To address this challenge, we introduce ReFineR: an offline framework to finely register individual images to pre-reconstructed point clouds. ReFineR builds upon differentiable point splatting rasterization and a smooth Gradient Orientation Measure (GOM) to iteratively refine the camera pose through analysis-by-synthesis optimization. We evaluated the method on a curated dataset of manually calibrated images collected from heterogeneous data sources. Furthermore, we show that an automatic point cloud rescale factor can be estimated prior to optimization, enabling robust refinement across a wide range of scenarios using a single scale-invariant formulation. The project page can be found at \url{https://refiner-project.github.io/}
\end{abstract}

\section{Introduction}
\label{sec:intro}
\begin{figure}[tb]
    \centering
    \includesvg[width=\linewidth]{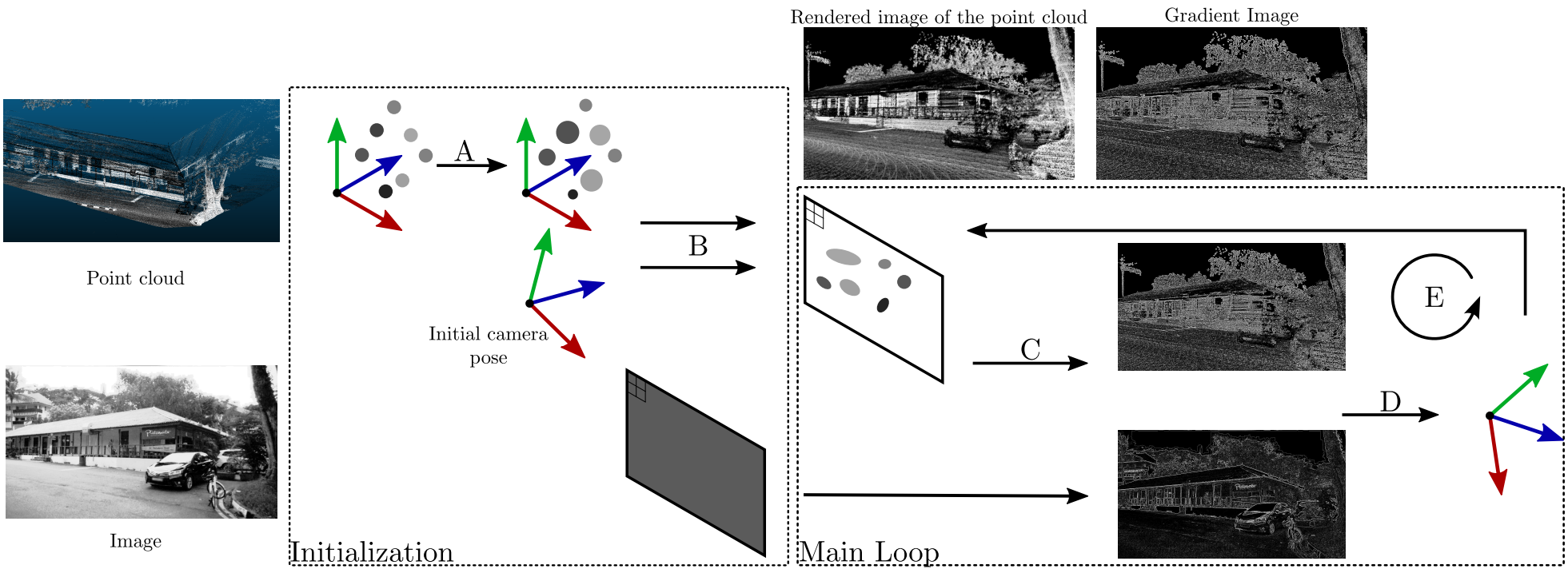}
    \caption{ReFineR optimization loop.
    A. During initialization, points are spatially extended to prepare the differentiable rendering step.
    B. Differentiable point splatting creates a view of the point cloud at the initial camera position (\cref{subsec:gs_init}). 
    C. Gradients are computed using a Sobel filter. 
    D. Both modalities are compared using the GOM loss and selected comparison channels (\cref{subsec:loss} and \cref{subsec:comp}). 
    E. The camera pose is optimized by backpropagation through the differentiable image, and the new pose iteratively renders new views (\cref{subsec:optim}).}
    \label{fig:pipeline}
\end{figure}

Accurate camera-LiDAR alignment is often treated as a solved pre-processing step, assuming a fixed extrinsic calibration between sensors across an entire sequence. However, in practice, even well-calibrated multi-sensor platforms exhibit residual reprojection errors that vary from frame to frame due to temporal misalignment, mechanical vibrations, rolling-shutter effects, and SLAM drift. In large-scale autonomous driving datasets such as nuScenes~\cite{Caesar2019nuScenesAM}, reprojection discrepancies of 7--10 pixels are commonly observed despite official calibration parameters (details in Supplementary~\cref{app:calib_results}). Although such errors may appear negligible, they can lead to significant metric misalignments: a 10-pixel error can induce up to 40 cm of lateral misalignment at moderate distances. This magnitude critically affects downstream applications like dense cross-modal supervision, semantic label transfer from image to LiDAR, HD map construction, and fine-grained scene understanding, where centimeter-level consistency is required. Importantly, these errors are not constant over time and cannot be corrected by a single global calibration step, making per-image fine alignment essential. Therefore, beyond coarse calibration, sub-pixel single-camera-LiDAR registration is a fundamental problem for high-precision multi-sensor perception. 

Advances in performant SLAM-LiDAR algorithms~\cite{dellenbachCTICPRealtimeElastic2021} allow the reconstruction of dense and precise point clouds from the aggregation of multiple LiDAR scans. When used for rasterizing new images, these dense point clouds create high-quality images containing rich visual information that can be leveraged for image-to-point cloud registration. Previous work~\cite{guislainFineScaleImage2017, koideGeneralSingleshotTargetless2023, Taylor2015MultiModalSC, chengLoGSVisualLocalization2024, Chen2020CategoryLO, Lin2020iNeRFIN} has shown that analysis-by-synthesis -- the principle of iteratively rendering the point cloud in artificial images and comparing each render with the real image in a loop -- performs significantly well for the task of fine-level registration. While earlier methods used non-differentiable algorithms for optimization~\cite{Taylor2015MultiModalSC, guislainFineScaleImage2017}, our work takes advantage of the recent development of differentiable rendering techniques~\cite{YifanDSS2019, kerbl3DGaussianSplatting2023, Lassner2021PulsarES, wilesSynSinEndtoEndView2019} to propose a fully differentiable pipeline, from the camera pose parameters to the loss value.

This paper presents a threefold contribution. \textbf{1)} As pixel-level registration is in fact absent from many LiDAR-image datasets~\cite{Caesar2019nuScenesAM, Sun2019ScalabilityIP, CarlevarisBianco2016UniversityOM}, we propose an open dataset, created from six heterogeneous sources of point cloud and image data (ETH3D~\cite{schoeps2017cvpr}, CARLA~\cite{Dosovitskiy2017CARLAAO}, nuScenes~\cite{Caesar2019nuScenesAM}, Tanks and Temples~\cite{knapitschTanksTemples2017}, RGBDScenes~\cite{Lai2014UnsupervisedFL}, and data from a Livox sensor~\cite{yuanPixelLevelExtrinsicSelf2021}) with manually calibrated images on reconstructed dense point clouds for three of these sets. \textbf{2)} We propose an automatic rescaling method, derived from first-order pinhole camera model analysis, to provide a principled solution to current scene-dependent hyperparameter tuning. \textbf{3)} We introduce ReFineR for \textbf{Re}gistration at \textbf{Fine} level using differentiable \textbf{R}endering, between single images and any pre-reconstructed point clouds. ReFineR is a training-free method that handles heterogeneous point channels at once. Prioritizing registration accuracy over speed, ReFineR was created for offline processing (HD mapping, semantic transfer...) rather than being a localization brick in real-time systems. 

\section{Related work}
Knowing correspondences between 3D points and 2D pixels, closed-form solutions exist to estimate the camera's 6 degrees of freedom~\cite{lepetitEPnPAccurateSolution2008}. For more robustness to matching noise, most common approaches use either the EPnP algorithm~\cite{lepetitEPnPAccurateSolution2008} combined with the RANSAC~\cite{fischlerRandomSampleConsensus1981} scheme, or a residual minimization of the reprojection error using the Levenberg-Marquardt optimizer~\cite{levenbergMETHODSOLUTIONCERTAIN1944,marquardtAlgorithmLeastSquaresEstimation1963,koideGeneralSingleshotTargetless2023}. For the relaxed problem of camera alignment without known matches, a recent survey from An \etal~\cite{An2024SurveyOE} classifies automatic registration methods as explicit or implicit from their estimation of point-pixel pairs or not.

\subsection{Explicit registration}
In order to remove the manual task of selecting matches, some work has been focusing on automatically estimating such correspondences, either via feature matching algorithms in 2D, such as SuperGlue~\cite{sarlinSuperGlueLearningFeature2019, koideGeneralSingleshotTargetless2023} or MatchAnything~\cite{He2025MatchAnythingUC}, or with direct estimation of 2D-3D matches~\cite{renCorri2pDeepImagetopoint2022,kangCoFiI2PCoarsetoFineCorrespondences2023,li2D3DMATR2D3DMatching2023,wangFreeRegImagetoPointCloud2023, Wang2023EndtoEnd2R, Mu2025Diff2I2PDI}. CoFiI2P~\cite{kangCoFiI2PCoarsetoFineCorrespondences2023} trains a transformer model to generate matches between image and point features at different scales. CorrI2P~\cite{renCorri2pDeepImagetopoint2022} adds global descriptors on both modalities, given to an overlapping region detector network that produces frustum-consistent matches. I2PNet~\cite{Wang2023EndtoEnd2R} estimates soft matches between points from the LiDAR scan and pixels, themselves fed to a pose regression network. Diff2I2P~\cite{Mu2025Diff2I2PDI} uses a deformable correspondence tuning layer to estimate matching offsets and a diffusion loss to compare RGB images and depth maps. UniCorrn~\cite{goswami2026unicorrn} embeds 2D and 3D features in a common space to predict correspondences between any modality pair, regrouping 2D-2D, 2D-3D, and 3D-3D matching into a single algorithm. A key limitation of most 2D-3D matching algorithms \cite{renCorri2pDeepImagetopoint2022,kangCoFiI2PCoarsetoFineCorrespondences2023,li2D3DMATR2D3DMatching2023,wangFreeRegImagetoPointCloud2023,Wang2023EndtoEnd2R,Mu2025Diff2I2PDI, goswami2026unicorrn} is their generalization ability to various point clouds. Since the models were generally trained on individual frames (LiDAR or RGBD), they tend to show weaker results given denser point clouds as input (for instance, acquired from a dense static LiDAR scanner or aggregated scans from a mobile LiDAR). Other approaches try to directly predict pointmaps (a Width $\times$ Height $\times$ 3 tensor representing the position in space of each pixel): 
Brachmann \etal~\cite{brachmannAcceleratedCoordinateEncoding2023,brachmannVisualCameraReLocalization2021,yangSANetSceneAgnostic2019}  trained a deep neural network with a differentiable version of RANSAC~\cite {brachmannVisualCameraReLocalization2021} for this task. SANet~\cite{yangSANetSceneAgnostic2019} estimated a pointmap from deep matching between a query image and the point cloud and reference views' features. Similarly, both these methods are typically trained on specific datasets or scenes, which limits their ability to generalize across different environments or scenes. In contrast, our approach does not require any training and can be applied directly to a wide variety of image–LiDAR datasets without parameter tuning.

\subsection{Implicit registration}
A parallel line of work focuses on synthesizing views from the point cloud using coarse initialization, then comparing the rendered and real images to refine the position and orientation, with the goal of overlapping both images (\textit{a.k.a.} analysis-by-synthesis). Iterations can either be carried out using neural network comparison of images~\cite{lvLCCNetLiDARCamera2021,iyerCalibNetSelfSupervisedExtrinsic2018} -- CalibNet~\cite{iyerCalibNetSelfSupervisedExtrinsic2018} concatenates image and depth map features before feeding them to a fully connected neural network, LCCNet~\cite{lvLCCNetLiDARCamera2021} introduced a cost volume layer instead, and Ou~\etal~\cite{Ou2024IterativeCE} added a trained diffusion model as a denoiser to the pose predicted by LCCNet -- or optimization of the pose's parameters from handcrafted loss functions between images~\cite{koideGeneralSingleshotTargetless2023,Taylor2015MultiModalSC,Castorena2016AutocalibrationOL,yuanPixelLevelExtrinsicSelf2021,Liu2021TargetlessEC,guislainFineScaleImage2017,Pandey2012AutomaticTE}. Early work from Taylor~\etal~\cite{Taylor2015MultiModalSC} compared a rasterized view of the point cloud to the image using Gradient Orientation Measure which minimizes the angle between each pixel's gradient. This loss was then optimized using the non-differentiable Nelder-Mead optimizer~\cite{nelderSimplexMethodFunction1965}. Koide \etal's method~\cite{koideGeneralSingleshotTargetless2023} minimized the normalized information distance, a comparison between the distributions of RGB values and synthesized LiDAR reflectivity to optimize the pose. Guislain \etal~\cite{guislainFineScaleImage2017} refined the camera position using a mix of entropic and gradient information thanks to Mutual Information and Histogram of Oriented Gradients (HOG) features~\cite{dalalHistogramsOrientedGradients2005}. Yuan \etal~\cite{yuanPixelLevelExtrinsicSelf2021} first extracted depth continuous edges from the point cloud and image, and minimized a line reprojection error over the Lie algebra of rigid transforms. Our approach (ReFineR) falls into the category of implicit registration.

\subsection{Differentiable rendering}
Differentiable rendering creates images whose pixel values are differentiable with respect to the 3D scene's parameters (point positions, camera extrinsics and intrinsics, etc). These methods arose jointly with smooth 3D representation and reconstruction engines like NeRF~\cite{mildenhallNeRFRepresentingScenes2020} or Gaussian Splatting~\cite{kerbl3DGaussianSplatting2023}, and involve mechanisms like volumetric rendering~\cite{mildenhallNeRFRepresentingScenes2020, Blanc2024RayGaussVG} or Gaussian/surfel splatting~\cite{kerbl3DGaussianSplatting2023,wilesSynSinEndtoEndView2019, Lassner2021PulsarES, YifanDSS2019}. For instance, DSS~\cite{YifanDSS2019} projects isotropic flat Gaussians centered on each point onto the image plane, and averages for each pixel the contribution of each Gaussian. To account for the discontinuous point occlusion, they computed artificial gradients based on the relative movement between the point and the deprojected pixel. 3D Gaussian Splatting \cite{kerbl3DGaussianSplatting2023} took a completely differentiable path by rasterizing Gaussians onto the image plane and blending them using an opacity value. 

\subsection{Gaussian Splatting registration}
Setting aside image-to-point-cloud registration, some studies have been carried out on registering new images on pre-reconstructed Gaussian Splatting models~\cite{chengLoGSVisualLocalization2024, Liu2024GSLocEC, Xu2024DGSLAMRD, Bortolon20246DGS6P, Matsuki2023GaussianSS} using either feature matching on rendered views~\cite{chengLoGSVisualLocalization2024, Liu2024GSLocEC}, reprojection error minimization from rendered depth maps~\cite{Xu2024DGSLAMRD}, or reverse rendering to find rays intersecting the image plane~\cite{Bortolon20246DGS6P}. Matsuki \etal~\cite{Matsuki2023GaussianSS} proposed to directly integrate Lie gradients over the camera pose in the rendering process to achieve end-to-end differentiability from camera to pixel values. These methods do not directly apply to our problem as they require preliminary optimization of a Gaussian Splatting model from multiple images of the same scene (which in our single-image case is not available) and are often based on photometric error minimization~\cite{Matsuki2023GaussianSS, chengLoGSVisualLocalization2024}, which cannot be achieved with geometry- or reflectivity-valued LiDAR point clouds. They, however, prove that differentiable rendering can be a valuable asset for estimating fine camera positions.
\section{Preliminaries} 
\noindent
\par 

\textbf{Reprojection error and parallax: }
Pixel error $p_e$ (a.k.a. reprojection error) can be defined as the root-mean-square value of the displacement of pixels between two projection models $\pi_1$ and $\pi_2$, given a point cloud $(X_i)_{1\leq i\leq N}\in \left(\mathbb{R}^{3}\right)^N$:

\begin{equation}
    p_e = \sqrt{\frac{1}{N}\sum_{i=1}^N \left\| \pi_1(X_i) - \pi_2(X_i)\right\|^2}
\end{equation}

In our problem, pixel error is interesting because it defines the registration difficulty independently of the latent point cloud scale (which would not be the case when comparing distances using the translation error). For a small perturbation $\delta\tau\in\mathfrak{se}(3)$ of the camera extrinsics around $0$, we obtain:
\begin{align}
    \pi(X_i) - \pi_0(X_i) &= J(X_i) \delta\tau + o(\lVert\delta \tau\rVert) \label{eq:dl}
\end{align}
with ${J=\frac{\mathcal{D} \pi}{\mathcal{D} \tau}}$ (borrowing notation from~\cite{Sol2018AML}) computed explicitly in supplementary. In particular, the terms in $J$ associated with rotation only depend on point projections $(x_c, y_c) = (f_x X_{iX} / X_{iZ}, f_y X_{iY} / X_{iZ})$ on the image, with $f_x$, $f_y$ the camera intrinsics. In contrast, the ones for translation also depend on their depth $X_{iZ}$: the farther a point is, the less impact a translation will have on its reprojection error. This is the well-known phenomenon of \textit{parallax}. 

\section{Method}
We assume that a coarse initialization of the camera pose is available prior to optimization. Such an initialization may originate from a standard camera-LiDAR calibration performed on the robotic platform, combined with LiDAR-Odometry~\cite{Caesar2019nuScenesAM} or from Structure-from-Motion point clouds registered on a LiDAR point cloud~\cite{knapitschTanksTemples2017}. Importantly, estimating this coarse pose differs fundamentally from the fine alignment addressed in this work, as it is typically solved through geometric calibration or feature matching rather than analysis-by-synthesis optimization.

The ReFineR algorithm is structured into four main stages (see \cref{fig:pipeline}). \textbf{1)} The input point cloud scale is normalized following \cref{subsec:rescale}, allowing our method to be scale independent. \textbf{2)} A rendered view of the point cloud is created with differentiable point splatting (based on Gaussian Splatting's~\cite{kerbl3DGaussianSplatting2023} framework), as described in~\cref{subsec:gs_init}. \textbf{3)} Handcrafted features, based on gradient orientation, are computed on the synthetic and real images and compared according to the rendered channels (\cref{subsec:loss} and \cref{subsec:comp}). \textbf{4)} Backpropagation is carried out with computation of Lie gradients with respect to the camera parameter $\tau$ (\cref{subsec:optim}), and the process continues from step 2. This pipeline is entirely differentiable.

\subsection{Point cloud rescale estimation}
\label{subsec:rescale}
When handling heterogeneous point clouds (\eg., indoor, outdoor, or RGB-D data) and accounting for parallax effects, optimizing the camera pose using gradient-based algorithms requires prior knowledge of the point cloud’s scale in order to appropriately adjust the learning rate. In this section, we propose an expression for automatically estimating such a prior, allowing fixed learning rates for rotation and translation across all scenes.

Following \cref{eq:dl}, we can approximate the pixel error using:
\begin{equation}
    p_e^2 = \delta\tau^T\underbrace{\left(\frac{1}{N}\sum\limits_{i=1}^NJ(X_i)^TJ(X_i)\right)}_S\delta\tau + o(\lVert\delta \tau\rVert ^2) \label{eq:pixel}
\end{equation}
Intuitively, the symmetric positive matrix $S$ integrates the contributions and correlations to the pixel error of each transform parameter in $\tau$. Separating contributions from rotation and translation, and introducing the unknown scale coefficient $\alpha$ such that ${X_i = \alpha X_i'}$, this matrix can be rewritten as (details in Supplementary \cref{app:scale}):
\begin{equation}
    S = \begin{pmatrix}
        S_r & S_{rt} \\
        S_{rt}^T & S_t
    \end{pmatrix} = \begin{pmatrix}
        S_r & \frac{1}{\alpha}S_{rt}' \\
        \frac{1}{\alpha}S_{rt}'^T & \frac{1}{\alpha^2}S_t'
    \end{pmatrix}
\end{equation}
To select an appropriate $\alpha$, our criterion is to balance the effects of translation and rotation on the pixel error. Concretely, we would expect a translation of 1 (written without a unit on purpose), to have approximately the same effect on the pixel error as a rotation of 1. To do so, we choose to equalize the Hilbert-Schmidt norms between $S_r$ and $S_t'$. We obtain:
\begin{equation}
    \alpha = \left(\nicefrac{\sum\limits_{i = 1}^3\lambda_{r, i}^2}{\sum\limits_{i = 1}^3\lambda_{t, i}^2}\right)^{1/4}
\end{equation}
with ${(\lambda_{r,i}, \lambda_{t, i})_i}$ the respective eigenvalues of $S_r$ and $S_t$. An example of rescaled scenes is shown in Supplementary \cref{fig:rescale}. Generally speaking, this rescaling procedure is a form of reparameterization to obtain better variable conditioning, which is a major issue in numerical optimization~\cite{Triggs1999BundleA}.

\subsection{Synthesis}
\label{subsec:gs_init}
After rescaling the point cloud, we initialize a Splatting model by creating an isotropic Gaussian for each 3D point (with standard deviation proportional to the distance to each point's nearest neighbor by a factor $p_s$), and setting a fixed opacity $o$. To limit the visual impact of isolated points (which would therefore be initialized with a large size), we limit the size of each Gaussian such that its projection does not exceed 4 pixels in diameter. We add each point channel value as a Gaussian feature (normals calculated from the point cloud, LiDAR reflectivity, etc.). 

\subsection{Analysis}
\label{subsec:loss}
Edge-based cues have been explored in prior work for cross-modal alignment~\cite{guislainFineScaleImage2017, Taylor2015MultiModalSC}. Building on the observation that image gradients correlate with both point cloud geometry and object texture, we follow Taylor \etal's work~\cite{Taylor2015MultiModalSC} and adapt the Gradient Orientation Measure (GOM). From two images $\left(I_p\right)_p$, $\left(I'_p\right)_p$ indexed by their pixels and their corresponding gradients $\left(G_p\right)_p$ and  $\left(G'_p\right)_p$, computed using (differentiable) Sobel filters, the GOM is defined as
\begin{equation}
    GOM(I, I') = \frac{\sum_p|\left(G_p | G'_p\right)|}{\sum_p\lVert G_p\rVert_2 \lVert G'_p\rVert_2}
\end{equation}
GOM aims at maximizing the dot product between gradients, therefore aligning edges from two images. 

\subsection{Comparison}
\label{subsec:comp}
Comparison between modalities is achieved by creating correspondences between point clouds and image channels, and computing the GOM loss function for each correspondence. For the point cloud, the different usable channels are: average normal direction ($p^n$), individual normal coordinates ($p^{\mathit{nx}}$, $p^{\mathit{ny}}$, $p^{\mathit{nz}}$), normalized projected depth ($p^d$), LiDAR reflectivity ($p^r$), or grayscale value per point ($p^g$) acquired from an embedded camera in the LiDAR sensor (for example the fixed terrestrial FARO scanner). For images, we compute their grayscale values ($i^g$) from the RGB channels. In our ablation study, we experiment with adding generated channels for images using the Marigold diffusion model~\cite{Ke2025MarigoldAA}, adding the artificial normal channels $i^\mathit{nx}, i^\mathit{ny}, i^\mathit{nz}$ and normalized depth $i^\mathit{d}$ (see \cref{app:ablation_channel}). 

The final multimodal loss function between the multichannel rendered image ${(I^c)_{c\in{\mathcal{C}_{\text{3D}}}}}$  and real image ${(I^{\prime c})_{{c\in \mathcal{C}_{\text{2D}}}}}$ is then computed as: 
\begin{equation}
    \mathcal{L}(I, I') = -\frac{1}{|\mathcal{C}_{\leftrightarrow}|}\sum\limits_{(c,c') \in\mathcal{C}_{\leftrightarrow}}GOM\left(I^c, I^{\prime c'}\right)
\end{equation}

With ${\mathcal{C}_{\text{3D}}}$ and ${\mathcal{C}_{\text{2D}} }$ denoting the sets of available point cloud and image channels and $\mathcal{C}_{\leftrightarrow} \subset \mathcal{C}_{3D}\times \mathcal{C}_{2D}$ a set of correspondences between 3D and 2D channels. In our experiments, we consider two common setups: the first, where the only available information is the point cloud geometry, and the second, with LiDAR reflectivity or grayscale value. For the first scenario, we therefore use ReFineR with ${\mathcal{C}_\leftrightarrow = \{(p^{nx}, i^g), (p^{ny}, i^g), (p^{nz}, i^g)\}}$ and for the second, we use ${\mathcal{C}_\leftrightarrow = \{(p^r, i^g)\}}$ or ${\mathcal{C}_\leftrightarrow = \{(p^g, i^g)\}}$.

\subsection{Optimization}
\label{subsec:optim}
Our pipeline is entirely differentiable from the camera parameter $\tau \in\mathfrak{se}(3)$ to the loss value $\mathcal{L}(I, I')$ and uses Gaussian Splatting SLAM's~\cite {Matsuki2023GaussianSS} codebase as the differentiable point splatting rasterizer. Optimization is carried out using the Lie version of gradient descent, with a fixed learning rate. Other optimization algorithms could be possible, as previous work experimented with integrating the Levenberg-Marquardt optimizer~\cite{Hllein20243DGSLMFG, levenbergMETHODSOLUTIONCERTAIN1944} through the Gaussian Splatting rasterizer, but their method additionally requires a large amount of GPU memory ($\sim$50GB), which would reduce the applicability of our method. The number of iterations is bounded between 50 and 1000, and a minimum step magnitude $s_m$ is fixed to dynamically stop optimization upon convergence.

\section{Experiments}

\subsection{Setup}
\noindent
\par \textbf{Data: } For the task of image-to-point cloud registration with a pixel-level goal, there is, to our knowledge, no unified evaluation framework. Generally, state-of-the-art methods either choose to run their experiments on their own data~\cite{koideGeneralSingleshotTargetless2023, guislainFineScaleImage2017, yuanPixelLevelExtrinsicSelf2021}, only on RGB-D images~\cite{li2D3DMATR2D3DMatching2023, wangFreeRegImagetoPointCloud2023} or on odometry-based calibration like KITTI Odometry or nuScenes~\cite{geigerVisionMeetsRobotics2013, renCorri2pDeepImagetopoint2022, kangCoFiI2PCoarsetoFineCorrespondences2023, lvLCCNetLiDARCamera2021} which do not have pixel-precision calibration (see \cref{tab:reproj} from Supplementary \cref{app:calib_results}) with registration on individual frames rather than the fully reconstructed point clouds. As part of our open contributions, we curated a dataset of point clouds and finely registered color images from six different sources: ETH3D~\cite{schoeps2017cvpr}, Tanks and Temples~\cite{knapitschTanksTemples2017}, nuScenes~\cite{Caesar2019nuScenesAM}, Livox~\cite{yuanPixelLevelExtrinsicSelf2021}, RGBDScenes~\cite{Lai2014UnsupervisedFL}, and synthetic data from the CARLA simulator~\cite{Dosovitskiy2017CARLAAO}. ETH3D is a multiview stereo benchmark with high-resolution images and high-quality point clouds acquired with a FARO scanner, used for evaluation. The ETH3D dataset provides subpixel camera registration, thanks to photometric error minimization across multiple views at once. This ideal scenario (RGB point cloud, high-precision scanner, multiple views) does not, however, reflect the full diversity of point clouds, and we therefore choose to enrich it from other sources. Tanks and Temples is closely related to ETH3D, as each point cloud was acquired using a similar FARO scanner, but we show (see Supplementary \cref{app:dataset}) that this dataset has coarser image registration of the order of 5 to 10 pixels on average. Indeed, the ground truth camera poses were obtained through manual registration of an SfM model to the point cloud, and errors originating from the sparse reconstruction directly impact the final camera poses. To improve on this error, we estimated the camera parameters for 20 images of this dataset by selecting points and pixel matches and minimizing the reprojection error with Gauss-Newton optimization. We obtained reprojection errors of $2$px on average after this refinement. More information on our manual registration procedure is given in \cref{app:dataset}. nuScenes data are outdoor sequences obtained from a Velodyne HDL32E LiDAR mounted on a car, alongside cameras, radars, and an IMU. We obtained dense point clouds by running CT-ICP~\cite{dellenbachCTICPRealtimeElastic2021}, a LiDAR odometry method that provides a precise pose for each LiDAR scan. After reconstruction and using the official sensor platform calibration, we still noticed residual errors and chose to manually register 20 images from this dataset, improving the registration from 7--10px to 3--5px. Livox data presented in Yuan \etal's paper~\cite{yuanPixelLevelExtrinsicSelf2021} consist of four scans obtained from a Livox sensor and a single image per scan. Each image was similarly manually registered. RGBDScenes consists of color and depth images recorded from a Kinect sensor, where point clouds are created by aligning frames using Patch Volumes Mapping~\cite{Henry2013PatchVS}. CARLA is an autonomous driving simulator that benefits from perfect registration, while still producing realistic point clouds and images. \textit{For the rest of this paper, except for the RGBDScenes dataset, we will consider images with a resized height of $1024$px. For RGBDScenes, height is kept at $480$px. Every pixel error will be expressed accordingly.} In total, this dataset is composed of 50 images from ETH3D, 50 images from CARLA, 20 images from Tanks and Temples, 20 images from nuScenes, 20 images from RGBDScenes, and 4 images from Livox, for a total of 164 finely registered images across 29 heterogeneous point cloud scenes.\\

\textbf{Methods: } ReFineR is evaluated against Koide \etal's method~\cite{koideGeneralSingleshotTargetless2023} and the intensity-free method from Yuan \etal~\cite{yuanPixelLevelExtrinsicSelf2021}. It is known that synthesized views of a point cloud can be used to find meaningful correspondences with real images in 2D \cite{schoeps2017cvpr, Huang2015SingleviewRV}. We thus compare ReFineR with the MatchAnything model~\cite{He2025MatchAnythingUC}, which extracts 2D matches between heterogeneous modality images. After matching in 2D using MatchAnything, we retrieve the 3D coordinates of non-empty pixels and estimate the pose using PoseLib's RANSAC. MatchAnything's default model configuration was kept. We added GS-SLAM, which uses photometric loss for comparison with the base work from Matsuki \etal~\cite{Matsuki2023GaussianSS}. The first two methods (Koide~\etal~\cite{koideGeneralSingleshotTargetless2023} and Yuan~\etal~\cite{yuanPixelLevelExtrinsicSelf2021}) require hyperparameter tuning based on the scene type and scale, which we detail in Supplementary \cref{app:benchmark}. Conversely, ReFineR is a scale-invariant approach, for which we fix ${s_m = 2 \times 10^{-5}}$, generally stopping our registration loop after 100 to 200 iterations. Gradient descent is handled by the Adam optimizer in PyTorch with learning rate ${l_r = 2\times10^{-4}}$. The splatting model is initialized using $o=0.2$ and $p_s=0.5$. All our experiments ran on a GeForce RTX 5090 graphics card with 32GB VRAM, but for most sets, 8GB VRAM would suffice. Indeed, ReFineR's computational cost is directly related to the number of points to be registered to, which in general doesn't exceed 10M (including frustum culling). We experimented with pretrained deep learning-based approaches~\cite{iyerCalibNetSelfSupervisedExtrinsic2018, Yuan2020RGGNetTA, goswami2026unicorrn, wangFreeRegImagetoPointCloud2023}, but noticed that they tend to predict abnormal 2D-3D correspondences for our task of fine-level camera calibration to dense point clouds. These methods were pretrained on individual nuScenes scans or RGBD depth maps, which we suspect introduced a distribution gap with inference on dense depth maps. A more detailed reflection is shared in \cref{app:unicorrn}. \\ 

\textbf{Randomized Pose Initialization: }Since point clouds are acquired at different scales and to keep consistency with our scale-invariant approach, we perturb our camera poses such that the resulting \textit{average} pixel error has a fixed value. This can be done using a first-order approximation of the pixel error (see \cref{eq:dl}) and selecting a random transform on the ellipsoid of constant error. Computation details are given in Supplementary \cref{app:init}. For the sets that don't have manual matches between points and pixels (e.g., ETH3D, CARLA, RGBDScenes), we sample 100 random points from each image's depth map. This initialization technique has the advantage of avoiding a fixed translation perturbation, which might have a varying effect on the reprojection error due to parallax. We chose an initial perturbation of $10$ px as it was, on average, our observed residual reprojection errors for the nuScenes and Tanks and Temples datasets. In practice, an error of about 10 pixels can induce translation discrepancies of several tens of centimeters in outdoor scenes (\textit{e.g.} CARLA). Each camera pose was perturbed from random starting positions twice for the CARLA and ETH3D sets, and 5 times for the others, adding up to a common 100 registrations per set (except Livox, which has 20). \\

\textbf{Metrics: }We report results in terms of percentage of registrations below a given threshold $t_h$ in pixels, denoted pix@$t_h$ with $\text{pix@}t_h = \frac{1}{N}\sum\limits_i\mathbbm{1}(p_i < t_h)$, $N$ the number of evaluated pairs, and $p_i$ the resulting pixel error of sample $i$. We also give median pixel errors across all registered scenes per set. Because translation error is scale-dependent, we only report it along with rotation error in supplementary \cref{app:rot_results}. Both metrics are respectively defined for two rigid transform matrices {\tiny ${T = \begin{pmatrix}
    R & t \\
    0 & 1
\end{pmatrix}}$} and {\tiny ${T' = \begin{pmatrix}
    R' & t' \\
    0 & 1
\end{pmatrix}}$ } as $\text{transl} = \lVert t' - t \rVert$ and $\text{rot} = \arccos((\text{trace}(R'^TR) - 1) / 2)$.

\subsection{Registration with raw point cloud}
\label{subsec:exp_geom}
In this first setup, we assume that no other information than the point cloud geometry is available. Indeed, this is a rather common setup, as for low-cost LiDARs, the point intensity is strongly correlated with the laser range or incidence angle, and not the underlying material, which makes this channel unusable when fusing multiple scans in a dense point cloud~\cite{s151128099}. From the raw point cloud, we compute normal directions using 50-nearest-neighbor PCA. Since the benchmarking methods~\cite{He2025MatchAnythingUC, yuanPixelLevelExtrinsicSelf2021, koideGeneralSingleshotTargetless2023, Matsuki2023GaussianSS} only take as input mono-intensity images, we synthesize the mean normal value $p^n$ and compare it with the image grayscale intensity $i^g$. For our method, which is capable of handling multiple channels at once, we use each normal's coordinate individually, compared with the image's grayscale value $i_g$:\\
$\mathcal{C}_{\leftrightarrow} = \{(p^\mathit{nx}, i^g), (p^\mathit{ny}, i^g), (p^\mathit{nz}, i^g)\}$.

\begin{table}[h] 
    \centering \scriptsize \resizebox{\linewidth}{!}{%
\addtolength{\tabcolsep}{-0.4em}
\begin{tabular}{@{}lrrrrrrrrrrrrrrrrrrrrrrr}
\toprule
& \multicolumn{4}{c}{ETH3D}
& \multicolumn{4}{c}{Tanks and Temples}
& \multicolumn{4}{c}{CARLA}
& \multicolumn{3}{c}{Livox}
& \multicolumn{4}{c}{nuScenes}
& \multicolumn{4}{c}{RGBDScenes}
\\
\cmidrule(lr){2-5}
\cmidrule(lr){6-9}
\cmidrule(lr){10-13}
\cmidrule(lr){14-16}
\cmidrule(lr){17-20}
\cmidrule(lr){21-24}
& \multicolumn{3}{c}{pix@ (\%$\uparrow$)}  & $p_e$(px$\downarrow$)
& \multicolumn{3}{c}{pix@ (\%$\uparrow$)}  & $p_e$(px$\downarrow$)
& \multicolumn{3}{c}{pix@ (\%$\uparrow$)}  & $p_e$(px$\downarrow$)
& \multicolumn{2}{c}{pix@ (\%$\uparrow$)}  & $p_e$(px$\downarrow$)
& \multicolumn{3}{c}{pix@ (\%$\uparrow$)}  & $p_e$(px$\downarrow$)
& \multicolumn{3}{c}{pix@ (\%$\uparrow$)}  & $p_e$(px$\downarrow$)
\\
& 1 & 2 & 5 & 
& 2 & 3 & 5 & 
& 1 & 2 & 5 & 
& 5 & 10 & 
& 3 & 5 & 10 & 
& 3 & 5 & 10 &\\
\midrule 
\textbf{One stage} \\
GS-SLAM~\cite{Matsuki2023GaussianSS}
	 & {0} & {0} & {0} & {\smallpm{123.23}{64.59}} & {0} & {0} & {0} & {\smallpm{264.12}{181.23}} & {0} & {0} & {0} & {\smallpm{371.52}{214.75}} & {0} & {0} & {\smallpm{113.85}{34.13}} & \cellcolor{orange!40} {0} & {0} & {0} & {\smallpm{309.77}{110.45}} & \cellcolor{orange!40} {0} & {0} & {0} & {\smallpm{90.32}{32.97}}\\
Yuan \etal~\cite{yuanPixelLevelExtrinsicSelf2021}
	 & {0} & {2} & {20} & {\smallpm{10.29}{9.74}} & {1} & {17} & {52} & {\smallpm{4.39}{7.71}} & {0} & {2} & {4} & {\smallpm{15.20}{41.70}} & {25} & \cellcolor{red!40} {100} & {\smallpm{5.83}{1.21}} & \cellcolor{orange!40} {0} & {0} & {5} & {\smallpm{21.10}{37.53}} & \cellcolor{orange!40} {0} & {0} & {5} & {\smallpm{26.95}{26.01}}\\
Koide \etal~\cite{koideGeneralSingleshotTargetless2023}
	 & {0} & {3} & {25} & {\smallpm{11.10}{30.86}} & {0} & {0} & {4} & {\smallpm{29.00}{67.16}} & \cellcolor{orange!40} {2} & {8} & {28} & {\smallpm{8.82}{10.73}} & {25} & {75} & {\smallpm{6.61}{2.60}} & \cellcolor{orange!40} {0} & {0} & {40} & {\smallpm{12.03}{12.21}} & \cellcolor{orange!40} {0} & {0} & {5} & {\smallpm{36.19}{25.14}}\\
Taylor \etal~\cite{Taylor2015MultiModalSC}
	 & {0} & {1} & {27} & {\smallpm{7.13}{3.00}} & \cellcolor{orange!40} {2} & {5} & {18} & {\smallpm{7.79}{2.47}} & {0} & {0} & {10} & {\smallpm{8.87}{3.75}} & {0} & {75} & {\smallpm{8.35}{1.09}} & \cellcolor{orange!40} {0} & {0} & {25} & {\smallpm{14.61}{26.77}} & \cellcolor{orange!40} {0} & \cellcolor{orange!40} {10} & {25} & {\smallpm{12.77}{7.73}}\\
MatchAnything~\cite{He2025MatchAnythingUC}
	 & \cellcolor{orange!40} {8} & \cellcolor{orange!40} {65} & \cellcolor{red!40} {100} & \cellcolor{orange!40} {\smallpm{1.80}{0.83}} & {1} & \cellcolor{orange!40} {46} & \cellcolor{red!40} {95} & \cellcolor{orange!40} {\smallpm{3.05}{0.94}} & {0} & \cellcolor{orange!40} {14} & \cellcolor{orange!40} {65} & \cellcolor{orange!40} {\smallpm{4.12}{3.88}} & \cellcolor{red!40} {65} & \cellcolor{red!40} {100} & \cellcolor{red!40} {\smallpm{4.44}{0.86}} & \cellcolor{orange!40} {0} & \cellcolor{orange!40} {28} & \cellcolor{red!40} {88} & \cellcolor{red!40} {\smallpm{5.71}{2.56}} & \cellcolor{orange!40} {0} & \cellcolor{red!40} {28} & \cellcolor{red!40} {95} & \cellcolor{red!40} {\smallpm{6.07}{1.90}}\\
ReFineR
	 & \cellcolor{red!40} {41} & \cellcolor{red!40} {73} & \cellcolor{orange!40} {79} & \cellcolor{red!40} {\smallpm{1.10}{3.17}} & \cellcolor{red!40} {64} & \cellcolor{red!40} {88} & \cellcolor{orange!40} {94} & \cellcolor{red!40} {\smallpm{1.86}{1.87}} & \cellcolor{red!40} {12} & \cellcolor{red!40} {44} & \cellcolor{red!40} {74} & \cellcolor{red!40} {\smallpm{2.36}{3.45}} & \cellcolor{orange!40} {30} & \cellcolor{orange!40} {90} & \cellcolor{orange!40} {\smallpm{5.50}{2.13}} & \cellcolor{red!40} {9} & \cellcolor{red!40} {41} & \cellcolor{orange!40} {87} & \cellcolor{orange!40} {\smallpm{6.29}{3.08}} & \cellcolor{red!40} {1} & {8} & \cellcolor{orange!40} {72} & \cellcolor{orange!40} {\smallpm{8.96}{2.08}}\\
\midrule 
\textbf{Two stages} \\
MatchAnything + GS-SLAM~\cite{Matsuki2023GaussianSS}
	 & {0} & {0} & {0} & {\smallpm{122.55}{64.26}} & {0} & {0} & {0} & {\smallpm{219.32}{204.55}} & {0} & {0} & {0} & {\smallpm{356.62}{492.77}} & {0} & {0} & {\smallpm{117.12}{33.88}} & \cellcolor{orange!40} {0} & {0} & {0} & {\smallpm{320.79}{110.81}} & {0} & {0} & {0} & {\smallpm{86.55}{43.01}}\\
MatchAnything + Yuan \etal~\cite{yuanPixelLevelExtrinsicSelf2021}
	 & {0} & {1} & {30} & {\smallpm{8.99}{48.80}} & {0} & {11} & {55} & {\smallpm{4.56}{7.63}} & {0} & {2} & {13} & {\smallpm{12.32}{17.25}} & \cellcolor{orange!40} {50} & \cellcolor{red!40} {100} & \cellcolor{red!40} {\smallpm{4.94}{1.30}} & \cellcolor{orange!40} {0} & {2} & {14} & {\smallpm{21.78}{44.35}} & {0} & {0} & {5} & {\smallpm{32.08}{20.41}}\\
MatchAnything + Koide \etal~\cite{koideGeneralSingleshotTargetless2023}
	 & {1} & {14} & {45} & {\smallpm{5.90}{17.58}} & {0} & {1} & {7} & {\smallpm{22.25}{69.57}} & {0} & {5} & {29} & {\smallpm{7.70}{22.69}} & {0} & \cellcolor{orange!40} {85} & {\smallpm{7.16}{1.92}} & \cellcolor{orange!40} {0} & {13} & {54} & {\smallpm{8.38}{46.37}} & {0} & {0} & {12} & {\smallpm{37.68}{26.45}}\\
MatchAnything + Taylor \etal~\cite{Taylor2015MultiModalSC}
	 & \cellcolor{orange!40} {21} & \cellcolor{orange!40} {74} & {96} & \cellcolor{orange!40} {\smallpm{1.48}{1.85}} & \cellcolor{orange!40} {12} & {65} & {98} & {\smallpm{2.72}{0.81}} & {0} & {9} & {51} & {\smallpm{4.97}{5.77}} & {25} & \cellcolor{red!40} {100} & {\smallpm{5.69}{1.17}} & \cellcolor{orange!40} {0} & {31} & {80} & {\smallpm{6.62}{42.65}} & {1} & {27} & {63} & {\smallpm{7.31}{6.01}}\\
MatchAnything + MatchAnything~\cite{He2025MatchAnythingUC}
	 & {14} & {70} & \cellcolor{orange!40} {99} & {\smallpm{1.58}{0.85}} & {8} & \cellcolor{orange!40} {71} & \cellcolor{orange!40} {99} & \cellcolor{orange!40} {\smallpm{2.63}{0.77}} & \cellcolor{orange!40} {3} & \cellcolor{orange!40} {24} & \cellcolor{orange!40} {66} & \cellcolor{orange!40} {\smallpm{3.38}{4.31}} & \cellcolor{orange!40} {50} & \cellcolor{red!40} {100} & {\smallpm{5.13}{0.76}} & \cellcolor{orange!40} {0} & \cellcolor{orange!40} {32} & \cellcolor{red!40} {98} & \cellcolor{orange!40} {\smallpm{5.70}{1.93}} & \cellcolor{orange!40} {9} & \cellcolor{orange!40} {37} & \cellcolor{orange!40} {91} & \cellcolor{orange!40} {\smallpm{5.58}{2.51}}\\
MatchAnything* + ReFineR
	 & \cellcolor{red!40} {58} & \cellcolor{red!40} {96} & \cellcolor{red!40} {100} & \cellcolor{red!40} {\smallpm{0.90}{0.40}} & \cellcolor{red!40} {70} & \cellcolor{red!40} {95} & \cellcolor{red!40} {100} & \cellcolor{red!40} {\smallpm{1.76}{0.51}} & \cellcolor{red!40} {16} & \cellcolor{red!40} {68} & \cellcolor{red!40} {91} & \cellcolor{red!40} {\smallpm{1.54}{3.04}} & \cellcolor{red!40} {55} & \cellcolor{red!40} {100} & \cellcolor{orange!40} {\smallpm{4.94}{0.94}} & \cellcolor{red!40} {6} & \cellcolor{red!40} {54} & \cellcolor{orange!40} {97} & \cellcolor{red!40} {\smallpm{4.90}{2.75}} & \cellcolor{red!40} {13} & \cellcolor{red!40} {43} & \cellcolor{red!40} {92} & \cellcolor{red!40} {\smallpm{5.36}{4.25}}\\

    \bottomrule
        \end{tabular}} 
 \vspace{1pt}\caption{Fine image to raw point cloud registration results (\cref{subsec:exp_geom}). Each image was registered using rendered point cloud normals.}\label{tab:geom}
    \end{table}

Results from~\cref{tab:geom} (one stage) show that for the more geometrically accurate datasets, ReFineR greatly improves on MatchAnything's baseline ($41\%$ subpixel registration on ETH3D versus $8\%$, $44\%$ below $2$px on CARLA against $14\%$ for MatchAnything) but lacks robustness for noisier datasets (Livox and RGBDScenes) and underperforms in the 5--10px ranges in which MatchAnything excels. It is unsurprising that this experimental setup is a failing case for GS-SLAM and Koide \etal as both these methods directly compare point intensities with the image using photometric loss and Mutual Information, but here normal orientation values are very weakly correlated with real image color. This explains the large error values for GS-SLAM, since it completely diverges in most cases. Yuan \etal's method does not seem robust for this fine registration task for most datasets. Their method extracts edges from the point cloud and matches them with 2D edges after projection. This procedure may be efficient for the specific Livox scans, which were taken at fixed positions and therefore do not contain occluded edges. The occluded edges are not filtered by the algorithm and could create unwanted supervision and wrong alignment. Additionally, 3D edges might be more difficult to extract in the first place, using plane intersection for sparser point clouds, as in CARLA and nuScenes. Taking advantage of MatchAnything's and ReFineR's strengths, we combine the two algorithms into our complete registration pipeline (two stages). A multi-stage approach was also considered by Koide \etal using SuperGlue~\cite{sarlinSuperGlueLearningFeature2019}. To do so, we propose to use our point splatting rendering of the point cloud as input to MatchAnything instead of point rasterization in order to create a denser image (referred to as MatchAnything*). This choice is closely related to the reconstruction procedure of ETH3D, which fills the empty pixels using nearest neighbors before extracting SIFT~\cite{loweObjectRecognitionLocal1999} features, and we show in the ablation that our design slightly increases the reliability of MatchAnything alone. This combination appears fruitful as performance is increased under all registration thresholds ($58\%$ under $1$px on ETH3D, $68$\% under $2$px on CARLA). This also proves that ReFineR is particularly performant as a "refinement" tool capable of improving registration even at a few pixels of displacement. In comparison, two successive inferences of MatchAnything are not able to reach as fine results. As for computation times, MatchAnything is generally constant, under 2s per image, while the other methods vary between 10s and 40s because of the variable number of iterations and point cloud size. On CARLA, ReFineR and Koide \etal's methods took about 10s, while Yuan \etal's took 45s per image. We provide more insight on computation times in supplementary~\cref{app:times}.

\subsection{Registration with intensity point clouds}
\label{subsec:exp_intens}
In this second setup, we consider the point cloud intensity when available. This intensity may either be a reflectivity value returned by each emitted light ray, which is related to the measured range or material absorption (Livox, nuScenes), or it may be, in some cases, a color value. Color is specific to the Faro Focus scanner of ETH3D and Tanks and Temples, and the RGBD camera of RGBDScenes, as color comes from an embedded, pre-calibrated camera.

\begin{table}[h] 
    \centering \scriptsize \resizebox{\linewidth}{!}{%
\addtolength{\tabcolsep}{-0.4em}
\begin{tabular}{@{}lrrrrrrrrrrrrrrrrrrr}
\toprule
& \multicolumn{4}{c}{ETH3D}
& \multicolumn{4}{c}{Tanks and Temples}
& \multicolumn{3}{c}{Livox}
& \multicolumn{4}{c}{nuScenes}
& \multicolumn{4}{c}{RGBDScenes}
\\
\cmidrule(lr){2-5}
\cmidrule(lr){6-9}
\cmidrule(lr){10-12}
\cmidrule(lr){13-16}
\cmidrule(lr){17-20}
& \multicolumn{3}{c}{pix@ (\%$\uparrow$)}  & $p_e$(px$\downarrow$)
& \multicolumn{3}{c}{pix@ (\%$\uparrow$)}  & $p_e$(px$\downarrow$)
& \multicolumn{2}{c}{pix@ (\%$\uparrow$)}  & $p_e$(px$\downarrow$)
& \multicolumn{3}{c}{pix@ (\%$\uparrow$)}  & $p_e$(px$\downarrow$)
& \multicolumn{3}{c}{pix@ (\%$\uparrow$)}  & $p_e$(px$\downarrow$)
\\
& 1 & 2 & 5 & 
& 2 & 3 & 5 & 
& 5 & 10 & 
& 3 & 5 & 10 & 
& 3 & 5 & 10 &\\
\midrule 
\textbf{One stage} \\
GS-SLAM~\cite{Matsuki2023GaussianSS}
	 & {19} & {40} & {47} & {\smallpm{7.54}{39.00}} & {45} & {62} & {67} & {\smallpm{2.06}{59.53}} & {20} & {25} & {\smallpm{56.41}{32.79}} & {0} & {8} & {8} & {\smallpm{131.07}{94.68}} & {1} & {5} & {11} & {\smallpm{37.54}{29.24}}\\
Koide \etal~\cite{koideGeneralSingleshotTargetless2023}
	 & \cellcolor{orange!40} {57} & \cellcolor{red!40} {98} & \cellcolor{red!40} {100} & \cellcolor{orange!40} {\smallpm{0.93}{0.29}} & \cellcolor{orange!40} {51} & \cellcolor{red!40} {95} & \cellcolor{red!40} {100} & \cellcolor{orange!40} {\smallpm{1.99}{0.50}} & \cellcolor{red!40} {75} & \cellcolor{red!40} {100} & \cellcolor{orange!40} {\smallpm{4.66}{0.28}} & {0} & {10} & {40} & {\smallpm{11.14}{7.43}} & \cellcolor{orange!40} {25} & \cellcolor{orange!40} {45} & \cellcolor{orange!40} {85} & \cellcolor{orange!40} {\smallpm{5.32}{22.24}}\\
Taylor \etal~\cite{Taylor2015MultiModalSC}
	 & {1} & {1} & {29} & {\smallpm{6.76}{5.05}} & {3} & {7} & {23} & {\smallpm{7.25}{3.48}} & {0} & {75} & {\smallpm{9.09}{6.69}} & {0} & {0} & {15} & {\smallpm{22.44}{46.66}} & {0} & {5} & {55} & {\smallpm{9.07}{6.65}}\\
MatchAnything~\cite{He2025MatchAnythingUC}
	 & {35} & \cellcolor{orange!40} {87} & \cellcolor{red!40} {100} & {\smallpm{1.12}{0.61}} & {38} & {87} & \cellcolor{red!40} {100} & {\smallpm{2.13}{0.63}} & \cellcolor{red!40} {75} & \cellcolor{red!40} {100} & \cellcolor{red!40} {\smallpm{4.17}{0.92}} & \cellcolor{orange!40} {4} & \cellcolor{red!40} {64} & \cellcolor{red!40} {100} & \cellcolor{red!40} {\smallpm{4.72}{1.19}} & \cellcolor{red!40} {81} & \cellcolor{red!40} {100} & \cellcolor{red!40} {100} & \cellcolor{red!40} {\smallpm{1.81}{0.93}}\\
ReFineR
	 & \cellcolor{red!40} {61} & {70} & \cellcolor{orange!40} {73} & \cellcolor{red!40} {\smallpm{0.71}{3.62}} & \cellcolor{red!40} {71} & \cellcolor{orange!40} {91} & \cellcolor{orange!40} {98} & \cellcolor{red!40} {\smallpm{1.93}{0.94}} & \cellcolor{orange!40} {45} & \cellcolor{orange!40} {95} & {\smallpm{5.17}{1.68}} & \cellcolor{red!40} {6} & \cellcolor{orange!40} {35} & \cellcolor{orange!40} {84} & \cellcolor{orange!40} {\smallpm{6.99}{2.88}} & {12} & {22} & {81} & {\smallpm{8.07}{3.00}}\\
\midrule 
\textbf{Two stages} \\
MatchAnything + GS-SLAM~\cite{Matsuki2023GaussianSS}
	 & {32} & {66} & {73} & {\smallpm{1.30}{38.55}} & {36} & {74} & {83} & {\smallpm{2.12}{38.45}} & {25} & \cellcolor{orange!40} {25} & {\smallpm{65.64}{30.08}} & {0} & {11} & {17} & {\smallpm{139.64}{101.83}} & {5} & {5} & {10} & {\smallpm{35.67}{31.37}}\\
MatchAnything + Koide \etal~\cite{koideGeneralSingleshotTargetless2023}
	 & {57} & \cellcolor{red!40} {100} & \cellcolor{red!40} {100} & {\smallpm{0.95}{0.25}} & {48} & \cellcolor{red!40} {96} & \cellcolor{red!40} {100} & {\smallpm{2.02}{0.48}} & {50} & \cellcolor{red!40} {100} & {\smallpm{4.99}{0.71}} & {0} & {18} & {68} & {\smallpm{7.99}{10.92}} & {38} & {60} & {85} & {\smallpm{3.96}{27.30}}\\
MatchAnything + Taylor \etal~\cite{Taylor2015MultiModalSC}
	 & \cellcolor{orange!40} {68} & {89} & \cellcolor{orange!40} {98} & \cellcolor{orange!40} {\smallpm{0.81}{1.45}} & {36} & {91} & {97} & {\smallpm{2.23}{0.95}} & \cellcolor{orange!40} {70} & \cellcolor{red!40} {100} & {\smallpm{4.48}{0.63}} & \cellcolor{orange!40} {4} & {45} & \cellcolor{orange!40} {80} & {\smallpm{5.35}{26.49}} & {50} & \cellcolor{orange!40} {78} & \cellcolor{orange!40} {93} & {\smallpm{2.99}{3.53}}\\
MatchAnything + MatchAnything~\cite{He2025MatchAnythingUC}
	 & {62} & \cellcolor{orange!40} {96} & \cellcolor{red!40} {100} & {\smallpm{0.90}{0.43}} & \cellcolor{orange!40} {52} & \cellcolor{red!40} {96} & \cellcolor{orange!40} {99} & \cellcolor{orange!40} {\smallpm{1.98}{0.64}} & \cellcolor{red!40} {75} & \cellcolor{red!40} {100} & \cellcolor{red!40} {\smallpm{4.12}{0.57}} & {2} & \cellcolor{orange!40} {54} & \cellcolor{red!40} {100} & \cellcolor{orange!40} {\smallpm{4.83}{1.24}} & \cellcolor{orange!40} {87} & \cellcolor{red!40} {100} & \cellcolor{red!40} {100} & \cellcolor{orange!40} {\smallpm{2.00}{0.64}}\\
MatchAnything* + ReFineR
	 & \cellcolor{red!40} {86} & \cellcolor{red!40} {100} & \cellcolor{red!40} {100} & \cellcolor{red!40} {\smallpm{0.57}{0.29}} & \cellcolor{red!40} {74} & \cellcolor{orange!40} {95} & \cellcolor{red!40} {100} & \cellcolor{red!40} {\smallpm{1.90}{0.55}} & \cellcolor{red!40} {75} & \cellcolor{red!40} {100} & \cellcolor{orange!40} {\smallpm{4.17}{0.63}} & \cellcolor{red!40} {10} & \cellcolor{red!40} {62} & \cellcolor{red!40} {100} & \cellcolor{red!40} {\smallpm{4.29}{1.40}} & \cellcolor{red!40} {99} & \cellcolor{red!40} {100} & \cellcolor{red!40} {100} & \cellcolor{red!40} {\smallpm{1.39}{0.54}}\\

    \bottomrule
        \end{tabular}} 
 \vspace{1pt}\caption{Fine image to intensity point cloud registration results (\cref{subsec:exp_intens}). Each image was registered using rendered point cloud reflectivity or grayscale value.}\label{tab:intens}
    \end{table}
    
We observe that intensity provides, in general, richer information for registration. This is naturally the case for the color channel, but also for reflectivity, as drawn patterns (white arrows on roads...) will now appear on reflectivity images, thanks to the uneven absorption strength of colored surfaces. Examples of rendered images are shown in supplementary \cref{fig:synth}. Thanks to the color information, photometric losses become a relevant supervision channel, which allows GS-SLAM to perform well on ETH3D and Tanks and Temples and to obtain competitive results on RGBDScenes, but it fails on the remaining reflectivity sets. Koide \etal's method is a little more resilient as it compares intensity histograms to extract correlation information between image color and LiDAR reflectivity. Our 2-stage variant with the MatchAnything* + ReFineR combination obtained near $100\%$ subpixel registration for ETH3D and best results at lower thresholds for all sets. We show qualitative results of both experiments in supplementary \cref{fig:qualit_geom} and \cref{fig:qualit_intens}.

\subsection{Ablation studies}
\label{subsec:ablation}
Our design choices are evaluated in \cref{tab:ablation} on all datasets with raw point clouds. As before, each experiment is initialized at a $10$px mean error. Experiments (I-IV) incorporate the MatchAnything* first stage with variations of ReFineR. Experiment (V) shows the performance gain from MatchAnything* alone. Experiment (VI) compares two stages of MatchAnything and MatchAnything*.

\begin{table}[h] 
    \centering \scriptsize \resizebox{\linewidth}{!}{%
\addtolength{\tabcolsep}{-0.4em}
\begin{tabular}{@{}lrrrrrrrrrrrrrrrrrrrrrrrr}
\toprule
& \multicolumn{4}{c}{ETH3D}
& \multicolumn{5}{c}{Tanks and Temples}
& \multicolumn{4}{c}{CARLA}
& \multicolumn{3}{c}{Livox}
& \multicolumn{4}{c}{nuScenes}
& \multicolumn{4}{c}{RGBDScenes}
\\
\cmidrule(lr){2-5}
\cmidrule(lr){6-10}
\cmidrule(lr){11-14}
\cmidrule(lr){15-17}
\cmidrule(lr){18-21}
\cmidrule(lr){22-25}
& \multicolumn{3}{c}{pix@ (\%$\uparrow$)}  & $p_e$(px$\downarrow$)
& \multicolumn{4}{c}{pix@ (\%$\uparrow$)}  & $p_e$(px$\downarrow$)
& \multicolumn{3}{c}{pix@ (\%$\uparrow$)}  & $p_e$(px$\downarrow$)
& \multicolumn{2}{c}{pix@ (\%$\uparrow$)}  & $p_e$(px$\downarrow$)
& \multicolumn{3}{c}{pix@ (\%$\uparrow$)}  & $p_e$(px$\downarrow$)
& \multicolumn{3}{c}{pix@ (\%$\uparrow$)}  & $p_e$(px$\downarrow$)
\\
& 1 & 2 & 5 & 
& 1 & 2 & 3 & 5 & 
& 1 & 2 & 5 & 
& 5 & 10 & 
& 3 & 5 & 10 & 
& 3 & 5 & 10 &\\
\midrule 
\textbf{(I)} \\
W/o rescale
	 & \cellcolor{orange!40} {57} & \cellcolor{red!40} {97} & \cellcolor{red!40} {100} & \cellcolor{orange!40} {\smallpm{0.90}{0.38}} & \cellcolor{red!40} {0} & \cellcolor{red!40} {70} & \cellcolor{red!40} {95} & \cellcolor{red!40} {100} & \cellcolor{red!40} {\smallpm{1.75}{0.48}} & \cellcolor{red!40} {17} & \cellcolor{orange!40} {61} & \cellcolor{orange!40} {88} & \cellcolor{orange!40} {\smallpm{1.66}{2.97}} & \cellcolor{orange!40} {50} & \cellcolor{red!40} {100} & \cellcolor{orange!40} {\smallpm{5.10}{0.96}} & \cellcolor{red!40} {9} & \cellcolor{orange!40} {50} & \cellcolor{orange!40} {95} & \cellcolor{orange!40} {\smallpm{5.00}{3.59}} & \cellcolor{orange!40} {10} & \cellcolor{red!40} {44} & \cellcolor{red!40} {93} & \cellcolor{orange!40} {\smallpm{5.43}{3.35}}\\
MatchAnything* + ReFineR (W/ rescale)
	 & \cellcolor{red!40} {58} & \cellcolor{orange!40} {96} & \cellcolor{red!40} {100} & \cellcolor{red!40} {\smallpm{0.90}{0.40}} & \cellcolor{red!40} {0} & \cellcolor{red!40} {70} & \cellcolor{red!40} {95} & \cellcolor{red!40} {100} & \cellcolor{orange!40} {\smallpm{1.76}{0.51}} & \cellcolor{orange!40} {16} & \cellcolor{red!40} {68} & \cellcolor{red!40} {91} & \cellcolor{red!40} {\smallpm{1.54}{3.04}} & \cellcolor{red!40} {55} & \cellcolor{red!40} {100} & \cellcolor{red!40} {\smallpm{4.94}{0.94}} & \cellcolor{orange!40} {6} & \cellcolor{red!40} {54} & \cellcolor{red!40} {97} & \cellcolor{red!40} {\smallpm{4.90}{2.75}} & \cellcolor{red!40} {13} & \cellcolor{orange!40} {43} & \cellcolor{orange!40} {92} & \cellcolor{red!40} {\smallpm{5.36}{4.25}}\\
\midrule 
\textbf{(II)} \\
Opacity-0.5
	 & \cellcolor{orange!40} {51} & \cellcolor{red!40} {98} & \cellcolor{red!40} {100} & \cellcolor{orange!40} {\smallpm{0.99}{0.37}} & \cellcolor{red!40} {0} & \cellcolor{red!40} {70} & {88} & \cellcolor{red!40} {100} & \cellcolor{orange!40} {\smallpm{1.78}{0.55}} & \cellcolor{orange!40} {9} & \cellcolor{orange!40} {56} & \cellcolor{red!40} {93} & \cellcolor{orange!40} {\smallpm{1.89}{2.90}} & \cellcolor{red!40} {80} & \cellcolor{red!40} {100} & \cellcolor{red!40} {\smallpm{4.31}{1.18}} & \cellcolor{orange!40} {1} & \cellcolor{orange!40} {48} & \cellcolor{orange!40} {96} & \cellcolor{orange!40} {\smallpm{5.26}{3.05}} & \cellcolor{orange!40} {4} & \cellcolor{orange!40} {23} & \cellcolor{orange!40} {82} & \cellcolor{orange!40} {\smallpm{6.69}{3.36}}\\
Opacity-0.8
	 & {35} & \cellcolor{red!40} {98} & \cellcolor{red!40} {100} & {\smallpm{1.09}{0.39}} & \cellcolor{red!40} {0} & \cellcolor{red!40} {70} & \cellcolor{orange!40} {89} & \cellcolor{red!40} {100} & {\smallpm{1.86}{0.62}} & {2} & {40} & {81} & {\smallpm{2.12}{2.87}} & {25} & \cellcolor{red!40} {100} & {\smallpm{5.15}{1.44}} & {0} & {37} & {95} & {\smallpm{5.47}{2.30}} & \cellcolor{orange!40} {4} & {12} & {72} & {\smallpm{7.51}{3.69}}\\
MatchAnything* + ReFineR (Opacity-0.2)
	 & \cellcolor{red!40} {58} & \cellcolor{orange!40} {96} & \cellcolor{red!40} {100} & \cellcolor{red!40} {\smallpm{0.90}{0.40}} & \cellcolor{red!40} {0} & \cellcolor{red!40} {70} & \cellcolor{red!40} {95} & \cellcolor{red!40} {100} & \cellcolor{red!40} {\smallpm{1.76}{0.51}} & \cellcolor{red!40} {16} & \cellcolor{red!40} {68} & \cellcolor{orange!40} {91} & \cellcolor{red!40} {\smallpm{1.54}{3.04}} & \cellcolor{orange!40} {55} & \cellcolor{red!40} {100} & \cellcolor{orange!40} {\smallpm{4.94}{0.94}} & \cellcolor{red!40} {6} & \cellcolor{red!40} {54} & \cellcolor{red!40} {97} & \cellcolor{red!40} {\smallpm{4.90}{2.75}} & \cellcolor{red!40} {13} & \cellcolor{red!40} {43} & \cellcolor{red!40} {92} & \cellcolor{red!40} {\smallpm{5.36}{4.25}}\\
\midrule 
\textbf{(III)} \\
Size-0.2
	 & \cellcolor{red!40} {63} & \cellcolor{red!40} {98} & \cellcolor{red!40} {100} & \cellcolor{red!40} {\smallpm{0.85}{0.38}} & \cellcolor{red!40} {0} & \cellcolor{red!40} {70} & \cellcolor{red!40} {95} & \cellcolor{red!40} {100} & \cellcolor{red!40} {\smallpm{1.71}{0.61}} & \cellcolor{orange!40} {13} & \cellcolor{orange!40} {60} & {86} & \cellcolor{orange!40} {\smallpm{1.63}{3.67}} & {30} & \cellcolor{red!40} {100} & {\smallpm{5.68}{1.30}} & \cellcolor{orange!40} {5} & {42} & {91} & {\smallpm{6.08}{12.93}} & \cellcolor{orange!40} {7} & \cellcolor{red!40} {43} & \cellcolor{red!40} {94} & \cellcolor{orange!40} {\smallpm{5.53}{3.19}}\\
Size-0.8
	 & {40} & {95} & \cellcolor{red!40} {100} & {\smallpm{1.15}{0.47}} & \cellcolor{red!40} {0} & \cellcolor{orange!40} {68} & \cellcolor{orange!40} {85} & \cellcolor{red!40} {100} & {\smallpm{1.89}{0.70}} & {6} & {55} & \cellcolor{red!40} {94} & {\smallpm{1.81}{3.04}} & \cellcolor{orange!40} {45} & \cellcolor{red!40} {100} & \cellcolor{orange!40} {\smallpm{5.02}{1.21}} & {0} & \cellcolor{orange!40} {49} & \cellcolor{orange!40} {94} & \cellcolor{orange!40} {\smallpm{5.03}{4.88}} & {5} & \cellcolor{orange!40} {38} & {86} & {\smallpm{5.68}{3.33}}\\
MatchAnything* + ReFineR (Size-0.5)
	 & \cellcolor{orange!40} {58} & \cellcolor{orange!40} {96} & \cellcolor{red!40} {100} & \cellcolor{orange!40} {\smallpm{0.90}{0.40}} & \cellcolor{red!40} {0} & \cellcolor{red!40} {70} & \cellcolor{red!40} {95} & \cellcolor{red!40} {100} & \cellcolor{orange!40} {\smallpm{1.76}{0.51}} & \cellcolor{red!40} {16} & \cellcolor{red!40} {68} & \cellcolor{orange!40} {91} & \cellcolor{red!40} {\smallpm{1.54}{3.04}} & \cellcolor{red!40} {55} & \cellcolor{red!40} {100} & \cellcolor{red!40} {\smallpm{4.94}{0.94}} & \cellcolor{red!40} {6} & \cellcolor{red!40} {54} & \cellcolor{red!40} {97} & \cellcolor{red!40} {\smallpm{4.90}{2.75}} & \cellcolor{red!40} {13} & \cellcolor{red!40} {43} & \cellcolor{orange!40} {92} & \cellcolor{red!40} {\smallpm{5.36}{4.25}}\\
\midrule 
\textbf{(IV)} \\
MatchAnything features
	 & {5} & {34} & \cellcolor{orange!40} {81} & {\smallpm{3.09}{5.99}} & \cellcolor{red!40} {0} & {12} & {51} & \cellcolor{orange!40} {89} & {\smallpm{2.96}{1.56}} & {0} & {2} & {50} & {\smallpm{4.97}{5.83}} & \cellcolor{red!40} {70} & \cellcolor{red!40} {100} & \cellcolor{red!40} {\smallpm{4.73}{0.82}} & {0} & {21} & {77} & {\smallpm{6.91}{6.95}} & {4} & {34} & {87} & {\smallpm{5.86}{3.39}}\\
HOG features
	 & \cellcolor{red!40} {63} & \cellcolor{red!40} {98} & \cellcolor{red!40} {100} & \cellcolor{red!40} {\smallpm{0.86}{0.41}} & \cellcolor{red!40} {0} & \cellcolor{orange!40} {65} & \cellcolor{red!40} {100} & \cellcolor{red!40} {100} & \cellcolor{orange!40} {\smallpm{1.79}{0.41}} & \cellcolor{orange!40} {13} & \cellcolor{orange!40} {58} & \cellcolor{red!40} {93} & \cellcolor{orange!40} {\smallpm{1.72}{2.36}} & {45} & \cellcolor{red!40} {100} & {\smallpm{5.05}{0.65}} & \cellcolor{red!40} {10} & \cellcolor{red!40} {60} & \cellcolor{red!40} {99} & \cellcolor{red!40} {\smallpm{4.55}{2.11}} & \cellcolor{orange!40} {11} & \cellcolor{red!40} {48} & \cellcolor{red!40} {95} & \cellcolor{red!40} {\smallpm{5.08}{2.99}}\\
MatchAnything* + ReFineR (GOM features)
	 & \cellcolor{orange!40} {58} & \cellcolor{orange!40} {96} & \cellcolor{red!40} {100} & \cellcolor{orange!40} {\smallpm{0.90}{0.40}} & \cellcolor{red!40} {0} & \cellcolor{red!40} {70} & \cellcolor{orange!40} {95} & \cellcolor{red!40} {100} & \cellcolor{red!40} {\smallpm{1.76}{0.51}} & \cellcolor{red!40} {16} & \cellcolor{red!40} {68} & \cellcolor{orange!40} {91} & \cellcolor{red!40} {\smallpm{1.54}{3.04}} & \cellcolor{orange!40} {55} & \cellcolor{red!40} {100} & \cellcolor{orange!40} {\smallpm{4.94}{0.94}} & \cellcolor{orange!40} {6} & \cellcolor{orange!40} {54} & \cellcolor{orange!40} {97} & \cellcolor{orange!40} {\smallpm{4.90}{2.75}} & \cellcolor{red!40} {13} & \cellcolor{orange!40} {43} & \cellcolor{orange!40} {92} & \cellcolor{orange!40} {\smallpm{5.36}{4.25}}\\
\midrule 
\textbf{(V)} \\
MatchAnything
	 & \cellcolor{orange!40} {8} & \cellcolor{orange!40} {65} & \cellcolor{red!40} {100} & \cellcolor{orange!40} {\smallpm{1.80}{0.83}} & \cellcolor{red!40} {0} & \cellcolor{orange!40} {1} & \cellcolor{orange!40} {46} & \cellcolor{orange!40} {95} & \cellcolor{orange!40} {\smallpm{3.05}{0.94}} & \cellcolor{red!40} {0} & \cellcolor{red!40} {14} & \cellcolor{orange!40} {65} & \cellcolor{orange!40} {\smallpm{4.12}{3.88}} & \cellcolor{red!40} {65} & \cellcolor{red!40} {100} & \cellcolor{red!40} {\smallpm{4.44}{0.86}} & \cellcolor{orange!40} {0} & \cellcolor{orange!40} {28} & \cellcolor{orange!40} {88} & \cellcolor{orange!40} {\smallpm{5.71}{2.56}} & \cellcolor{orange!40} {0} & \cellcolor{orange!40} {28} & \cellcolor{red!40} {95} & \cellcolor{orange!40} {\smallpm{6.07}{1.90}}\\
MatchAnything*
	 & \cellcolor{red!40} {12} & \cellcolor{red!40} {66} & \cellcolor{red!40} {100} & \cellcolor{red!40} {\smallpm{1.61}{0.74}} & \cellcolor{red!40} {0} & \cellcolor{red!40} {14} & \cellcolor{red!40} {77} & \cellcolor{red!40} {100} & \cellcolor{red!40} {\smallpm{2.67}{0.59}} & \cellcolor{red!40} {0} & \cellcolor{orange!40} {9} & \cellcolor{red!40} {85} & \cellcolor{red!40} {\smallpm{3.21}{2.44}} & \cellcolor{orange!40} {50} & \cellcolor{red!40} {100} & \cellcolor{orange!40} {\smallpm{4.93}{1.31}} & \cellcolor{red!40} {5} & \cellcolor{red!40} {30} & \cellcolor{red!40} {100} & \cellcolor{red!40} {\smallpm{5.69}{1.48}} & \cellcolor{red!40} {6} & \cellcolor{red!40} {38} & \cellcolor{orange!40} {83} & \cellcolor{red!40} {\smallpm{5.99}{3.20}}\\
\midrule 
\textbf{(VI)} \\
MatchAnything + MatchAnything
	 & \cellcolor{orange!40} {14} & \cellcolor{orange!40} {70} & \cellcolor{orange!40} {99} & {\smallpm{1.58}{0.85}} & \cellcolor{red!40} {0} & {8} & {71} & \cellcolor{orange!40} {99} & {\smallpm{2.63}{0.77}} & {3} & {24} & {66} & {\smallpm{3.38}{4.31}} & \cellcolor{orange!40} {50} & \cellcolor{red!40} {100} & {\smallpm{5.13}{0.76}} & {0} & {32} & \cellcolor{red!40} {98} & {\smallpm{5.70}{1.93}} & \cellcolor{orange!40} {9} & {37} & \cellcolor{orange!40} {91} & \cellcolor{orange!40} {\smallpm{5.58}{2.51}}\\
MatchAnything* + MatchAnything*
	 & {11} & {69} & \cellcolor{red!40} {100} & {\smallpm{1.63}{0.72}} & \cellcolor{red!40} {0} & {9} & {80} & \cellcolor{red!40} {100} & {\smallpm{2.60}{0.51}} & {0} & {12} & {79} & {\smallpm{3.33}{2.51}} & \cellcolor{orange!40} {50} & \cellcolor{red!40} {100} & \cellcolor{red!40} {\smallpm{4.93}{1.26}} & \cellcolor{orange!40} {5} & {29} & \cellcolor{orange!40} {97} & {\smallpm{5.62}{3.63}} & {5} & {34} & {80} & {\smallpm{7.11}{3.84}}\\
MatchAnything + ReFineR
	 & \cellcolor{red!40} {58} & \cellcolor{red!40} {96} & \cellcolor{red!40} {100} & \cellcolor{orange!40} {\smallpm{0.92}{0.43}} & \cellcolor{red!40} {0} & \cellcolor{red!40} {71} & \cellcolor{orange!40} {90} & \cellcolor{red!40} {100} & \cellcolor{orange!40} {\smallpm{1.84}{0.55}} & \cellcolor{red!40} {17} & \cellcolor{red!40} {69} & \cellcolor{red!40} {92} & \cellcolor{red!40} {\smallpm{1.53}{2.96}} & {35} & \cellcolor{red!40} {100} & {\smallpm{5.18}{1.48}} & \cellcolor{red!40} {6} & \cellcolor{orange!40} {52} & \cellcolor{red!40} {98} & \cellcolor{red!40} {\smallpm{4.89}{1.69}} & {5} & \cellcolor{orange!40} {39} & {82} & {\smallpm{5.88}{8.47}}\\
MatchAnything* + ReFineR
	 & \cellcolor{red!40} {58} & \cellcolor{red!40} {96} & \cellcolor{red!40} {100} & \cellcolor{red!40} {\smallpm{0.90}{0.40}} & \cellcolor{red!40} {0} & \cellcolor{orange!40} {70} & \cellcolor{red!40} {95} & \cellcolor{red!40} {100} & \cellcolor{red!40} {\smallpm{1.76}{0.51}} & \cellcolor{orange!40} {16} & \cellcolor{orange!40} {68} & \cellcolor{orange!40} {91} & \cellcolor{orange!40} {\smallpm{1.54}{3.04}} & \cellcolor{red!40} {55} & \cellcolor{red!40} {100} & \cellcolor{orange!40} {\smallpm{4.94}{0.94}} & \cellcolor{red!40} {6} & \cellcolor{red!40} {54} & \cellcolor{orange!40} {97} & \cellcolor{orange!40} {\smallpm{4.90}{2.75}} & \cellcolor{red!40} {13} & \cellcolor{red!40} {43} & \cellcolor{red!40} {92} & \cellcolor{red!40} {\smallpm{5.36}{4.25}}\\

    \bottomrule
        \end{tabular}} 
 \vspace{1pt}\caption{Ablation study (\cref{subsec:ablation}).}\label{tab:ablation}
    \end{table}

\textbf{Automatic rescaling (I): }
First, we test the impact of not using our automatic point cloud rescaling factor, keeping the point cloud at its acquired scale. In practice, this implies that since the learning rate remains fixed, translation is less updated at each iteration, and the optimization is more prone to fall into local translation minima. This is why we observe a more significant impact on the CARLA set, as this set has the largest point clouds, being simulated outdoors and on large streets and with large depth ranges. Practically, adding an automatic rescale factor also serves the purpose of fixing the hyperparameters independently of the point cloud scale (learning rate, point size).\\

\textbf{Splatting parameters (II) \& (III): }
We change the values of the opacity and point size parameters used for splatting. These parameters are directly linked to the loss landscape of the optimization, as a larger point size creates larger splats, which can reduce the fine registration ability of the method by having too smooth images, but a smaller point size creates rendering holes, which can bias the gradient computations. We found that using a size $0.5$ gave reasonable results. Similarly, opacity controls a form of smoothness in the optimization, as low opacity increases the individual contribution of each Gaussian, making the loss more robust, but also enables occluded structures to appear in the render. We chose to fix the opacity at $0.2$.\\

\textbf{Loss functions (IV): }
We experimented with the idea of using intermediate MatchAnything features as a replacement for the GOM loss. The intuition was to leverage the model's cross-modality features at a dense and multiscale level. We compared features at scales 2, 4, and 8, computed for the initial RGB and rendered images, using an L2 loss. Due to memory constraints, rendered images at $1024$px were differentiably resized to $664$px before giving them as input to MatchAnything, which we suspect had an impact on the comparatively weak results for this experiment. Additionally, we guess that the presence of pooling layers in the CNN encoder reduces the fine-registration abilities, as these layers divide the image resolution by 2, 4, and 8. We believe that a less naive approach could certainly be proposed to better leverage the MatchAnything features, which may be investigated in future work.
Inspired by Guislain \etal's work~\cite{guislainFineScaleImage2017}, we propose to implement a differentiable version of the HOG features~\cite{dalalHistogramsOrientedGradients2005}. To compute the HOG features, one should first estimate the image gradients, then apply a smooth histogram convolution kernel onto the image, creating cell histograms, followed by a local block normalization of each histogram bin along blocks of cells (an explicit formulation of differentiable HOGs can be found in Supplementary~\cref{app:hog}). The idea behind this loss is quite similar to the GOM, which locally compares images' gradients. This is probably why we observe, on average, comparable results between GOM and HOG features. On average, GOM loss is a little faster than HOG, as the former took on average 0.05s per iteration, whereas the latter took 0.2s. \\

\textbf{MatchAnything (V) \& (VI): }
Comparing performance between MatchAnything inference on image rasters and Gaussian splatting renders, we observe a net gain in the final pixel error for the CARLA and Tanks and Temples datasets, hinting that a denser image is a better quality input for MatchAnything. A very small improvement can be observed on the ETH3D dataset, which is coherent, as this dataset already has very dense point clouds that completely fill the image, even with simple rasterization. When comparing two consecutive inferences of MatchAnything and MatchAnything*, this gap closes, demonstrating the limits of the MatchAnything model itself. The similar performances when adding ReFineR as a second stage tend to show that our method is capable of handling some of MatchAnything's weaker cases, but given results from (VI), we chose to keep MatchAnything* for our final pipeline.

\subsection{Registration at scale}
\label{subsec:exp_scale}
To capture the performance of the MatchAnything* + ReFineR combination, we vary the initialization of our registration experiment from $0$ to $100$ pixels (\cref{tab:scale}). Starting from a $0$px perturbation, we observe that ReFineR alone keeps the registration stable under $2$px $76\%$ of the time, which is about three times as often as MatchAnything alone, but when combined, the fine registration capabilities of ReFineR mitigate MatchAnything's instabilities. As the perturbation increases, our pipeline proves to stay relatively robust even at the large $100$px initialization, keeping its median final reprojection error below $2$px. Another interesting result is the net reduction in translation error, probably caused by the automatic rescaling factor, which allows better conditioning for the joint optimization of rotation and translation. A similar experiment was carried out on the ETH3D set in supplementary \cref{tab:scale2}.

\begin{table}[h] 
    \centering \scriptsize \resizebox{0.75\linewidth}{!}{%
\addtolength{\tabcolsep}{-0.4em}
\begin{tabular}{@{}lrrrrrrrrrrr}
\toprule
& \multicolumn{11}{c}{CARLA}
\\
\cmidrule(lr){2-12}
& \multicolumn{8}{c}{pix@ (\%$\uparrow$)}  & $p_e$(px$\downarrow$) & rot (°$\downarrow$) & transl (cm$\downarrow$)
\\
& 1 & 2 & 3 & 5 & 10 & 20 & 50 & 100 &  &  &\\
\midrule 
\textbf{0px initialization} \\
MatchAnything
	 & {3} & {26} & {44} & {85} & {98} & {100} & {100} & {100} & {\smallpm{3.20}{2.60}} & {\smallpm{0.14}{0.08}} & {\smallpm{33.94}{27.77}}\\
ReFineR
	 & {20} & {76} & {92} & {100} & {100} & {100} & {100} & {100} & {\smallpm{1.47}{0.79}} & {\smallpm{0.06}{0.04}} & {\smallpm{2.99}{1.74}}\\
MatchAnything* + ReFineR
	 & {16} & {71} & {83} & {95} & {96} & {99} & {100} & {100} & {\smallpm{1.49}{2.72}} & {\smallpm{0.05}{0.09}} & {\smallpm{3.03}{6.32}}\\
\midrule 
\textbf{5px initialization} \\
MatchAnything
	 & {2} & {17} & {39} & {78} & {95} & {99} & {100} & {100} & {\smallpm{3.35}{3.20}} & {\smallpm{0.13}{0.12}} & {\smallpm{32.50}{32.13}}\\
ReFineR
	 & {11} & {63} & {82} & {95} & {99} & {100} & {100} & {100} & {\smallpm{1.68}{1.72}} & {\smallpm{0.06}{0.07}} & {\smallpm{3.29}{3.82}}\\
MatchAnything* + ReFineR
	 & {16} & {66} & {84} & {96} & {99} & {100} & {100} & {100} & {\smallpm{1.49}{1.59}} & {\smallpm{0.06}{0.06}} & {\smallpm{3.05}{3.72}}\\
\midrule 
\textbf{10px initialization} \\
MatchAnything
	 & {0} & {14} & {30} & {65} & {90} & {100} & {100} & {100} & {\smallpm{4.12}{3.88}} & {\smallpm{0.15}{0.17}} & {\smallpm{37.95}{53.56}}\\
ReFineR
	 & {12} & {44} & {57} & {74} & {92} & {100} & {100} & {100} & {\smallpm{2.36}{3.45}} & {\smallpm{0.08}{0.25}} & {\smallpm{4.44}{7.52}}\\
MatchAnything* + ReFineR
	 & {16} & {68} & {81} & {91} & {97} & {99} & {100} & {100} & {\smallpm{1.54}{3.04}} & {\smallpm{0.06}{0.09}} & {\smallpm{3.05}{6.39}}\\
\midrule 
\textbf{20px initialization} \\
MatchAnything
	 & {0} & {11} & {19} & {49} & {85} & {96} & {100} & {100} & {\smallpm{5.13}{4.94}} & {\smallpm{0.19}{0.24}} & {\smallpm{43.47}{66.34}}\\
ReFineR
	 & {2} & {14} & {18} & {21} & {26} & {82} & {100} & {100} & {\smallpm{16.67}{7.70}} & {\smallpm{0.94}{0.64}} & {\smallpm{26.50}{16.88}}\\
MatchAnything* + ReFineR
	 & {18} & {66} & {82} & {90} & {94} & {98} & {100} & {100} & {\smallpm{1.54}{4.19}} & {\smallpm{0.06}{0.22}} & {\smallpm{3.06}{8.75}}\\
\midrule 
\textbf{50px initialization} \\
MatchAnything
	 & {0} & {8} & {18} & {34} & {65} & {93} & {98} & {99} & {\smallpm{6.79}{14.99}} & {\smallpm{0.23}{0.32}} & {\smallpm{50.53}{79.29}}\\
ReFineR
	 & {0} & {0} & {0} & {0} & {0} & {0} & {56} & {100} & {\smallpm{49.60}{6.50}} & {\smallpm{3.63}{1.19}} & {\smallpm{82.69}{29.06}}\\
MatchAnything* + ReFineR
	 & {17} & {54} & {73} & {80} & {92} & {96} & {99} & {100} & {\smallpm{1.64}{9.36}} & {\smallpm{0.06}{0.29}} & {\smallpm{3.35}{20.37}}\\
\midrule 
\textbf{100px initialization} \\
MatchAnything
	 & {1} & {3} & {9} & {31} & {55} & {77} & {95} & {100} & {\smallpm{8.32}{15.98}} & {\smallpm{0.28}{0.87}} & {\smallpm{81.66}{202.21}}\\
ReFineR
	 & {0} & {0} & {0} & {0} & {0} & {0} & {0} & {52} & {\smallpm{99.66}{13.05}} & {\smallpm{7.43}{2.28}} & {\smallpm{170.25}{56.97}}\\
MatchAnything* + ReFineR
	 & {12} & {52} & {60} & {69} & {80} & {89} & {95} & {99} & {\smallpm{1.98}{20.24}} & {\smallpm{0.07}{0.65}} & {\smallpm{4.29}{41.36}}\\

    \bottomrule
        \end{tabular}} 
 \vspace{3pt}\caption{Registration at scale experiment, each camera pose was initialized at increasing pixel errors (\cref{subsec:exp_scale}).}\label{tab:scale}
    \end{table}

\section{Limitations of this work}
Several limitations were not addressed in this paper and still represent challenges for the fine image-to-point cloud registration task, which we will discuss here. First, we show in this paper that there is a strong dependence between registration capabilities and the point cloud quality. In particular, floating individual points in the 3D model and in front of the camera might be rendered, biasing the optimization. This is also the case for dynamic objects in the 3D scene, which create discrepancies between the real image and the rendered view. This issue was mitigated by manually removing such points in preprocessing, but results on nuScenes vs ETH3D show that noisier point clouds represent a greater challenge. In general, for any analysis-by-synthesis method, the correct rendering of occlusion is a major challenge: by standing close to walls or having a sparse point cloud, one might render points that are supposedly concealed by others, which, in the end, creates discrepancies with the real image once again.

For RGBDScenes, ETH3D, and Tanks and Temples datasets, color is different from reflectivity as it does not correspond to real frequency information associated with a given light ray/point. Instead, color is captured by a separate camera sensor and synchronized in post-processing with the 3D data and precomputed calibration. This implies that some error in color projection can be caused by bad sensor registration, which is precisely the problem we are aiming to solve. For our current resolution of $1024$ px height, however, this error seems to be negligible (or less than a pixel), but it will surely appear at finer resolutions. 

Finally, our manual dataset and evaluation present their limitations. On the one hand, as we saw for the scale estimation method, pixel error is entirely dependent on the chosen point's depth: because of parallax, farther points will be less displaced than closer points for a fixed translation. This means that pixel error is simply a function of the depth distribution, which can be manipulated by selecting far points only. During our data acquisition, we ensured not to fall into this loophole by selecting evenly distributed points in depth. Second, camera estimation through point-pixel selection is itself a biased process toward certain kinds of rigid transforms. Indeed, because of the uncertainty matrix $S$ defined in~\cref{eq:pixel}, some transform directions have less observability than others (e.g., the $z$ axis in camera coordinates), which would not occur in multi-sensor setups with LiDAR-odometry registration. Third, due to uncertainty in selecting "the right pixel" for our manual registration, we were unable to achieve sub-pixel registration on our three manual sets. In \cref{tab:reproj} of the Supplementary, we observe final reprojection errors of $2$px for Tanks and Temples and $3.5$px for nuScenes on average. Additionally, the $1$px threshold column in the Tanks and Temples ablation in \cref{tab:ablation} displays a complete column of $0\%$, while our method achieves subpixel registration on the similar ETH3D set. This evidence suggests that, with ReFineR, we reached the limits of manual registration, and that the true capabilities of our method are probably slightly better at the pixel level than the results we obtained from manual registration.

\section{Conclusion}
Through the paradigm of analysis-by-synthesis, we present ReFineR: a generalist, differentiable, and training-free approach for fine-level image-to-point cloud registration. Combined with MatchAnything, ReFineR achieves state-of-the-art results on heterogeneous image and point cloud data and is capable of fine registration even at large camera perturbations up to $100$px. Furthermore, we have shown that thanks to a first-order analysis of the pinhole camera model, it is possible to remove the point cloud scale ambiguity from the optimization parameters, creating a unified method for both indoor and outdoor scenes.


\bibliography{egbib}
\clearpage

\appendix
\section{Theory and implementation}
\subsection{Pinhole camera model}
The pinhole camera model computes, for each point ${X\in\mathbb{R}^3}$ in space, the projection ${x = \pi(X)}$ onto the camera plane, in pixel units. Each camera is defined by its respective intrinsics and extrinsic parameters ${K_i\in\mathbb{R}^{3\times 3}}$ and ${K_e\in SE(3)}$ such that:
\begin{equation}
     \pi(X) = \varphi(K_i \varphi\left(K_e \varphi^{-1}(X)\right)) \label{eq:pinhole}
\end{equation}
with $\varphi$ and $\varphi^{-1}$ defined by 
\begin{align}
    \varphi(x_1, \dots, x_n) &= \left( x_1 / x_n, \dots, x_{n-1} / x_n\right) \\
    \varphi^{-1}(x_1, \dots, x_{n-1}) &= \left(x_1, \dots, x_{n-1}, 1\right)
\end{align}

\subsection{Reprojection error asymptotic development}
\label{app:parallax}
Writing ${K_i = \begin{pmatrix}
    f_x & 0 & c_x \\ 0 & f_y & c_y \\ 0 & 0 & 1
\end{pmatrix}}$ and ${A(\tau, X) = \text{exp}(\theta_\times)X + V(\theta)\rho = \begin{pmatrix}X_X & X_Y & X_Z \end{pmatrix}^T}$, with 
\begin{equation}{V(\theta) = I_3 + \frac{1-\cos(\lVert\theta\rVert)}{\lVert \theta\rVert^2}\theta_\times + \frac{\lVert\theta\rVert - \sin(\lVert\theta\rVert)}{\lVert\theta\rVert^3}\theta_\times^2}
\end{equation} and $._\times$ the antisymmetric operator, we then have from \cref{eq:pinhole}:
\begin{align}
    \pi(\tau, X) &= \varphi(K_i A(\tau, X))\\
    &= \begin{pmatrix}
        f_x\frac{X_X}{X_Z} + c_x \\
        f_y\frac{X_Y}{X_Z} + c_y
    \end{pmatrix}
\end{align}
Therefore
\begin{align}
    \!\!\frac{\mathcal{D}\pi}{\mathcal{D}\tau}(0, X) &= \frac{\partial \varphi}{\partial X}(K_i X)K_i\frac{\mathcal{D}A}{\mathcal{D}\tau}(0, X) \\
    &= \begin{pmatrix}
        \frac{1}{X_Z} & 0 & -\frac{f_x X_X + c_x X_Z}{X_Z^2} \\
        0 & \frac{1}{X_Z} & -\frac{f_y X_Y + c_y X_Z}{X_Z^2}
    \end{pmatrix}
    \begin{pmatrix}
        f_x & 0   & c_x \\
        0   & f_y & c_y \\
        0   & 0   & 1
    \end{pmatrix}
    \begin{pmatrix}
         0 & X_Z  & -X_Y & 1 & 0 & 0 \\
        -X_Z & 0  & X_X  & 0 & 1 & 0 \\
        X_Y  & -X_X & 0  & 0 & 0 & 1 
    \end{pmatrix} \\
    &= \begin{pmatrix}
    \frac{f_x}{X_Z} & 0 & -\frac{f_x X_X}{X_Z^2} \\
    0 & \frac{f_y}{X_Z} & -\frac{f_y X_Y}{X_Z^2}
    \end{pmatrix} 
    \begin{pmatrix}
         0 & X_Z  & -X_Y & 1 & 0 & 0 \\
        -X_Z & 0  & X_X  & 0 & 1 & 0 \\
        X_Y  & -X_X & 0  & 0 & 0 & 1 
    \end{pmatrix} \\
    &= \begin{pmatrix}
        -\frac{x_cy_c}{f_y} &\frac{x_c^2 + f_x^2}{f_x} &-\frac{f_xyc}{f_y} &\frac{f_x}{X_Z} &0 &-\frac{x_c}{X_Z} \\
        -\frac{y_c^2 + f_y^2}{f_y} &\frac{x_cy_c}{f_x} &\frac{f_yx_c}{f_x} &0 &\frac{f_y}{X_Z} &-\frac{y_c}{X_Z}
    \end{pmatrix}
\end{align}
with $(x_c, y_c) = (f_x X_X / X_Z, f_y X_Y / X_Z)$
\subsection{Scale estimation}
\label{app:scale}
\begin{figure}[tb]
    \centering
    \includegraphics[width=0.9\linewidth]{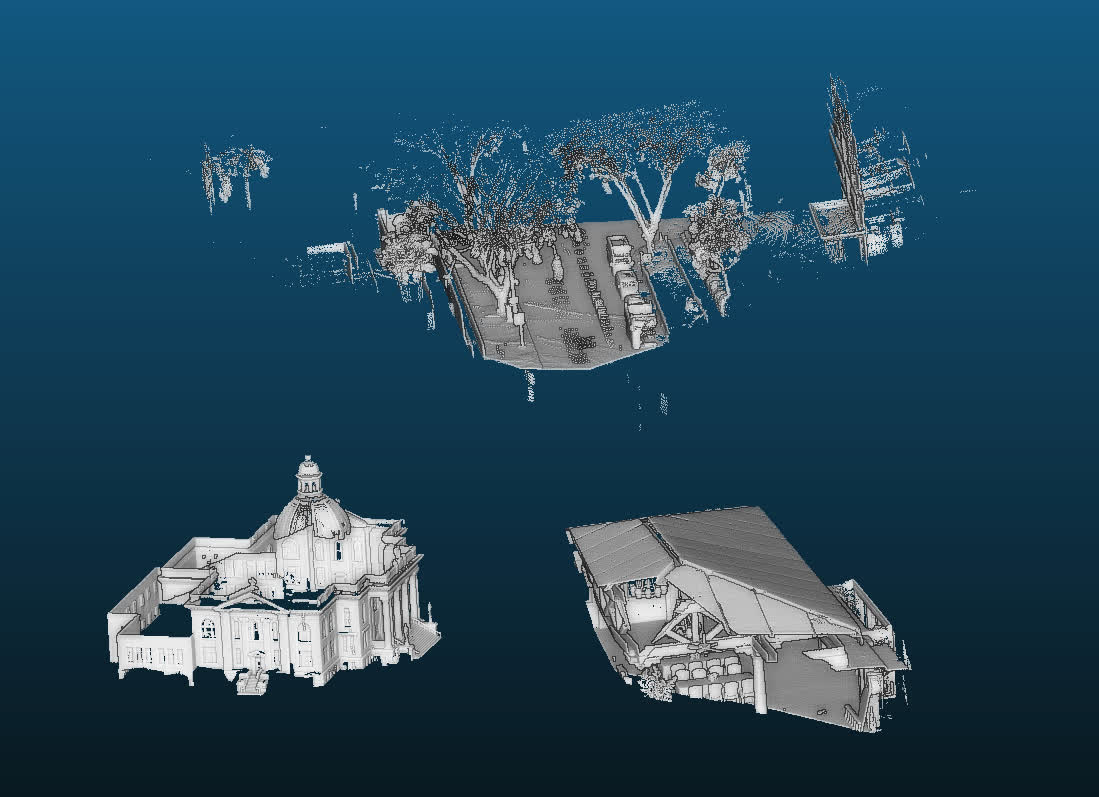}
    \caption{Three scenes from heterogeneous sources (nuScenes and Tanks and Temples), transformed to a unified scale using the method of \cref{subsec:rescale}.}
    \label{fig:rescale}
\end{figure}

Writing ${X = \alpha X'}$ and 
\begin{align}
J(X)&=  \frac{\mathcal{D}\pi}{\mathcal{D}\tau}(0, X)\\
&= \begin{pmatrix}
    J_r(X) & J_{t}(X)
\end{pmatrix} \\
&= \begin{pmatrix}
    J_r(X) & \frac{1}{\alpha}J_{t'}(X)
\end{pmatrix}
\end{align} then 
\begin{align}
    \!\!S &= \frac{1}{N}\sum\limits_{i=1}^NJ(X_i)^TJ(X_i) \\
    &= \frac{1}{N}\sum\limits_{i=1}^N\begin{pmatrix}
        J_r(X_i)^TJ_r(X_i) &\frac{1}{\alpha}J_r(X_i)^TJ_{t'}(X_i) \\
        \frac{1}{\alpha}J_{t'}(X_i)^TJ_r(X_i) &\frac{1}{\alpha^2}J_{t'}(X_i)^TJ_{t'}(X_i)
    \end{pmatrix} \\
    &= \begin{pmatrix}
        S_r & \frac{1}{\alpha}S_{rt'} \\
        \frac{1}{\alpha}S_{rt'}^T & \frac{1}{\alpha^2}S_{t'}
    \end{pmatrix} \\
    &= \begin{pmatrix}
        S_r & S_{rt} \\
        S_{rt}^T & S_{t}
    \end{pmatrix} 
\end{align}
The eigenvalues between $S_{t}$ and $S_{t'}$ are related by 
\begin{equation}
    \lambda_{t', i} = \alpha^2\lambda_{t,i}
\end{equation}
Therefore, equalizing the Hilbert-Schmidt norms:
\begin{align}
    \sum\limits_{i=1}^3\lambda_{t',i}^2 &= \sum\limits_{i=1}^3\lambda_{r, i}^2 \\
    \Leftrightarrow \alpha^4\sum\limits_{i=1}^3\lambda_{t,i}^2 &= \sum\limits_{i=1}^3\lambda_{r, i}^2\\
    \Leftrightarrow \alpha &= \left(\nicefrac{\sum\limits_{i=1}^3\lambda_{r, i}^2}{\sum\limits_{i=1}^3\lambda_{t,i}^2}\right)^{1/4}
\end{align}
\cref{fig:rescale} shows examples of 3 point clouds rescaled using the above equations for $\alpha$.

\subsection{Initialization at a target pixel error}
\label{app:init}
We explain here how to perturb the ground truth camera pose to obtain a fixed pixel error.
Starting back from \cref{eq:pixel} around the identity, and using the first-order approximation, we wish to find a transform $\delta\tau$ such that for a given pixel error $\Delta$ we have:
\begin{equation}
    \delta\tau ^TS\delta\tau = \Delta^2
\end{equation}
which by diagonalizing $S$ reads:
\begin{align}
    \Delta^2 &= (P\delta\tau)^TD(P\delta\tau) \\
             &= \sum\limits_{i=1}^6\lambda_i (P\delta\tau)_i^2
\end{align}
with $P$ an orthogonal matrix and ${D = \text{diag}(\lambda_1,\dots\lambda_6)}$.
Therefore, such a $\delta\tau$ can be found by randomly selecting a unit vector $\tau = (\tau_1,\dots, \tau_6)^T$ and computing:
\begin{equation}
    \delta\tau = \Delta P^T\begin{pmatrix}
        \frac{\tau_1}{\sqrt{\lambda_1}}\\
        \vdots\\
        \frac{\tau_6}{\sqrt{\lambda_6}}
    \end{pmatrix}
\end{equation} 
Note that this formulation remains an approximation of the pixel error due to the 1st order asymptotic development, which, in practice, might not produce exactly a pixel error of magnitude $\Delta$ for large perturbations.

\subsection{Implementation of differentiable HOG features}
\label{app:hog}
As a comparison to Guislain \etal~\cite{guislainFineScaleImage2017}, who proposed using HOG features~\cite{dalalHistogramsOrientedGradients2005}, we implement a differentiable version of this metric to fit our pipeline. In four steps: \\

\textbf{1. Gradient computation:} Starting from an image $(I_p)_p$ indexed by its pixel $p$, we compute the gradient magnitude and orientation images ${(G_p, \Theta_p)_p}$ using Sobel filters. \\

\textbf{2. Cell histogram:} A cell $C$ is defined as a set of locally adjacent pixels: ${C = \{p^C_1, \dots, p^C_{|C|}\}}$. In practice, we used a kernel with cell size ${k_C = 4}$, with padding and stride hyperparameters. The cell size is then $|C| = k_C^2$. For each cell, a differentiable histogram $F^C_m$ of unsigned orientation is computed with $M$ evenly spaced bin centers ${(m_i)_{1\leq i\leq M}} \in [0, \pi]^M$:
\begin{align}
    F^C_i &= \sum\limits_{p\in C}G_pf_\sigma(\Theta_p, m_i)
\end{align}
with 
$f_\sigma(\theta, m)$ the Gaussian's density with mean $m$ and variance $\sigma^2$, and $\sigma$ fixed to ${0.5*(m_2 - m_1)}$ \\

\textbf{3. Block normalization: } Similarly, a block $B$ is defined as a set of adjacent cells: ${B = \{C_1^B\dots C_{|B|}^B\}}$, and the block size is defined using block kernel size $|B| = k_B^2$. We used $k_B=4$. The normalized features $\left(F^{B, C}_i\right)_{B, C, i}$ are computed as:
\begin{align}
    N^B &= \sqrt{\frac{1}{M|B|}\sum\limits_{C\in B, 1\leq i\leq M}\left(F_i^C\right)^2} & 
    F^{B, C}_i &= \frac{F^C_i}{N_B}
\end{align} \\

\textbf{4. Loss function: } Given two images $I$ and $I'$, and the features $\left(F^{B, C}_i\right)_{B, C, i}$, $\left(F'^{B, C}_i\right)_{B, C, i}$, the full feature loss is:
\begin{equation}
    HOG(I, I') =\frac{\sum\limits_{B, C, i}\left(F^{B, C}_i - F'^{B, C}_i\right)^2}{|B||C|M}
\end{equation}

\section{Dataset preparation}
\label{app:dataset}

\subsection{ETH3D}
From the ETH3D dataset \cite{schoeps2017cvpr}, we reconstructed the global point clouds using ground truth alignment files and subsampled them using a $5$ mm voxel grid. Then the normals' orientations were computed using CloudCompare. Fifty images spanning ten scenes were then randomly chosen to form our dataset, along with their ground truth camera positions and intrinsics. Thanks to its pixel-level refinement scheme, this dataset presents sub-pixel camera registration for resolutions up to 6048 $\times$ 4032. Additionally, each point cloud was acquired using a Faro Focus X 330, providing centimeter-level point precision.

\subsection{Tanks and Temples}
We chose the preprocessed Church and Meetingroom LiDAR point clouds from Tanks and Temples~\cite{knapitschTanksTemples2017} and ten images from each scene. The available LiDAR point clouds were cleanly preprocessed by the original authors, and because of their high number of points, we randomly subsampled them, keeping $40\%$ of the points during our experiments. Since camera positions and intrinsics were not given by the authors, we used a manual point-pixel matching protocol to estimate the extrinsic parameters as well as the shared intrinsics and 5 distortion parameters following the OpenCV distortion model. The final reprojection errors were around $2$px on average. 

\subsection{nuScenes}
We chose four scenes (1, 3, 7, and 12) from the nuScenes~\cite{Caesar2019nuScenesAM} dataset. For each scene, the LiDAR scans were aggregated into a global point cloud using CT-ICP~\cite{dellenbachCTICPRealtimeElastic2021}, and we removed dynamic objects using the ground truth annotations and manual selection. We also applied histogram equalization on the LiDAR intensity values to enhance contrast. Five images were selected for each scene, all from the same front-right camera, taking care to choose images where the point cloud would be dense enough to fill most of the image when projected. Ground truth camera poses were similarly obtained using our manual matching.

\subsection{Livox}
We took 4 Livox scans with their associated images from Yuan \etal's work~\cite{yuanPixelLevelExtrinsicSelf2021}, consisting of outdoor scenes. Despite its relatively small size, we chose to include this set for a fair comparison with Yuan \etal's work \cite{yuanPixelLevelExtrinsicSelf2021}, whose authors built a method for the specific case of Livox sensors.

\subsection{RGBDScenes}
We selected 5 images from scenes 01, 07, 09, and 13 of the RGBDScenes dataset~\cite{Lai2014UnsupervisedFL}. Calibration was kept unchanged, and we used $(f_x, f_y, c_x, c_y)=(525, 525, 320, 240)$ as our intrinsics values. This set represents a drastic shift in point cloud quality from the other sets because of the sensor's resolution, but has the advantage of intrinsically having color intensity, which can be leveraged for our task.

\subsection{CARLA}
We used the CARLA simulator~\cite{Dosovitskiy2017CARLAAO} to generate five sequences of synthetic data. The sequences were recorded in Towns 2, 3, 4, 5, and 10. We placed a rotating LiDAR and two cameras on the main vehicle, one facing forward and the other facing right; the sensor parameters are given in \cref{tab:carla_sensors}. Parameters that are not mentioned are kept at their default values (see here\footnote{\url{https://carla.readthedocs.io/en/latest/ref_sensors/}}). We also provide for each modality the associated semantic information, which was used in \cref{app:seg_transfer} to assess a 2D to 3D feature transfer quality in regard to a perturbed camera position. We noticed that the CARLA LiDAR had inaccuracies (missing objects or buildings); therefore, we manually selected 10 images from each sequence to mitigate the discrepancies between images and point clouds. Being synthetically generated, this dataset has the benefit of having floating-point-perfect image registration.
\begin{table}[tb]
    \centering
        \begin{tabular}[width=\linewidth]{lc}
            \toprule
            Sensor / Parameter & value \\
            \midrule
            Camera \\
            \quad image\_size\_x & 1920 \\
            \quad image\_size\_y & 1080 \\
            \quad fov          &   90 \\

            \midrule
            LiDAR \\
            \quad points\_per\_second       & 500,000 \\
            \quad rotation\_frequency       &       5 \\
            \quad dropoff\_general\_rate    &       0 \\
            \quad dropoff\_intensity\_limit &       0 \\
            \quad range                     &     200 \\
            \quad channels                  &      64 \\
            \bottomrule
        \end{tabular}
    \vspace{5pt}
    \caption{CARLA sensor setup.}
    \label{tab:carla_sensors}
\end{table}

\subsection{Common preprocessing}
\begin{figure}
    \centering
    \begin{subfigure}{0.45\textwidth}
        \includegraphics[width=\textwidth]{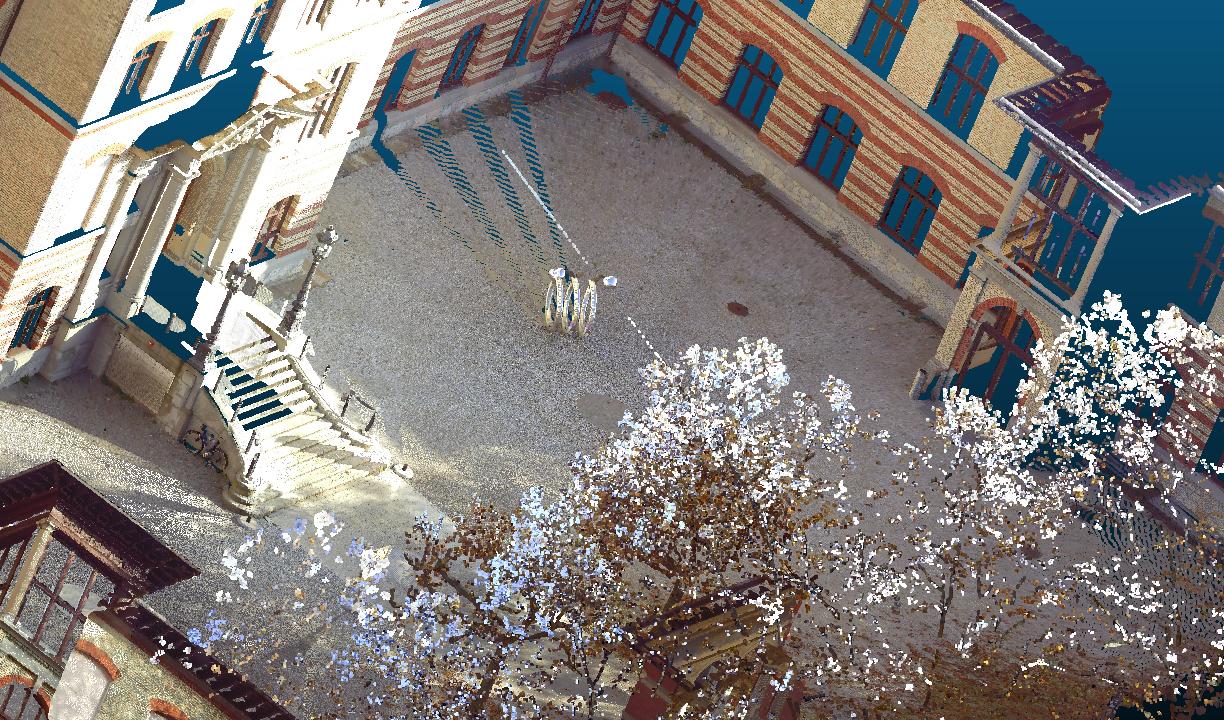}
        \caption{ETH3D}
    \end{subfigure}
    \begin{subfigure}{0.45\textwidth}
        \includegraphics[width=\textwidth]{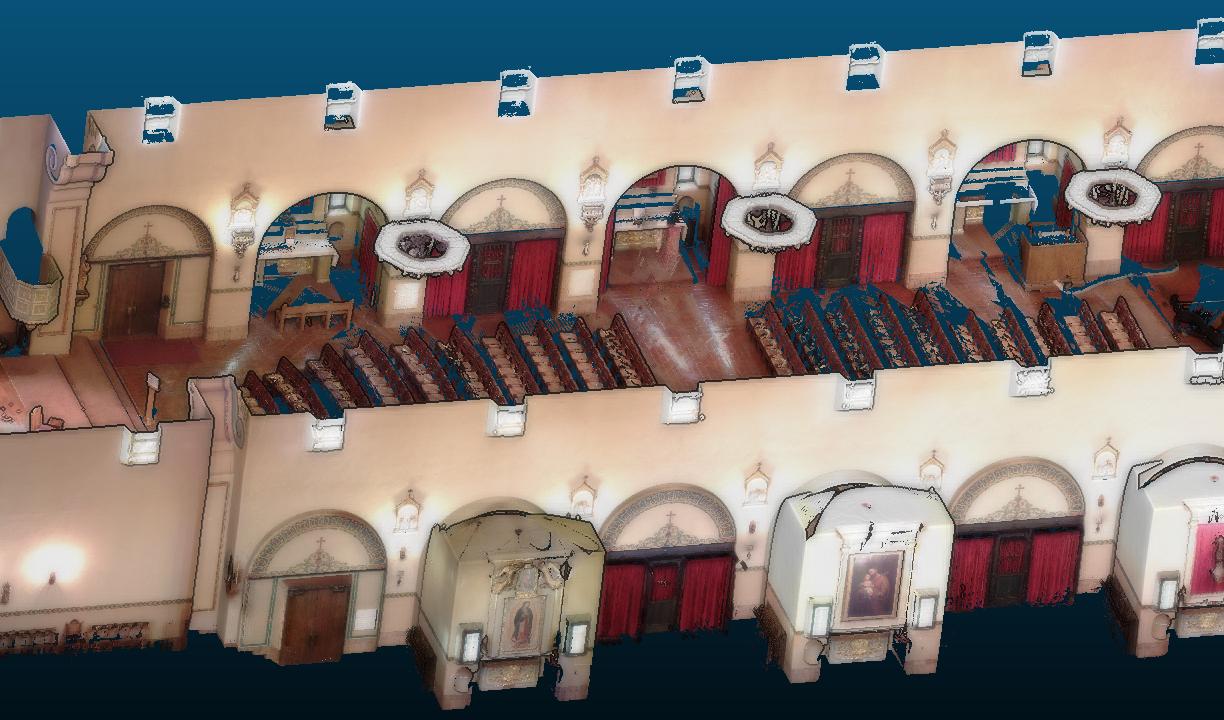}
        \caption{Tanks And Temples}
    \end{subfigure}
    \begin{subfigure}{0.45\textwidth}
        \includegraphics[width=\textwidth]{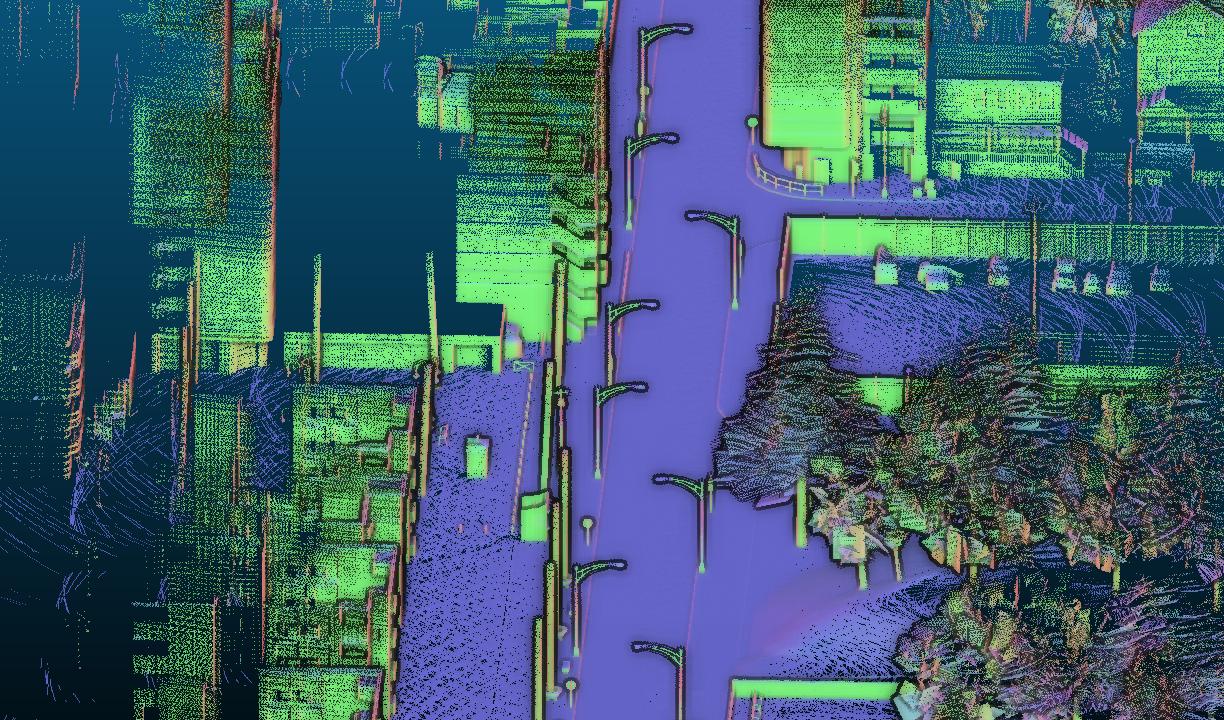}
        \caption{CARLA}
    \end{subfigure}
    \begin{subfigure}{0.45\textwidth}
        \includegraphics[width=\textwidth]{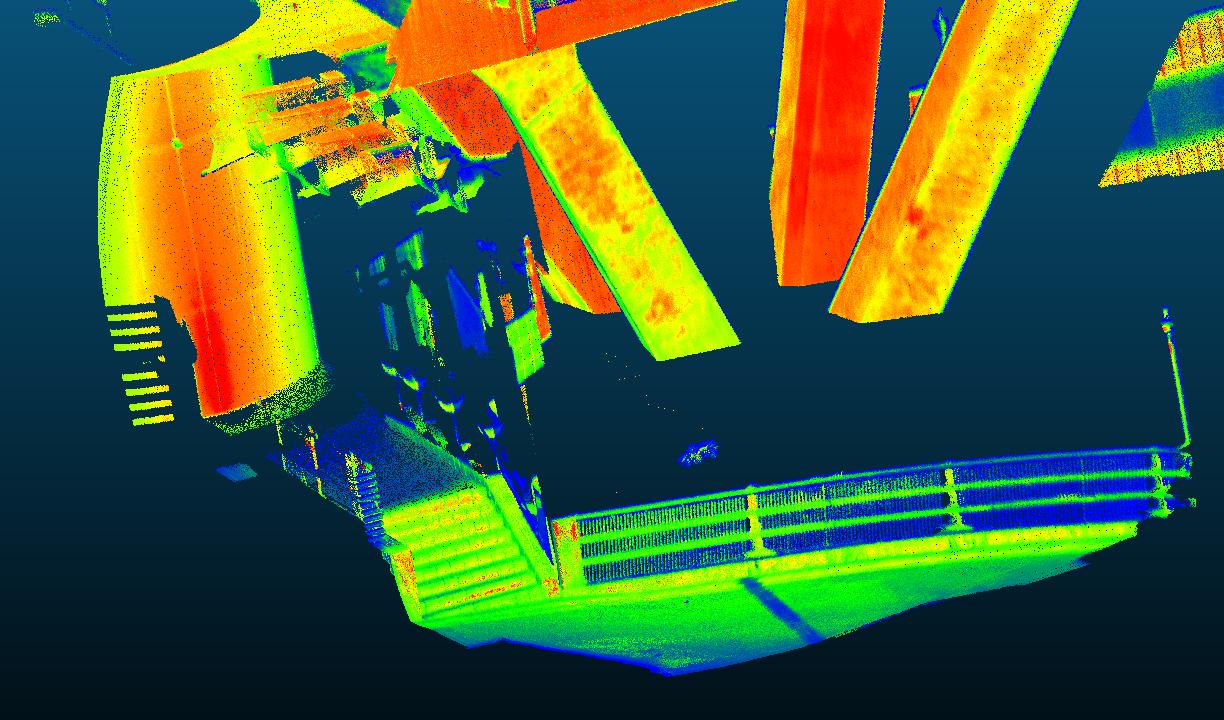}
        \caption{Livox}
    \end{subfigure}
    \begin{subfigure}{0.45\textwidth}
        \includegraphics[width=\textwidth]{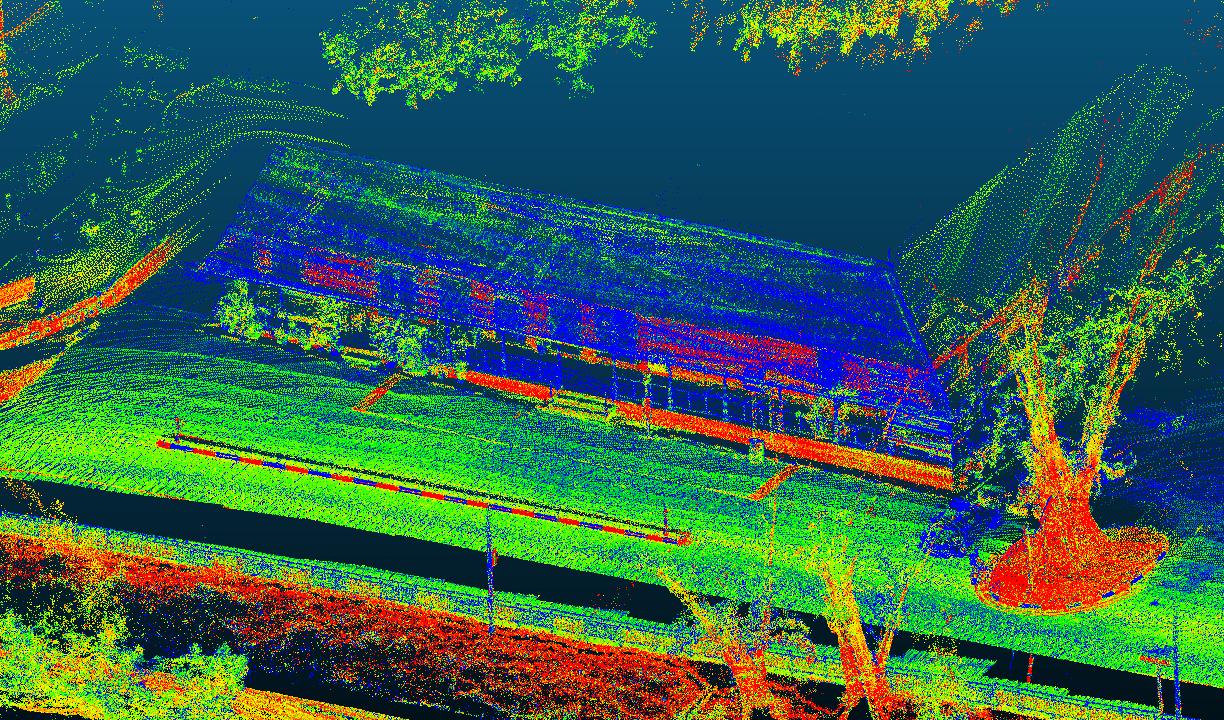}
        \caption{nuScenes}
    \end{subfigure}
    \begin{subfigure}{0.45\textwidth}
        \includegraphics[width=\textwidth]{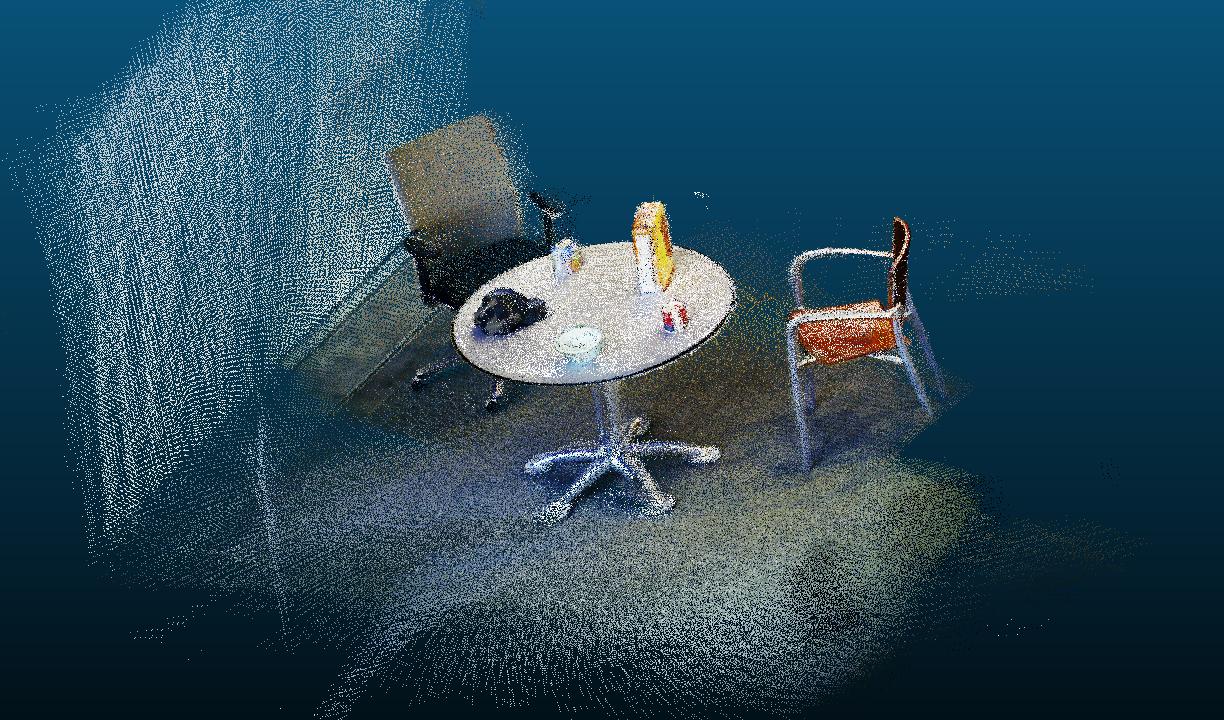}
        \caption{RGBDScenes}
    \end{subfigure}

    \caption{Example point clouds from our benchmark dataset.}
    \label{fig:exemples}
\end{figure}

\begin{figure}
    \centering
    \begin{subfigure}{0.30\textwidth}
        \includegraphics[width=\textwidth]{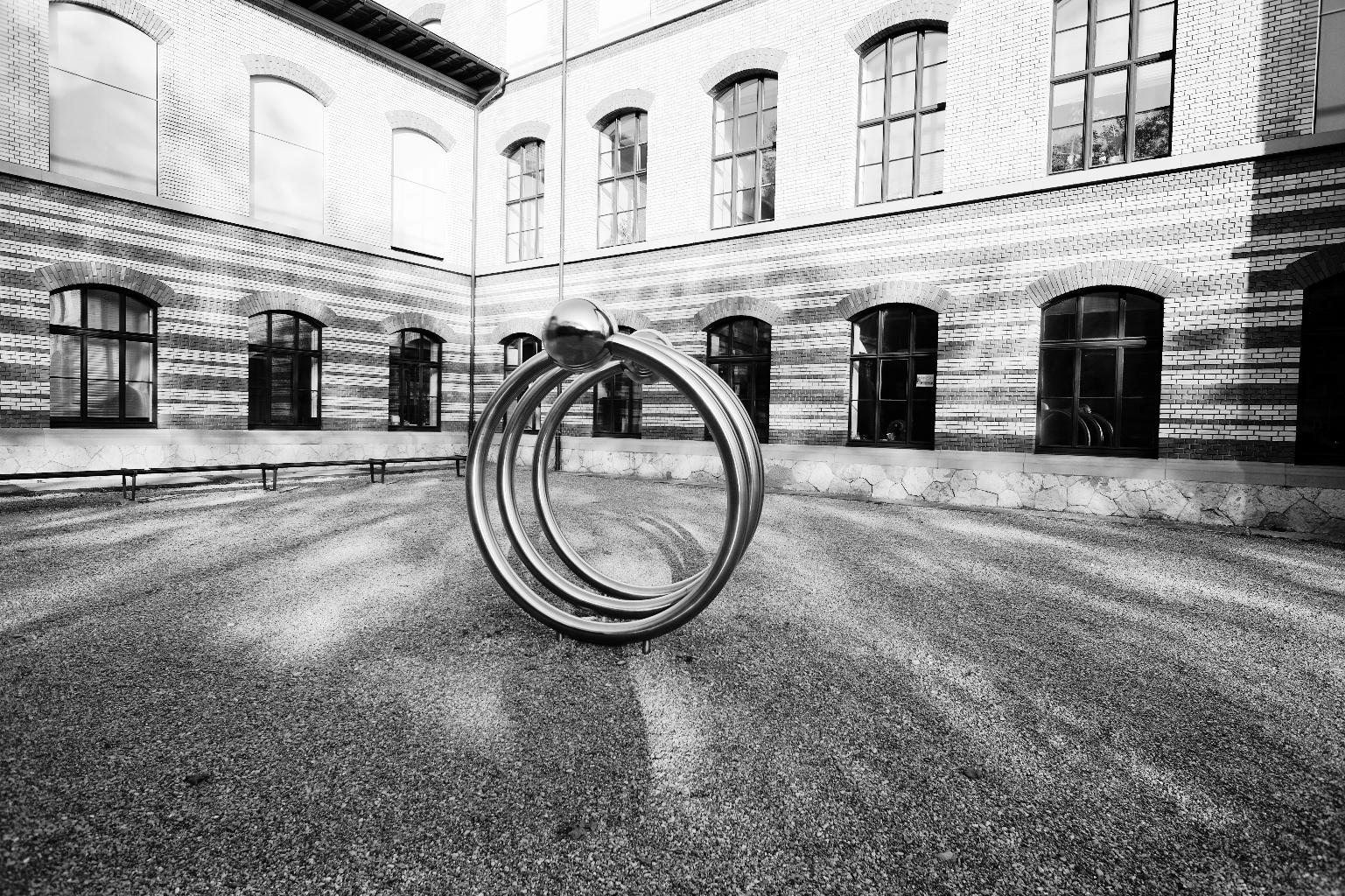}
        \caption{$i^g$}
    \end{subfigure}
    \begin{subfigure}{0.30\textwidth}
        \includegraphics[width=\textwidth]{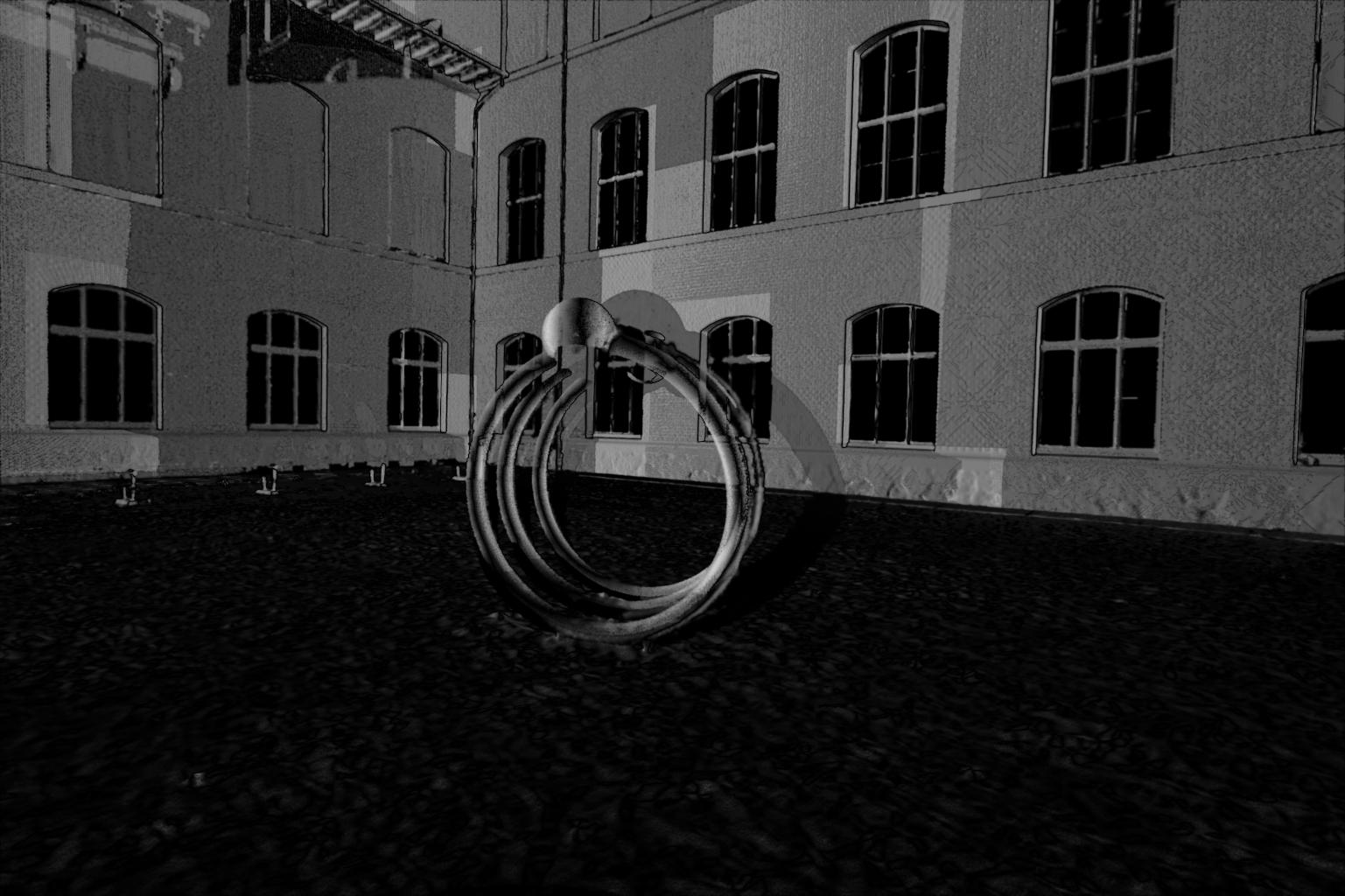}
        \caption{$p^{nx}$}
    \end{subfigure}  
    \begin{subfigure}{0.30\textwidth}
        \includegraphics[width=\textwidth]{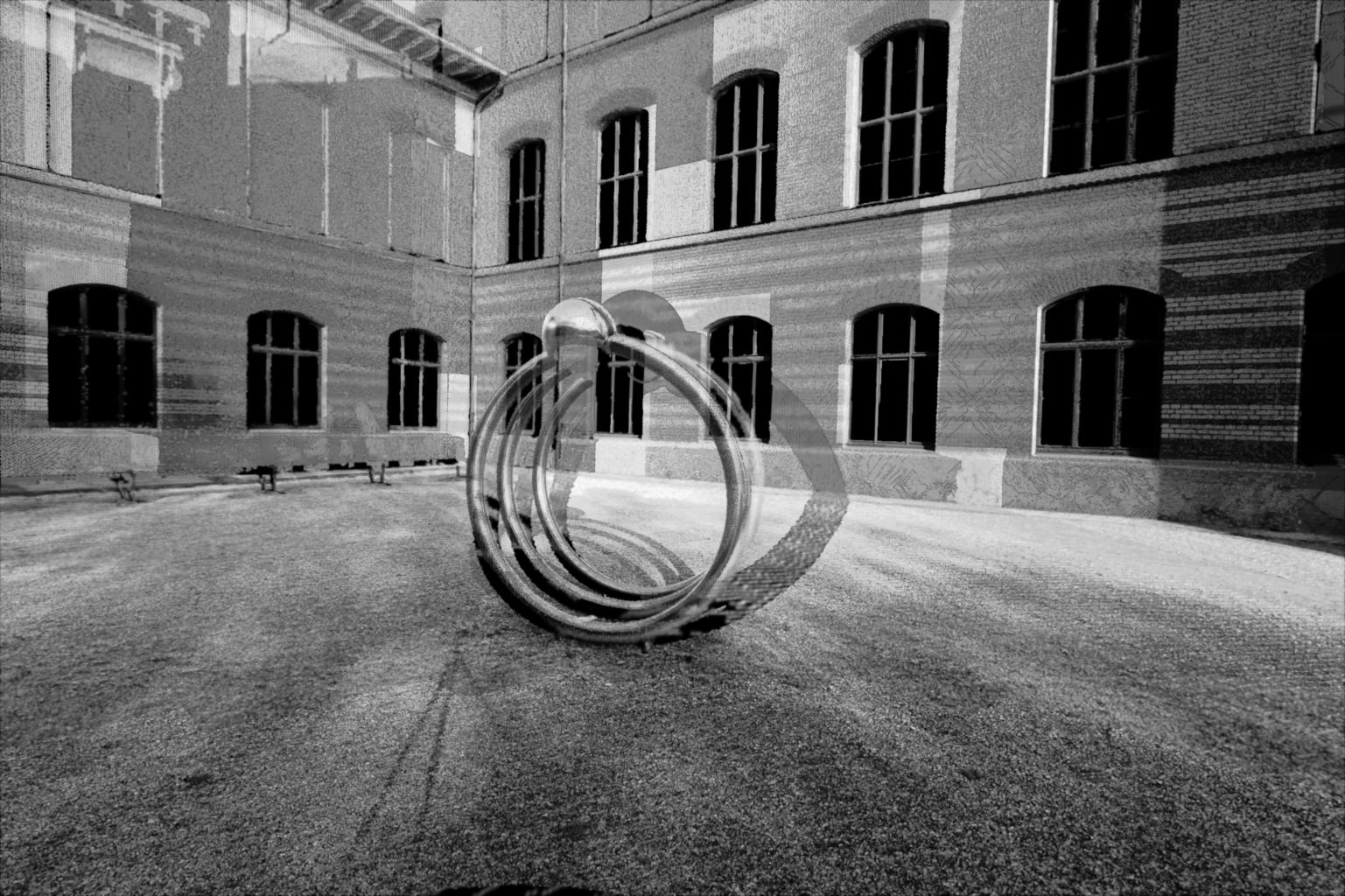}
        \caption{$p^g$}
    \end{subfigure} 

    \begin{subfigure}{0.30\textwidth}
        \includegraphics[width=\textwidth]{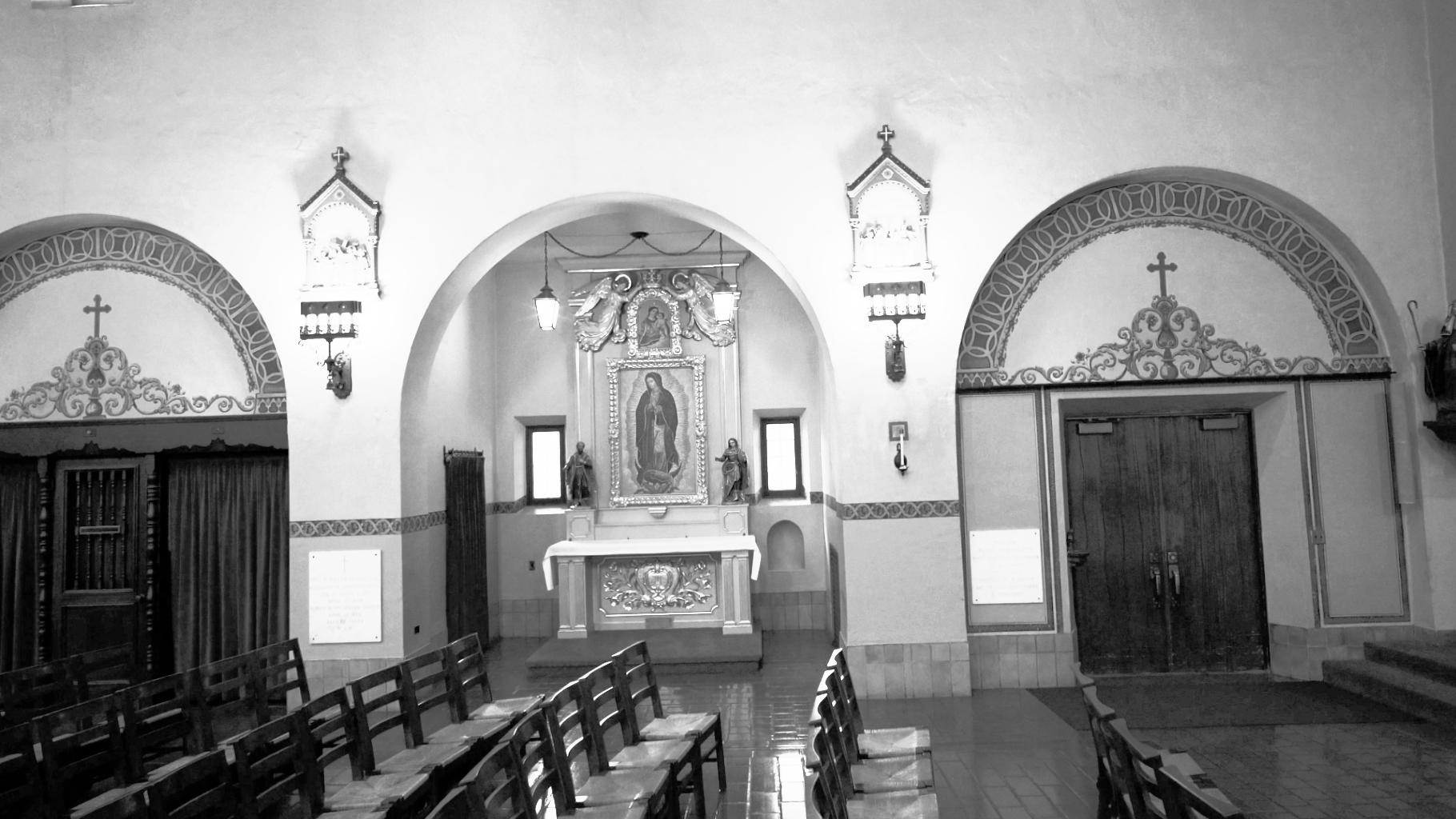}
        \caption{$i^g$}
    \end{subfigure}
    \begin{subfigure}{0.30\textwidth}
        \includegraphics[width=\textwidth]{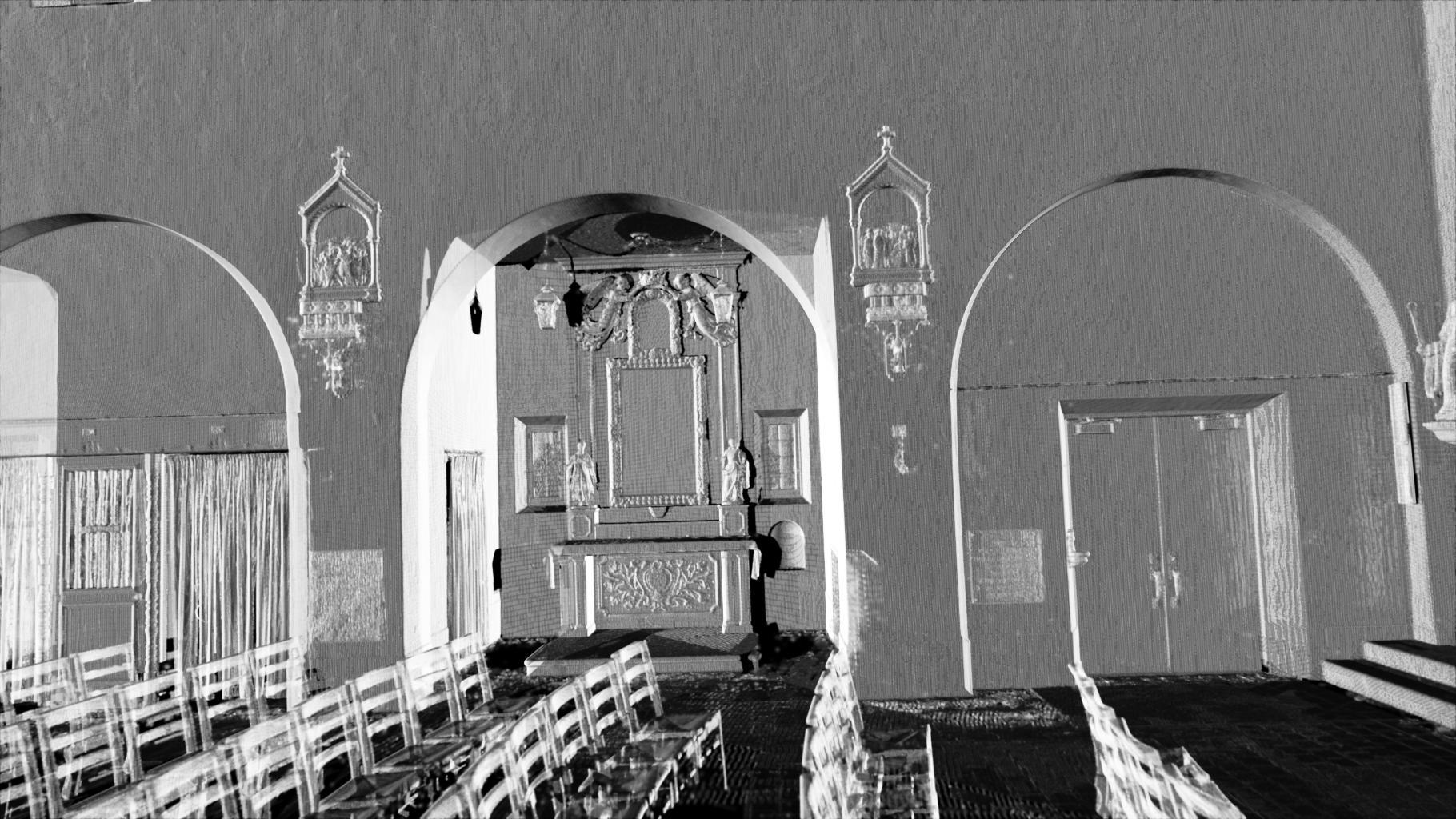}
        \caption{$p^{nx}$}
    \end{subfigure}  
    \begin{subfigure}{0.30\textwidth}
        \includegraphics[width=\textwidth]{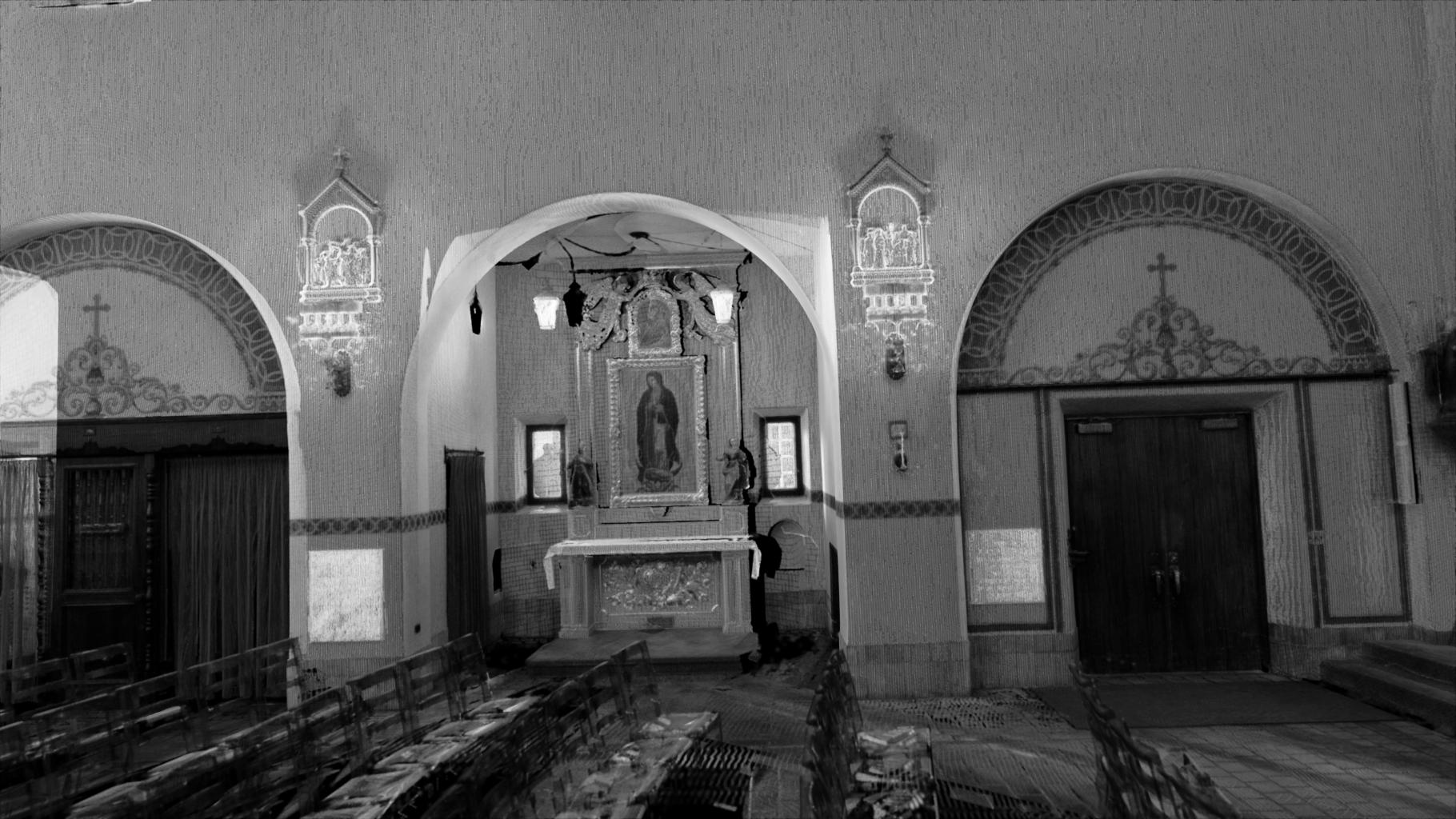}
        \caption{$p^g$}
    \end{subfigure} 

    \begin{subfigure}{0.30\textwidth}
        \includegraphics[width=\textwidth]{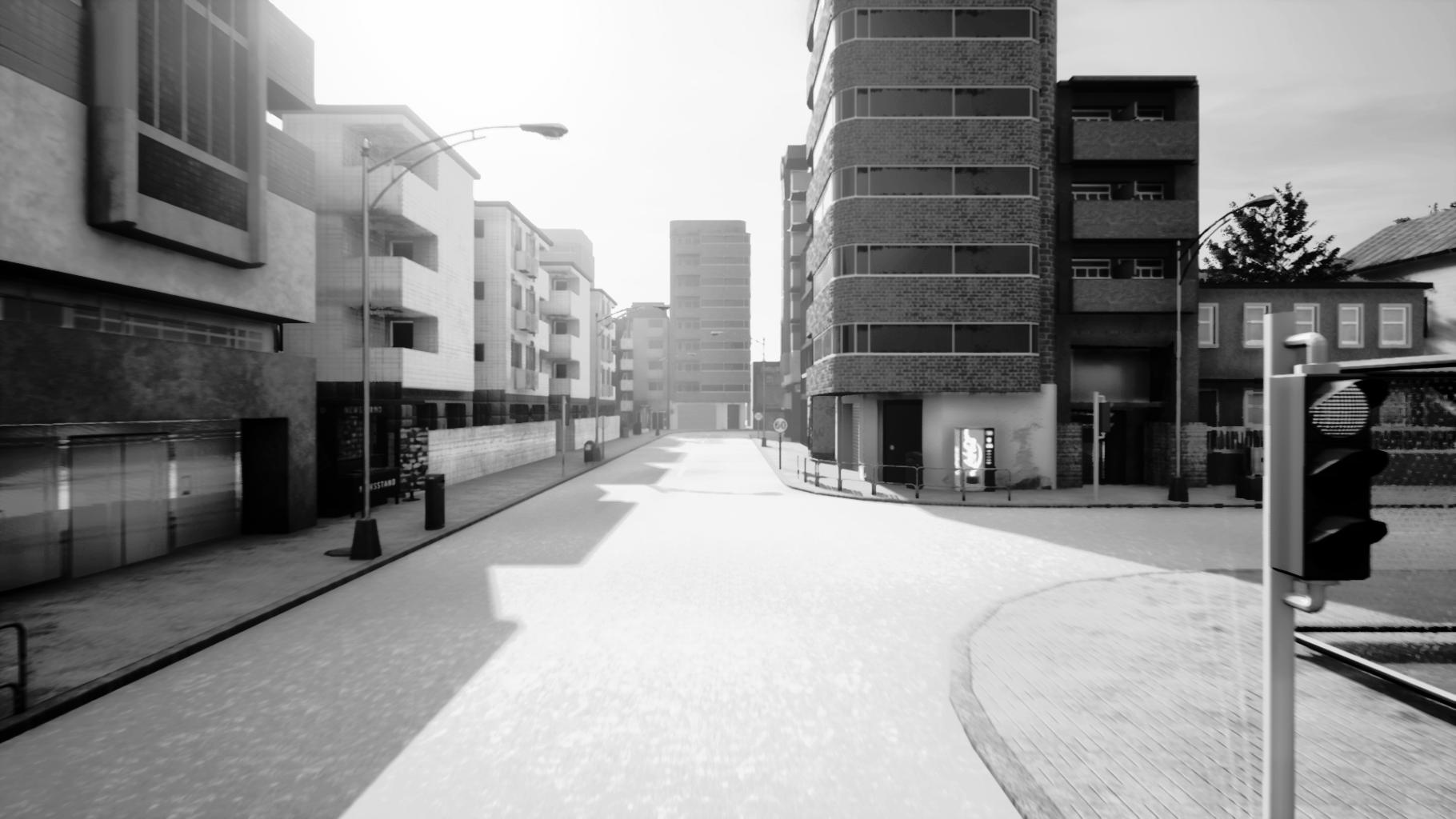}
        \caption{$i^g$}
    \end{subfigure}
    \begin{subfigure}{0.30\textwidth}
        \includegraphics[width=\textwidth]{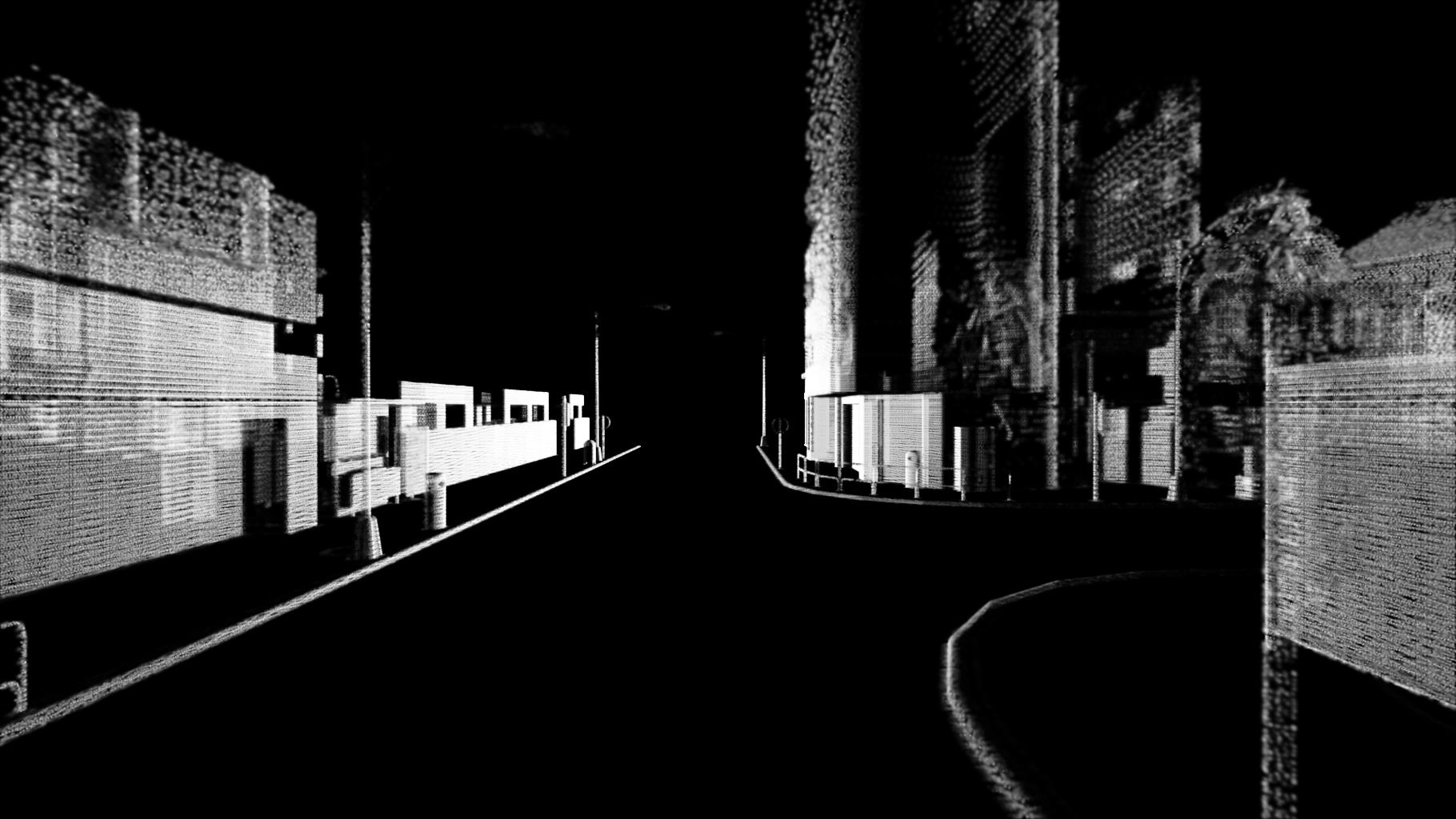}
        \caption{$p^{nx}$}
    \end{subfigure}  

    \begin{subfigure}{0.30\textwidth}
        \includegraphics[width=\textwidth]{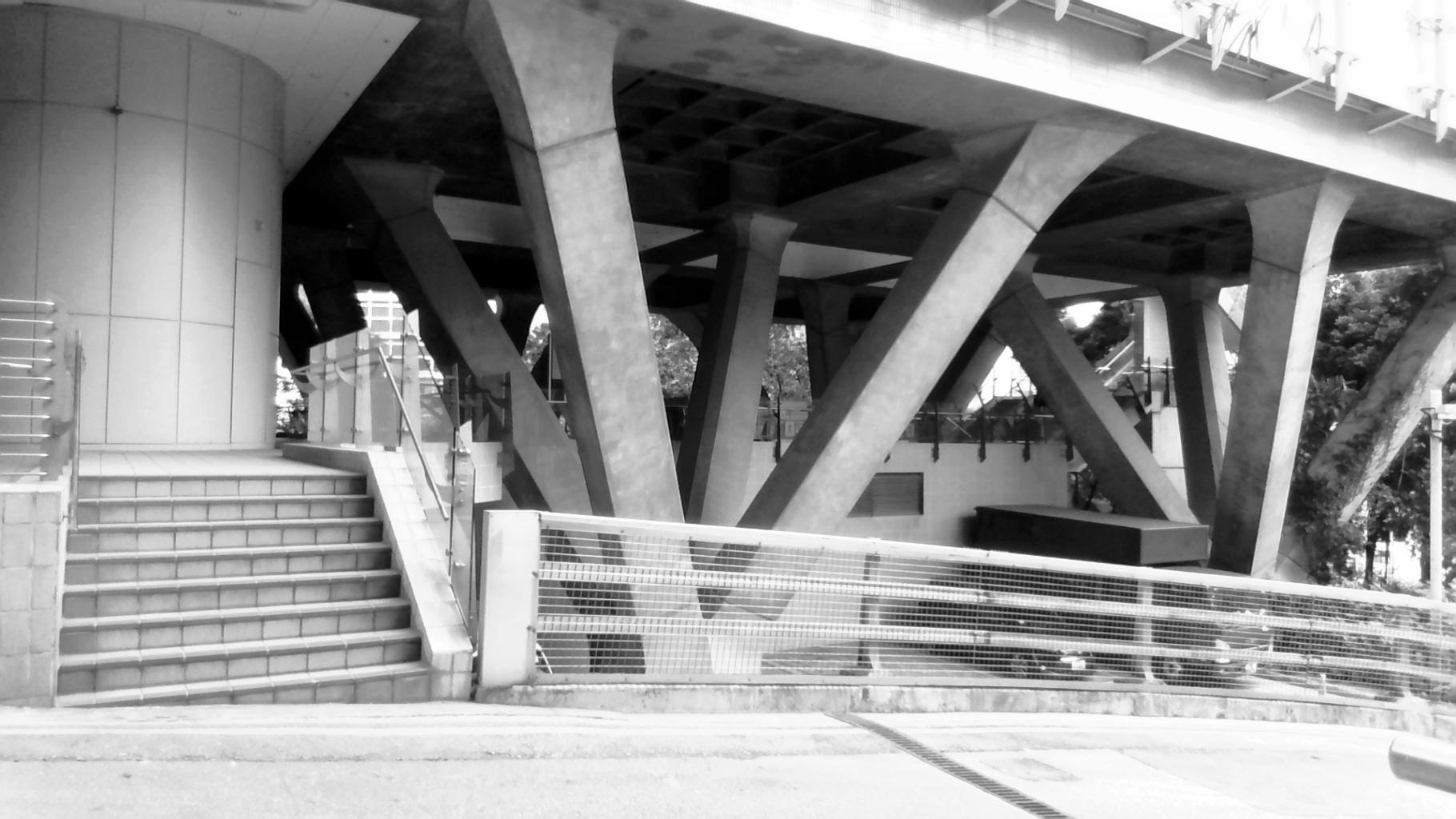}
        \caption{$i^g$}
    \end{subfigure}
    \begin{subfigure}{0.30\textwidth}
        \includegraphics[width=\textwidth]{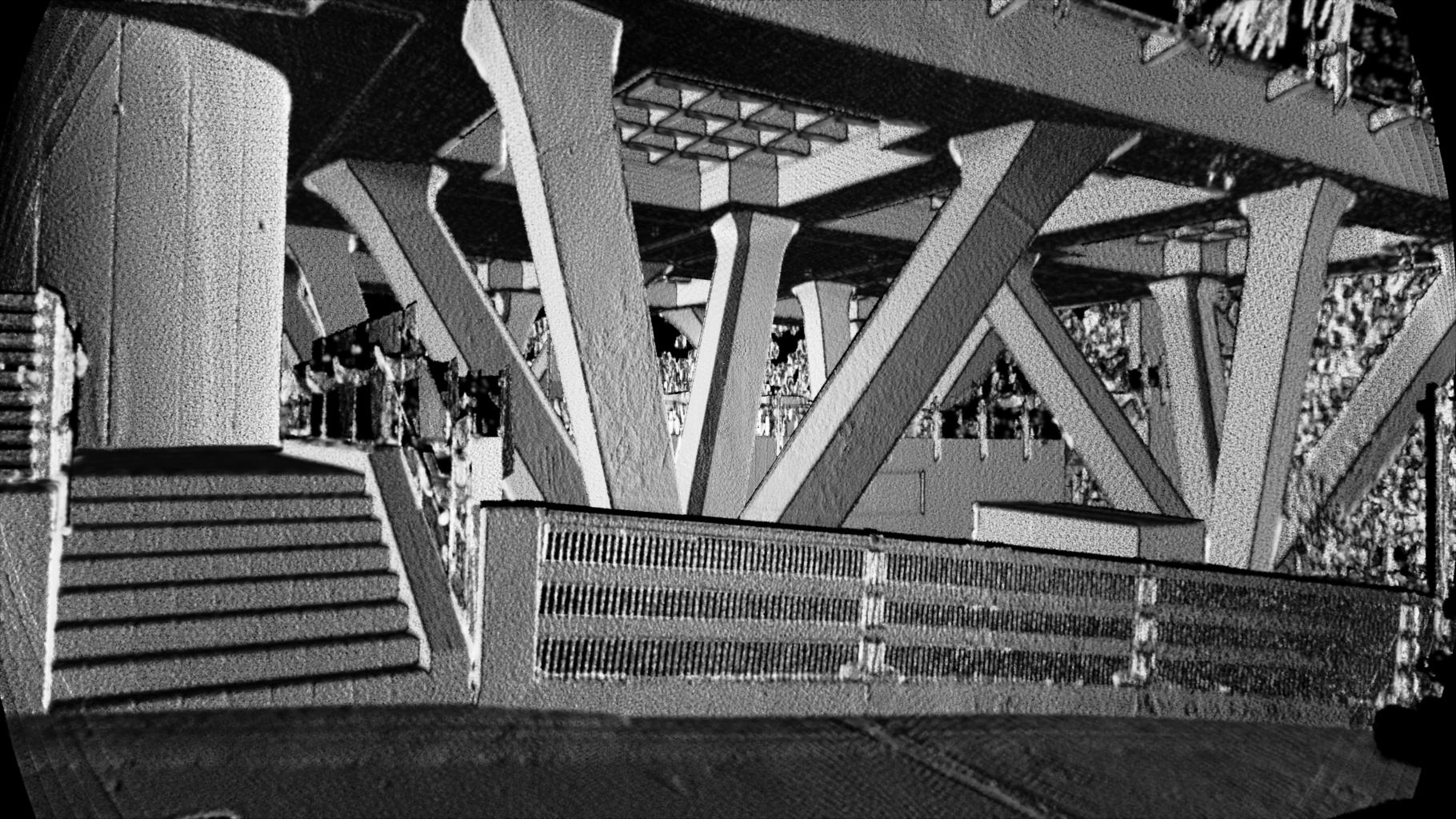}
        \caption{$p^{nx}$}
    \end{subfigure}  
    \begin{subfigure}{0.30\textwidth}
        \includegraphics[width=\textwidth]{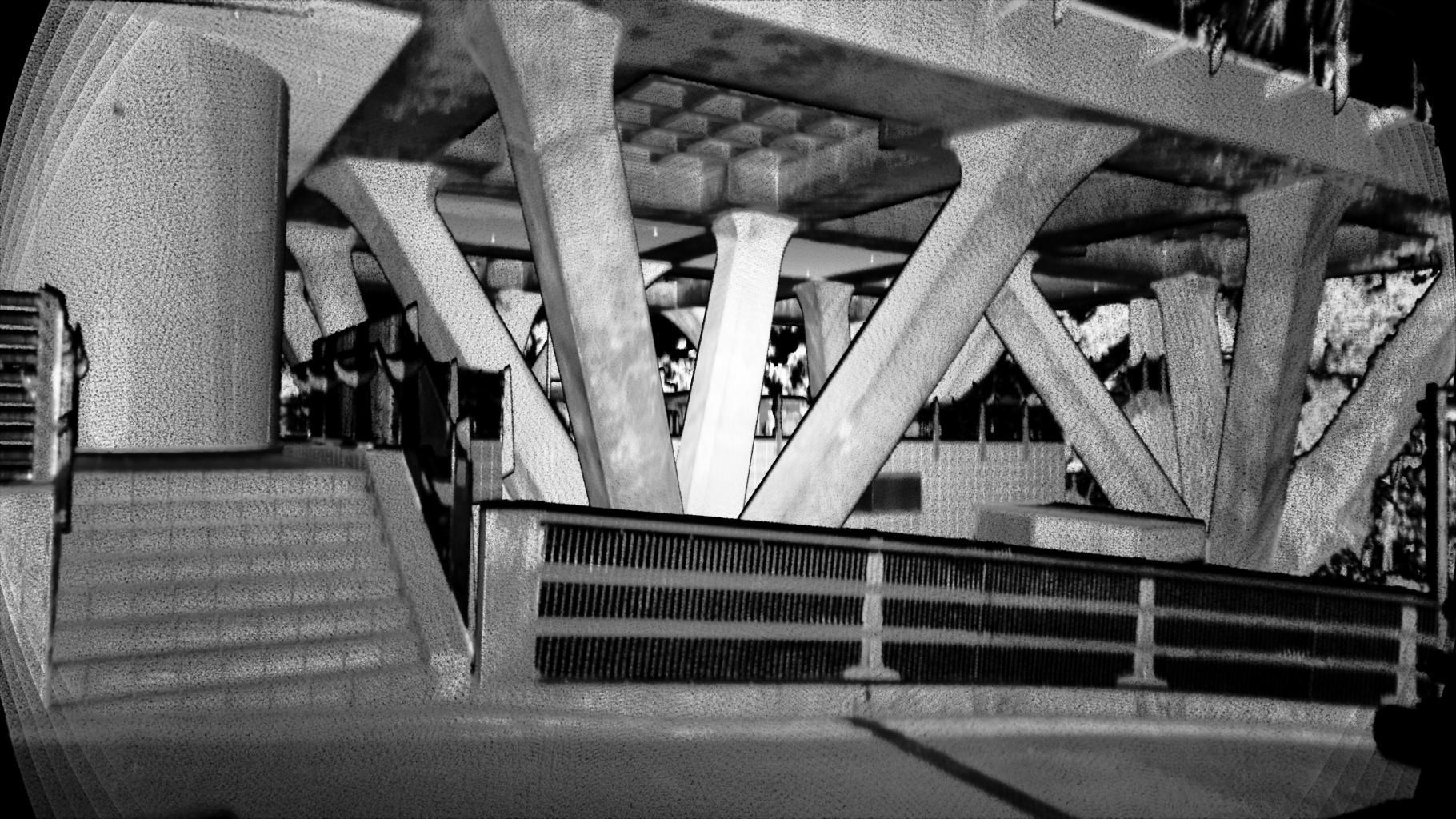}
        \caption{$p^r$}
    \end{subfigure} 

    \begin{subfigure}{0.30\textwidth}
        \includegraphics[width=\textwidth]{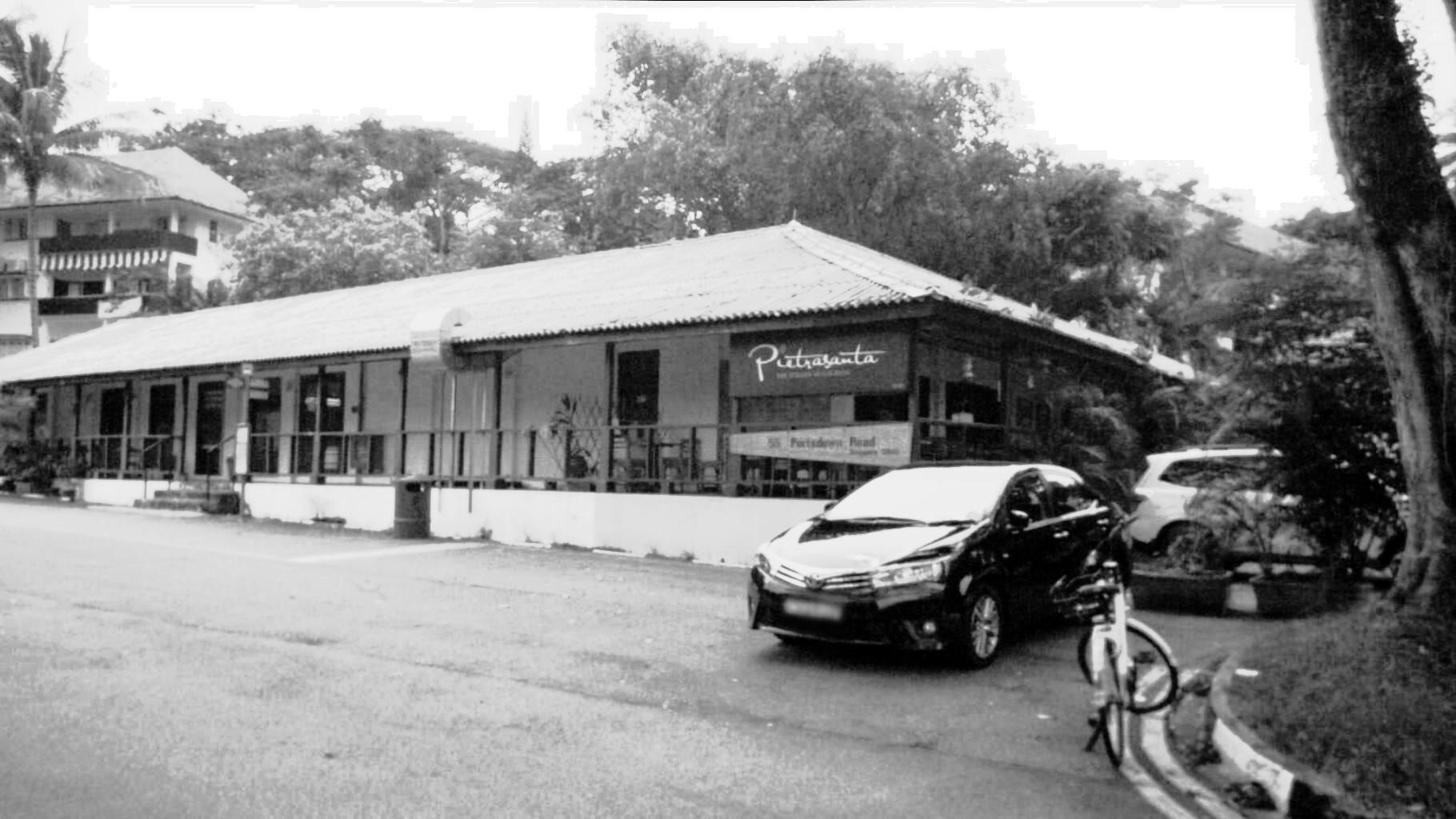}
        \caption{$i^g$}
    \end{subfigure}
    \begin{subfigure}{0.30\textwidth}
        \includegraphics[width=\textwidth]{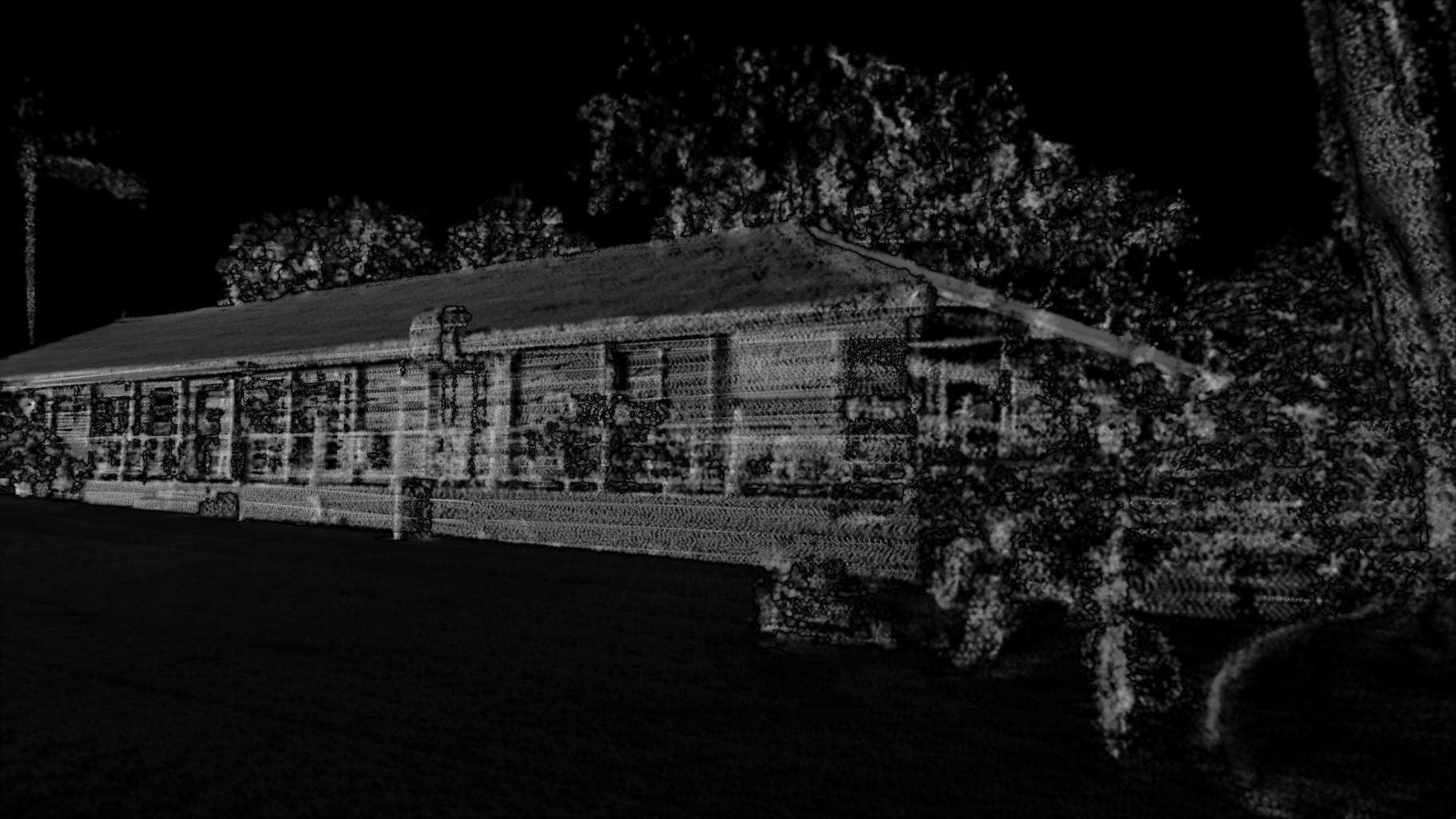}
        \caption{$p^{nx}$}
    \end{subfigure}  
    \begin{subfigure}{0.30\textwidth}
        \includegraphics[width=\textwidth]{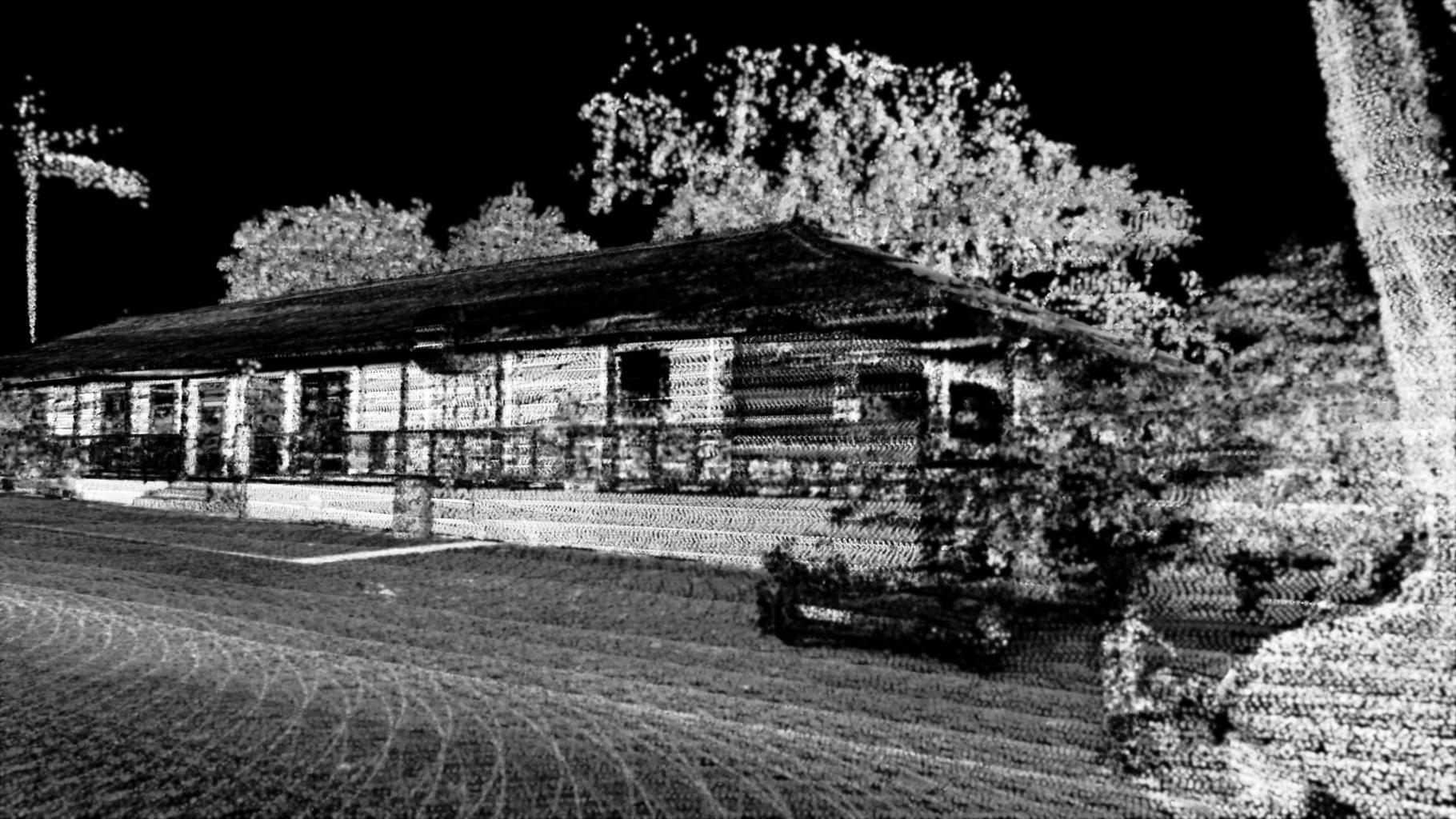}
        \caption{$p^r$}
    \end{subfigure} 

    \begin{subfigure}{0.30\textwidth}
        \includegraphics[width=\textwidth]{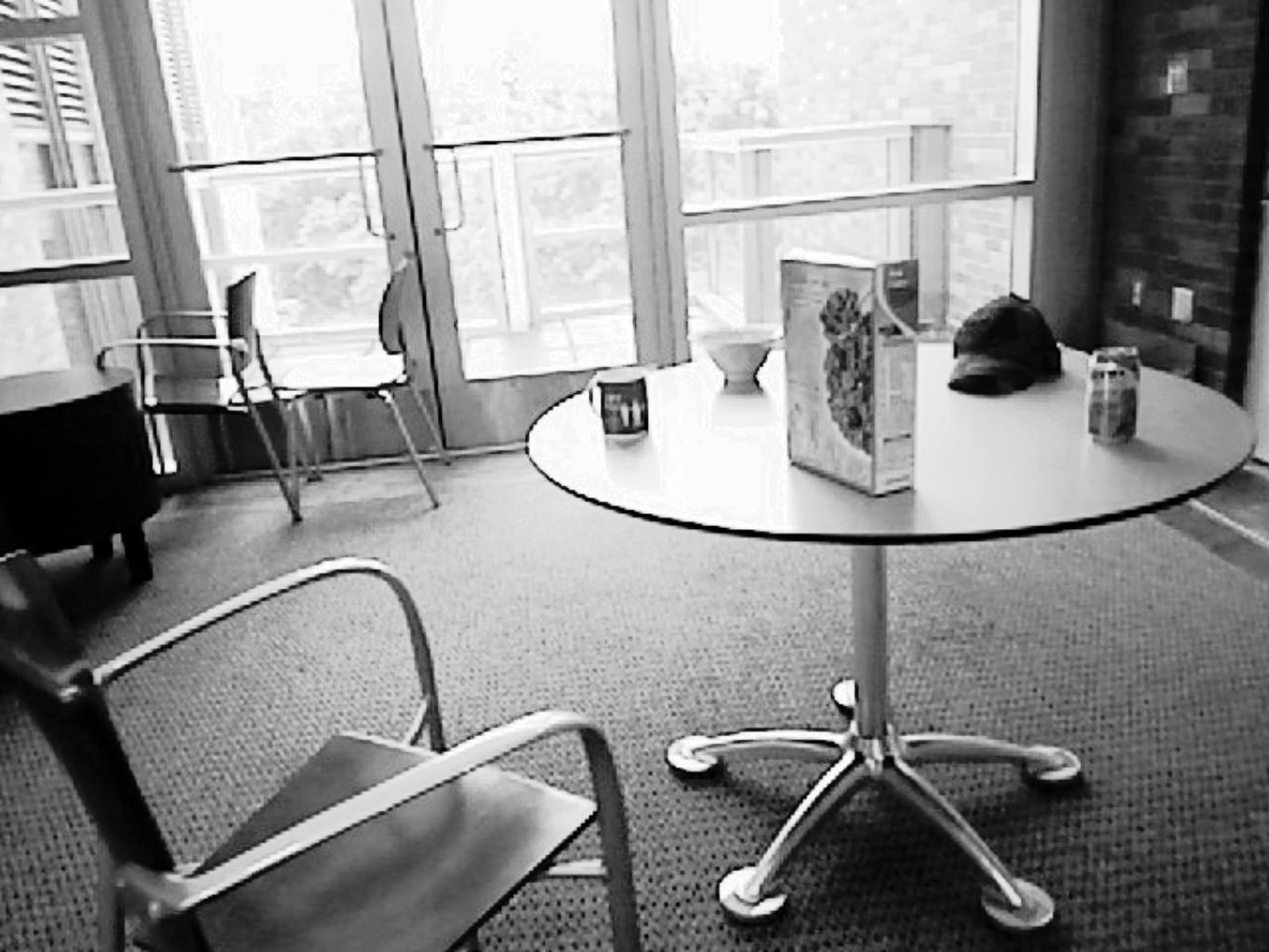}
        \caption{$i^g$}
    \end{subfigure}
    \begin{subfigure}{0.30\textwidth}
        \includegraphics[width=\textwidth]{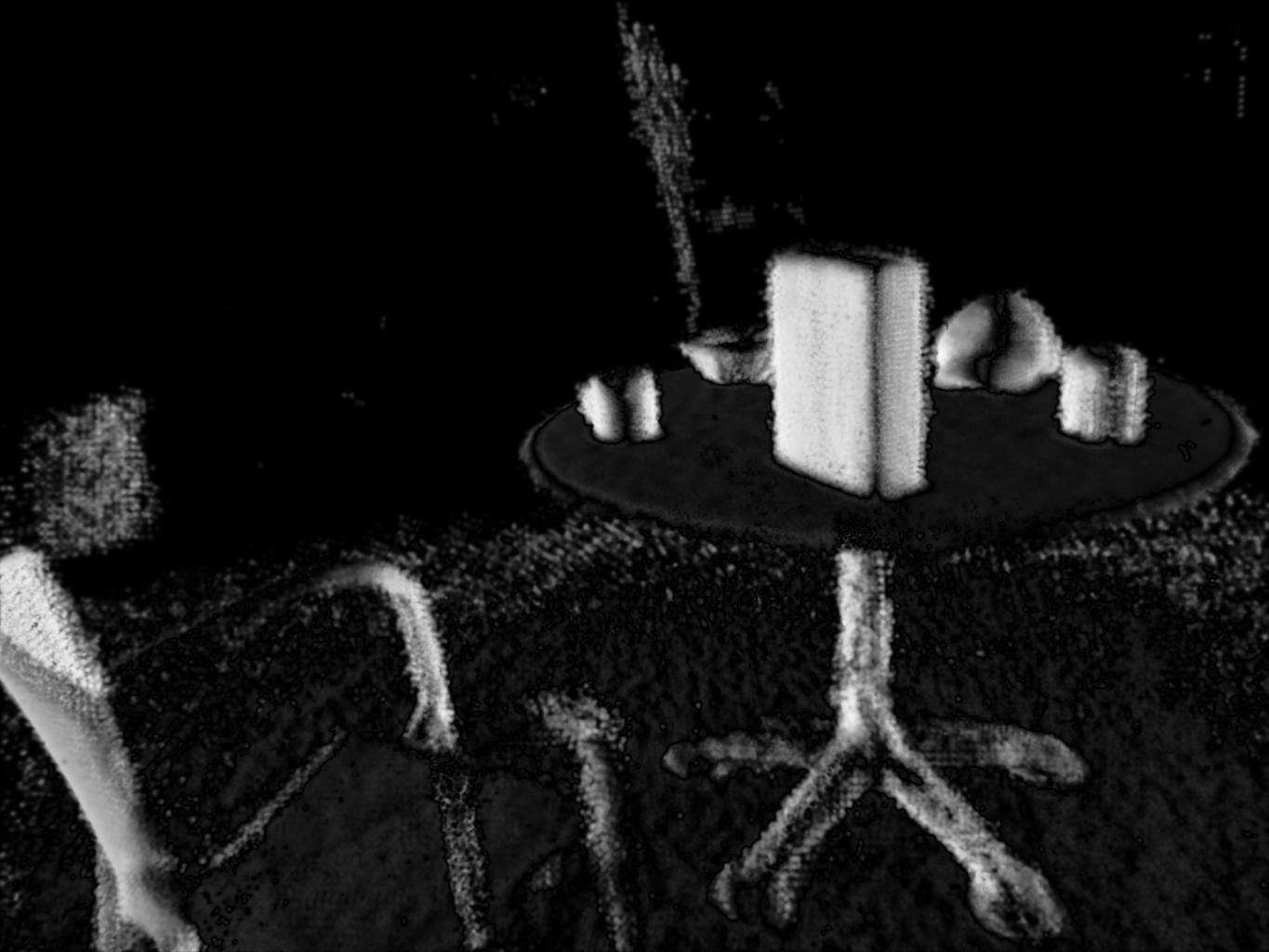}
        \caption{$p^{nx}$}
    \end{subfigure}  
    \begin{subfigure}{0.30\textwidth}
        \includegraphics[width=\textwidth]{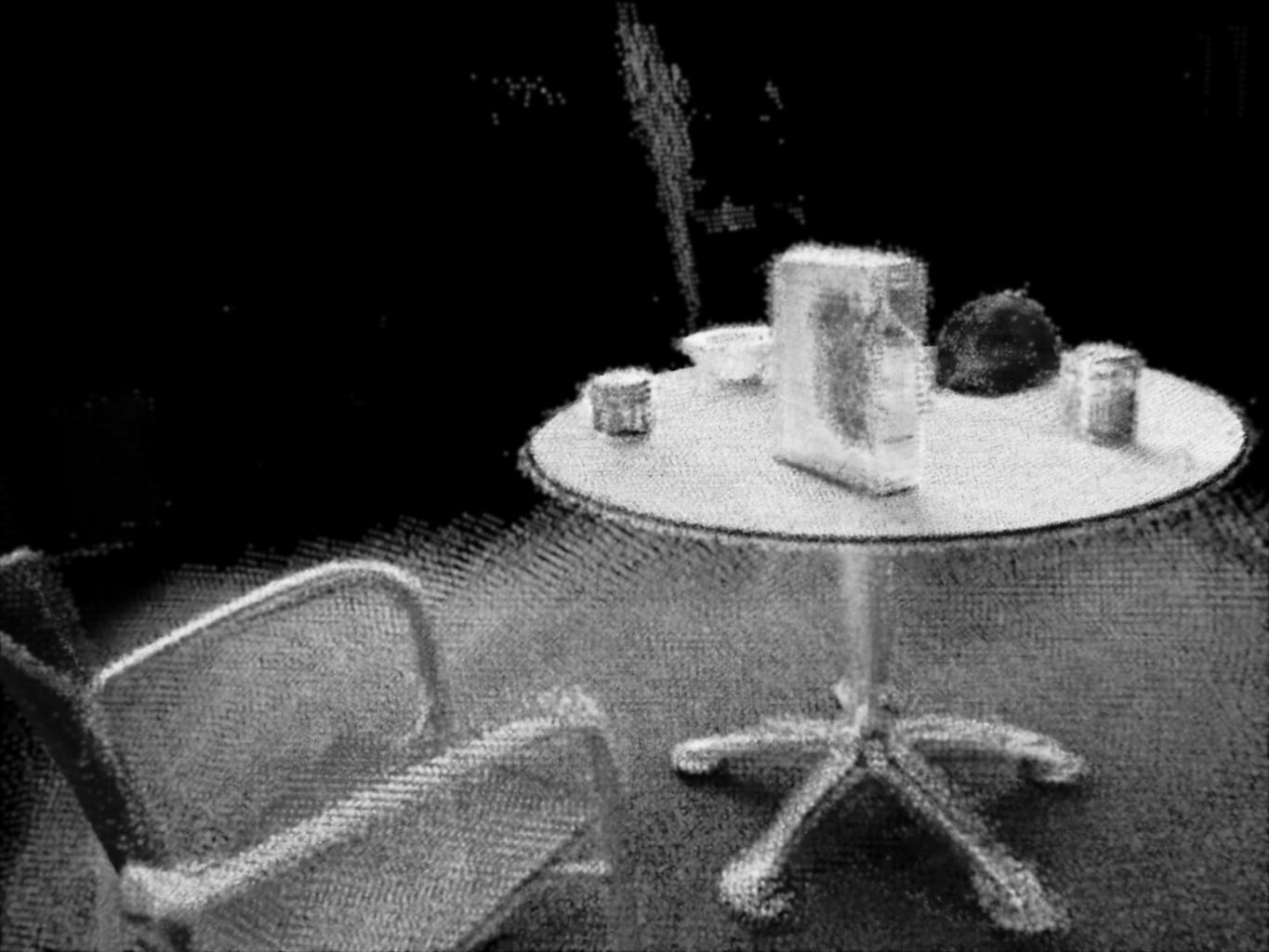}
        \caption{$p^g$}
    \end{subfigure}
    \caption{Example of images and rendered views from our dataset using point normals or intensity. From top to bottom: ETH3D, Tanks and Temples, CARLA, Livox, nuScenes, RGBDScenes.}
    \label{fig:synth}
\end{figure}

For these different sets, we computed normal directions using a local PCA of $50$ neighbors. All images were resized to a fixed height of $1024$px except for the RGBDScenes set, which was kept at a resolution of $640\times 480$. Example point clouds are shown in \cref{fig:exemples}, and images with point splatting renders are shown in \cref{fig:synth}.

\subsection{Manual registration}
\label{app:calib_results}
For the nuScenes, Tanks and Temples, and Livox datasets, we manually computed our ground truth from 2D-3D correspondences and least-squares minimization. For each image, we manually selected around 15 point-pixel pairs, which we used to find the camera position.

For Tanks and Temples, we also noticed the presence of camera distortion and approximate intrinsic values, which could not be solved by simply estimating the camera position. For this set, we therefore jointly optimized the camera extrinsics, intrinsics, and distortion through least-squares minimization, keeping a single camera model (intrinsics and distortion) among all the images of the same scene. Final reprojection errors are given in \cref{tab:reproj}. 

\begin{figure}
    \centering
    \begin{subfigure}{0.45\textwidth}
        \includegraphics[width=\textwidth]{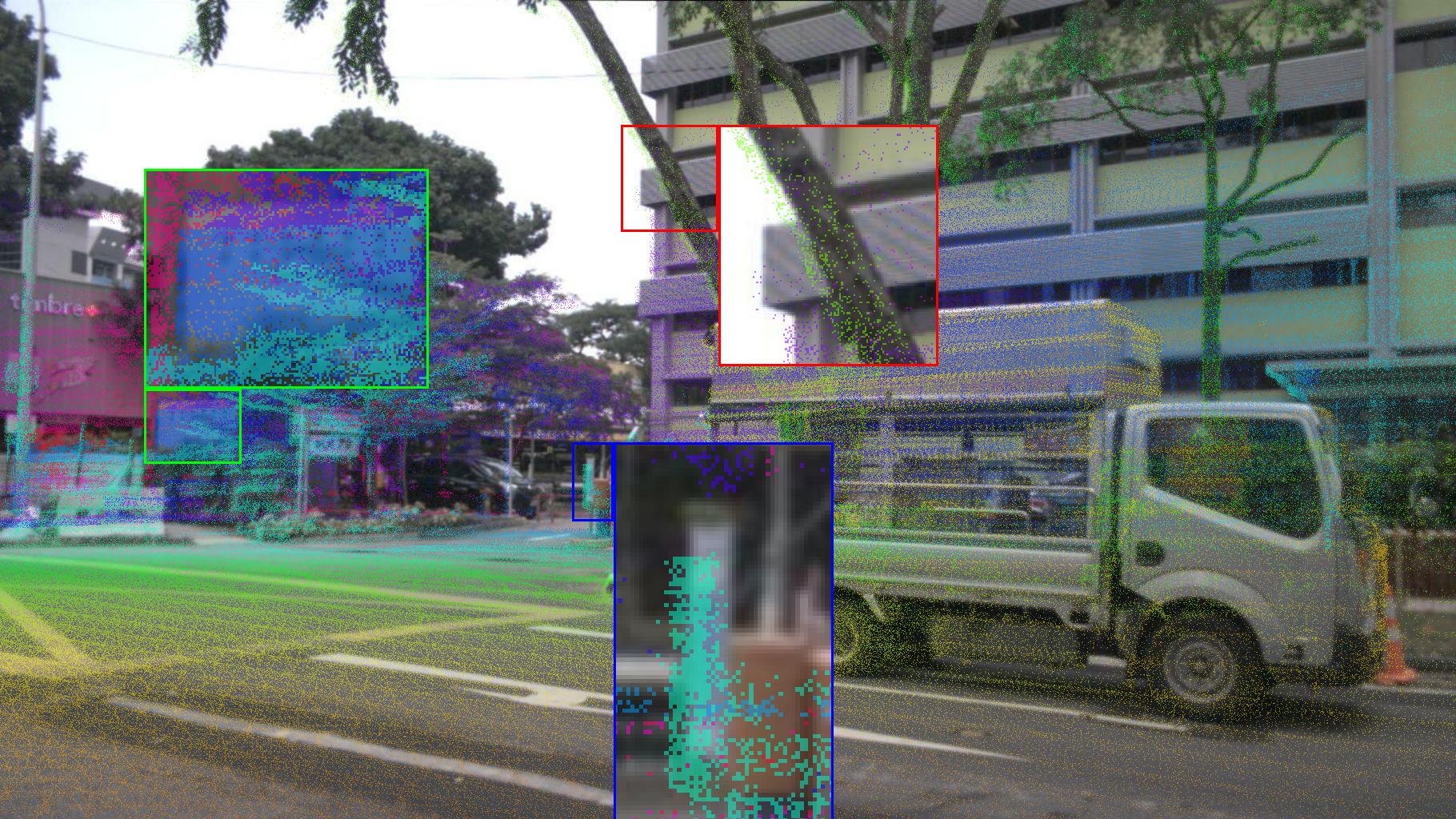}
    \end{subfigure}
    \begin{subfigure}{0.45\textwidth}
        \includegraphics[width=\textwidth]{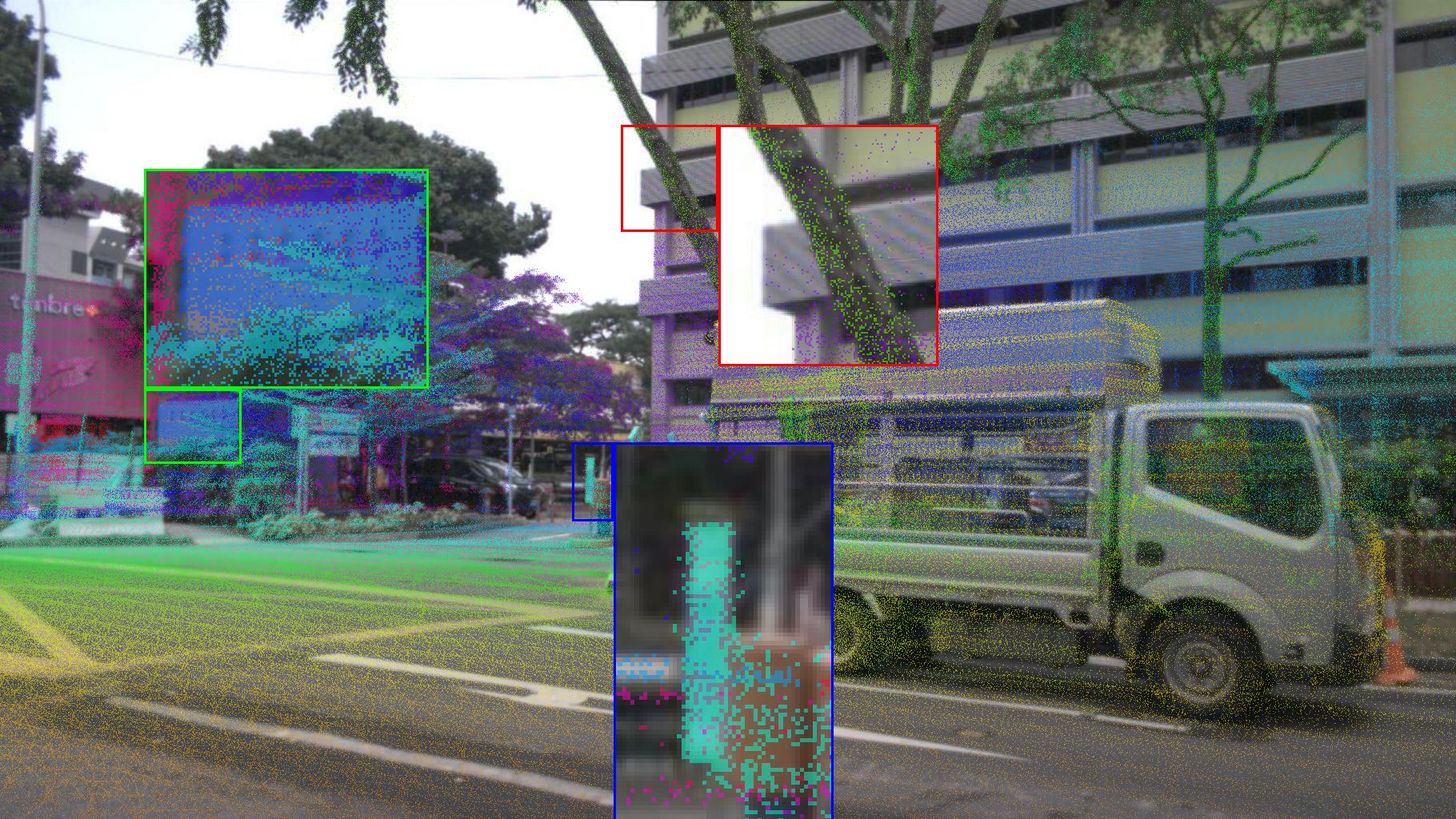}
    \end{subfigure}

    \begin{subfigure}{0.45\textwidth}
        \includegraphics[width=\textwidth]{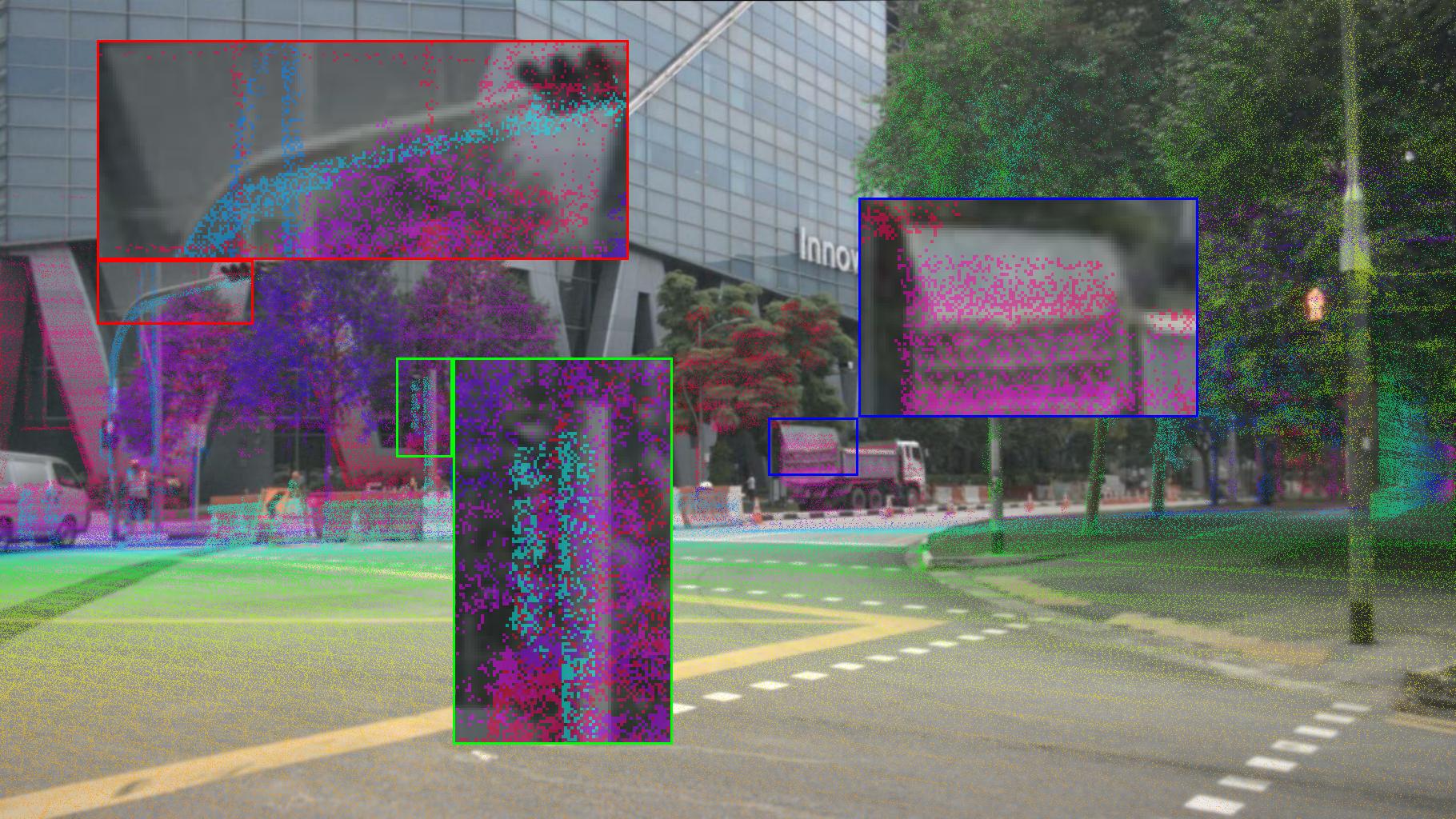}
    \end{subfigure}
    \begin{subfigure}{0.45\textwidth}
        \includegraphics[width=\textwidth]{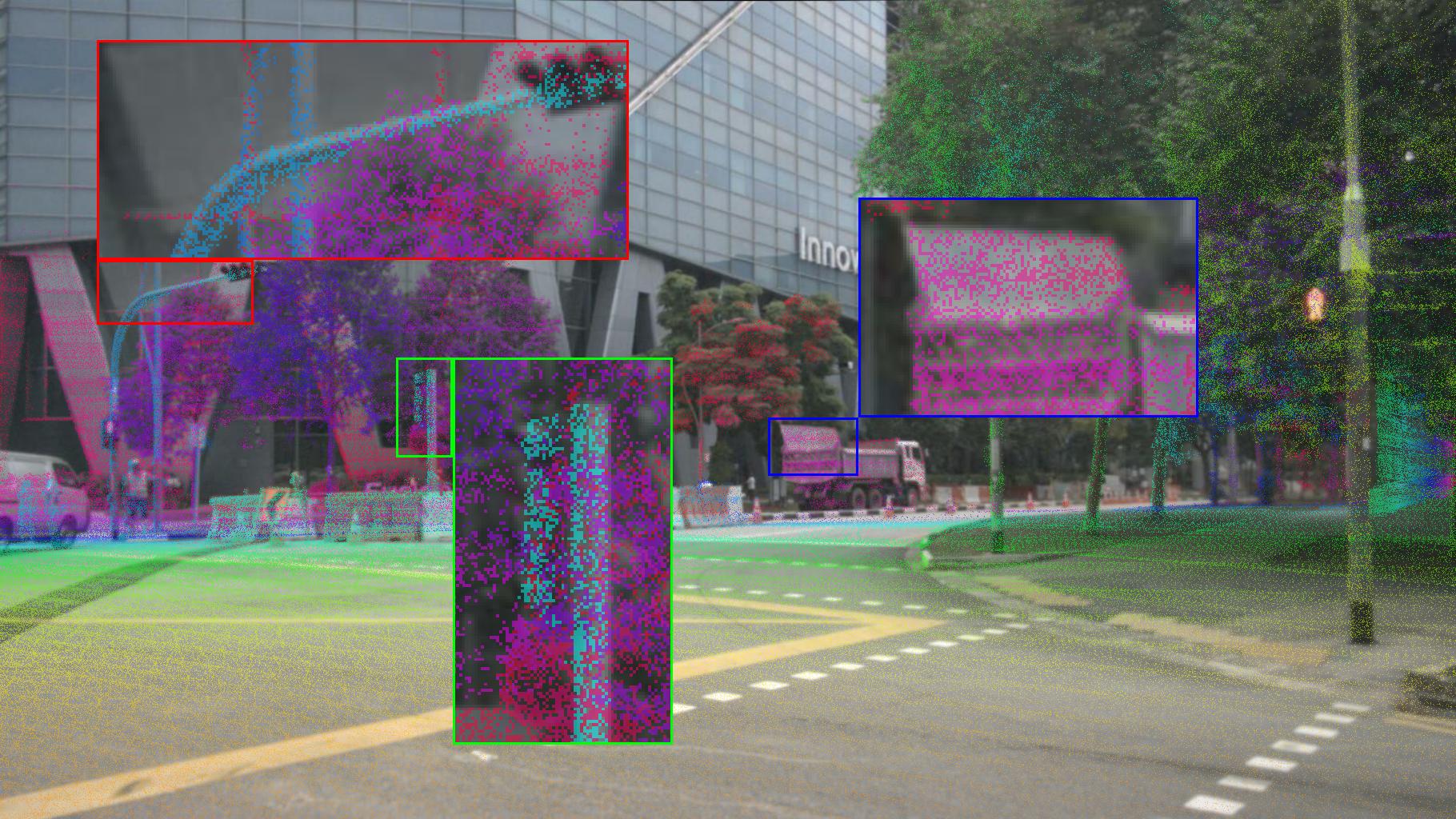}
    \end{subfigure}

    \begin{subfigure}{0.45\textwidth}
        \includegraphics[width=\textwidth]{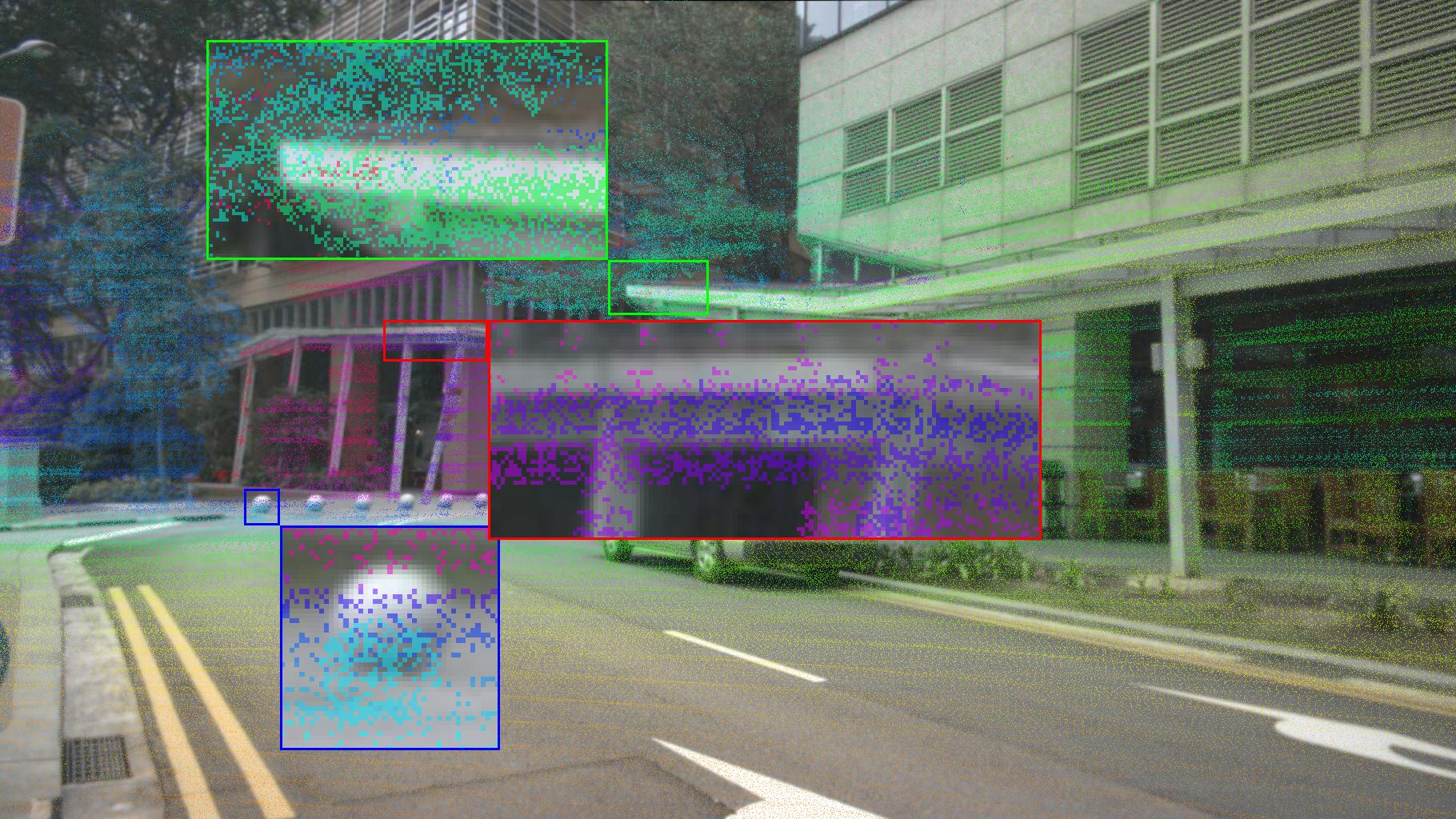}
    \end{subfigure}
    \begin{subfigure}{0.45\textwidth}
        \includegraphics[width=\textwidth]{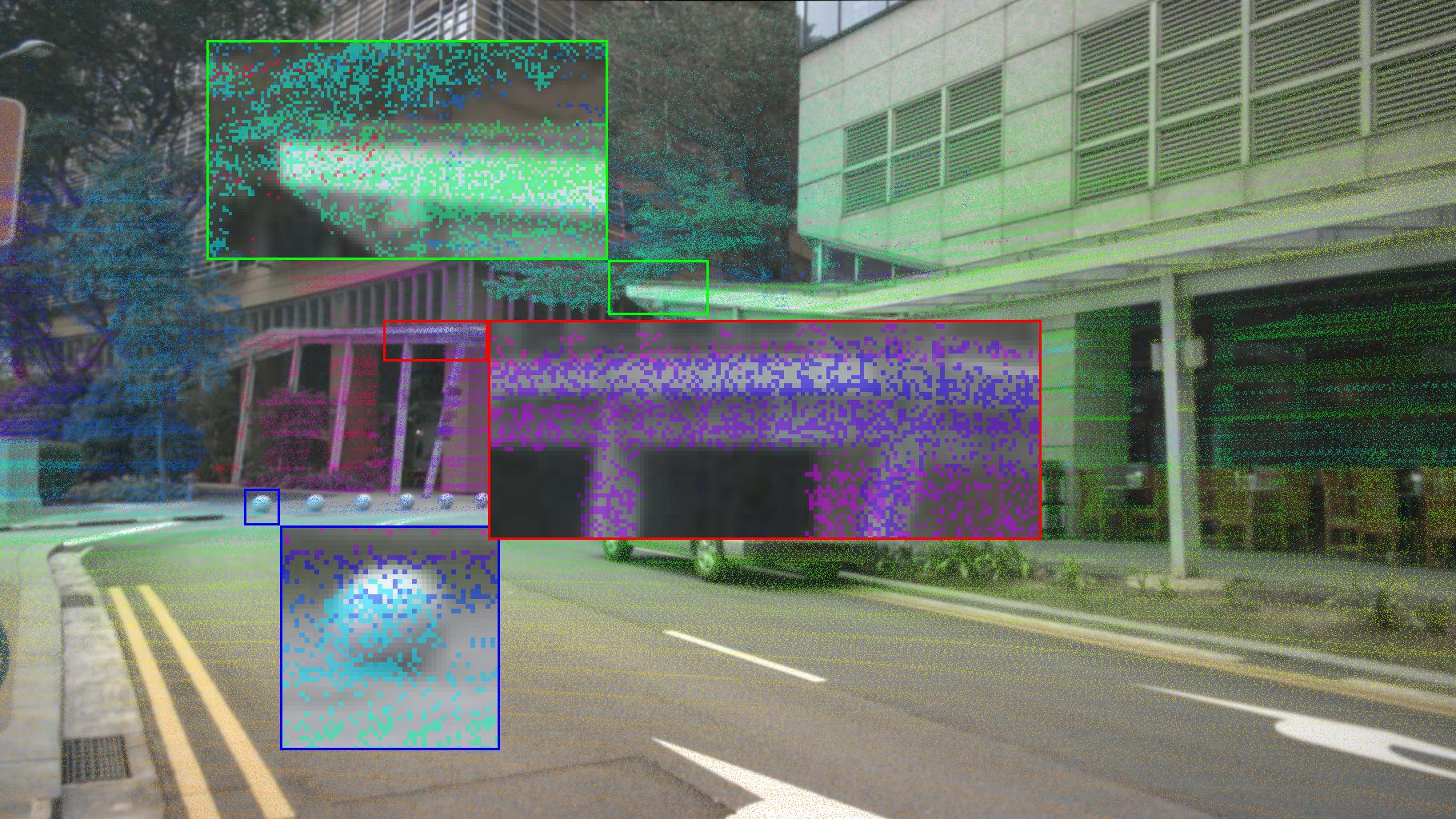}
    \end{subfigure}

        \begin{subfigure}{0.45\textwidth}
        \includegraphics[width=\textwidth]{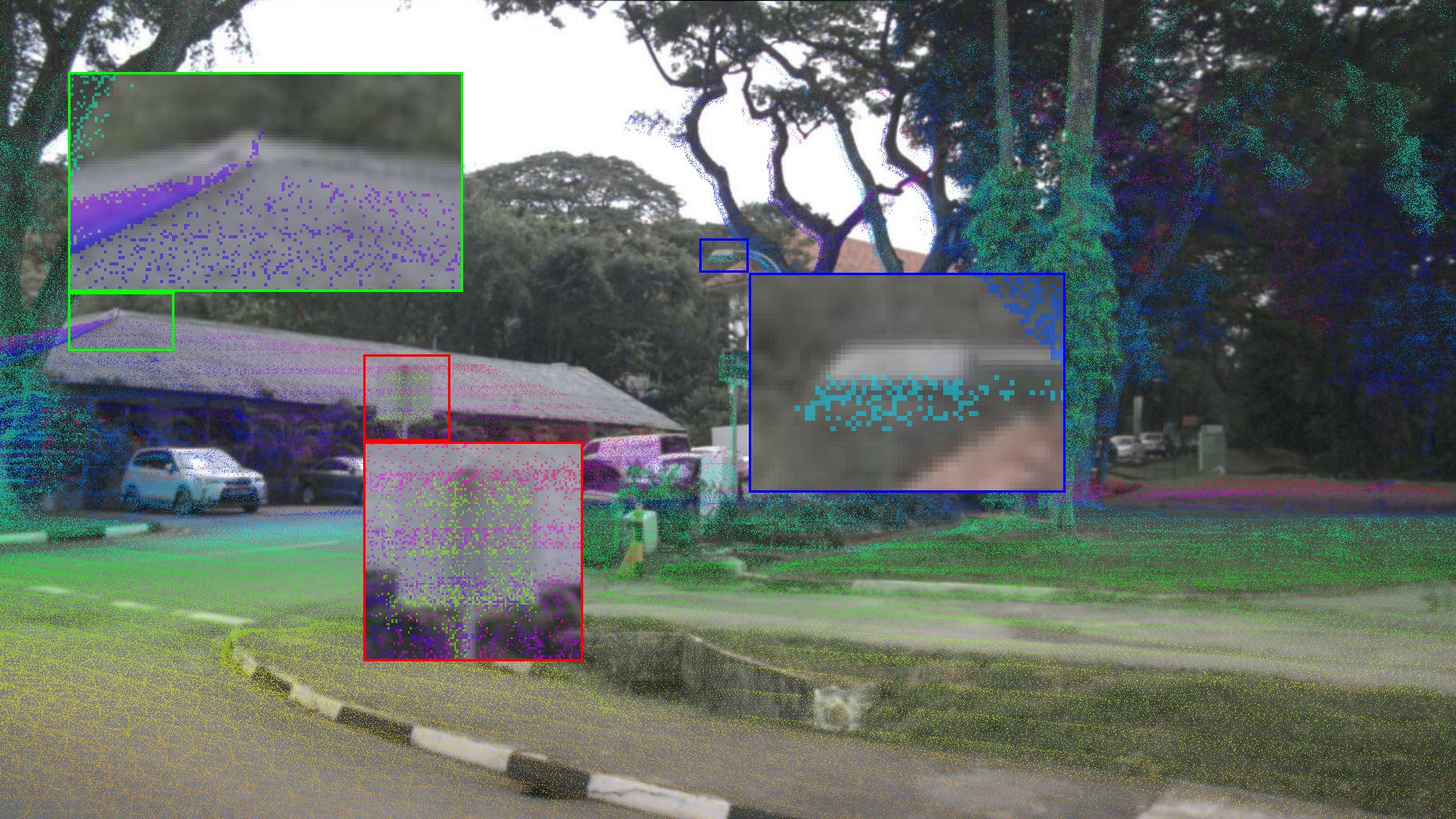}
    \end{subfigure}
    \begin{subfigure}{0.45\textwidth}
        \includegraphics[width=\textwidth]{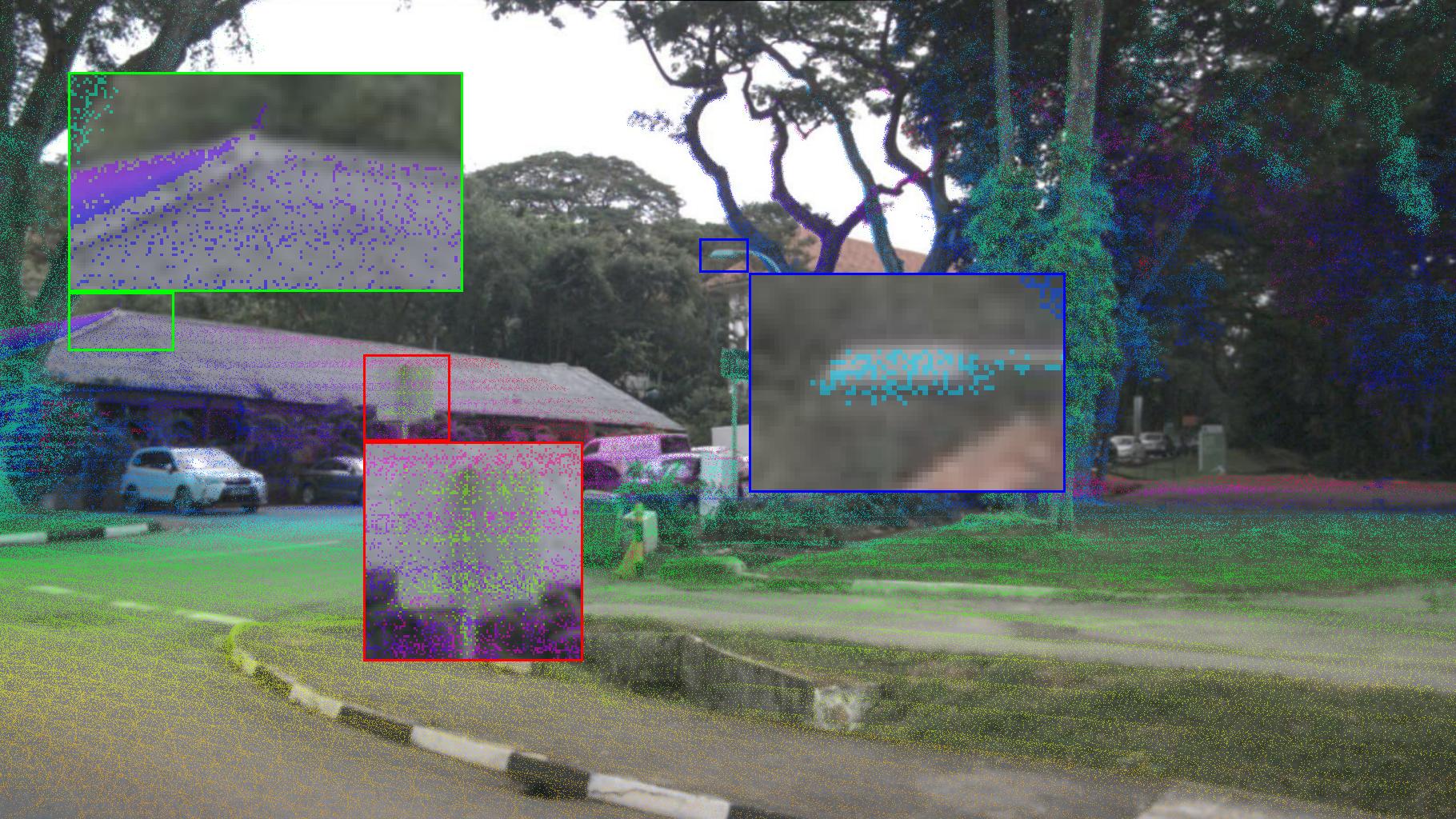}
    \end{subfigure}
    
    \caption{Manual registration results on nuScenes. Left: before manual registration, right: after.}
    \label{fig:manual_register_nu}
\end{figure}

\begin{figure}
    \centering
    \begin{subfigure}{0.45\textwidth}
        \includegraphics[width=\textwidth]{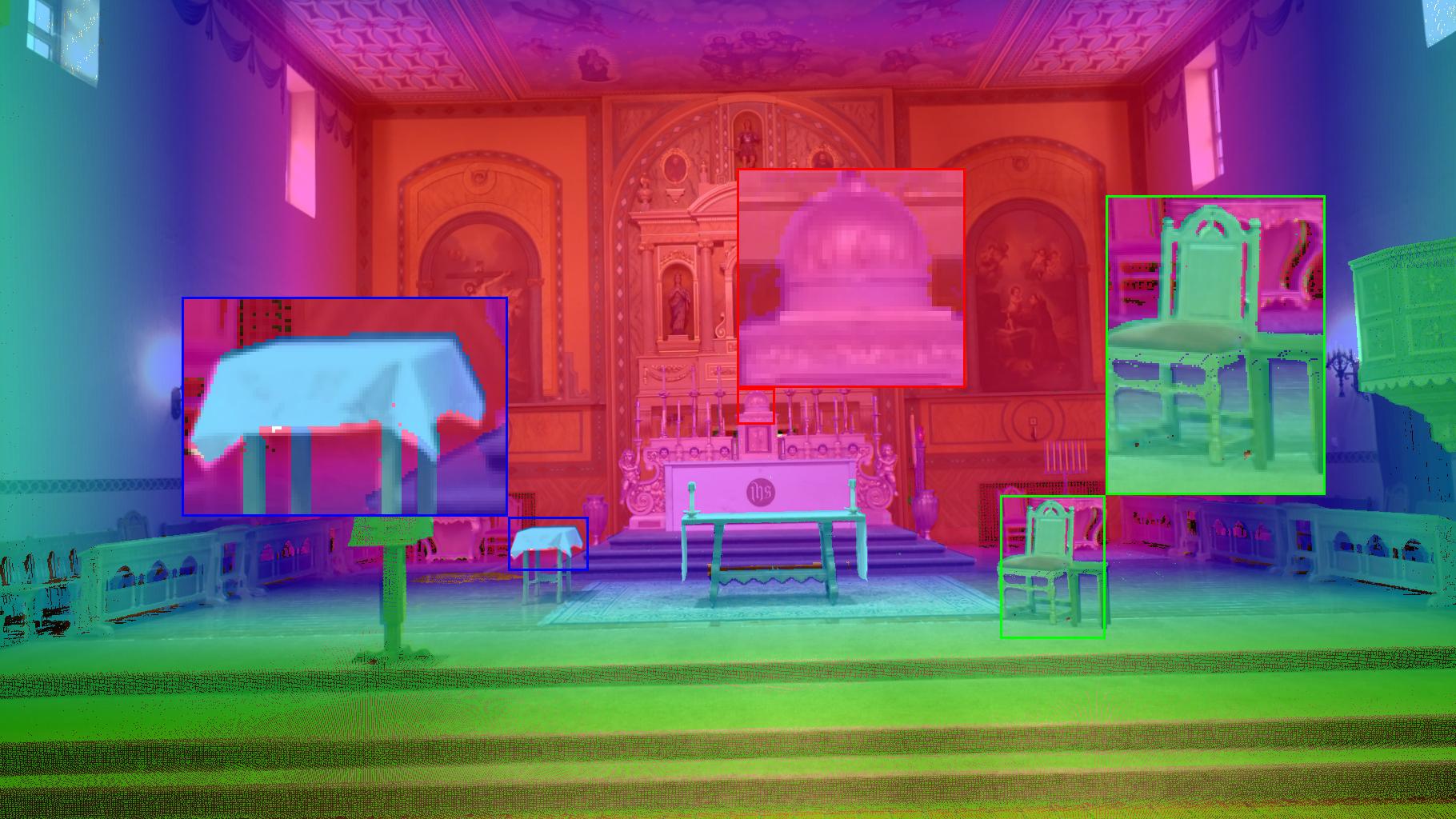}
    \end{subfigure}
    \begin{subfigure}{0.45\textwidth}
        \includegraphics[width=\textwidth]{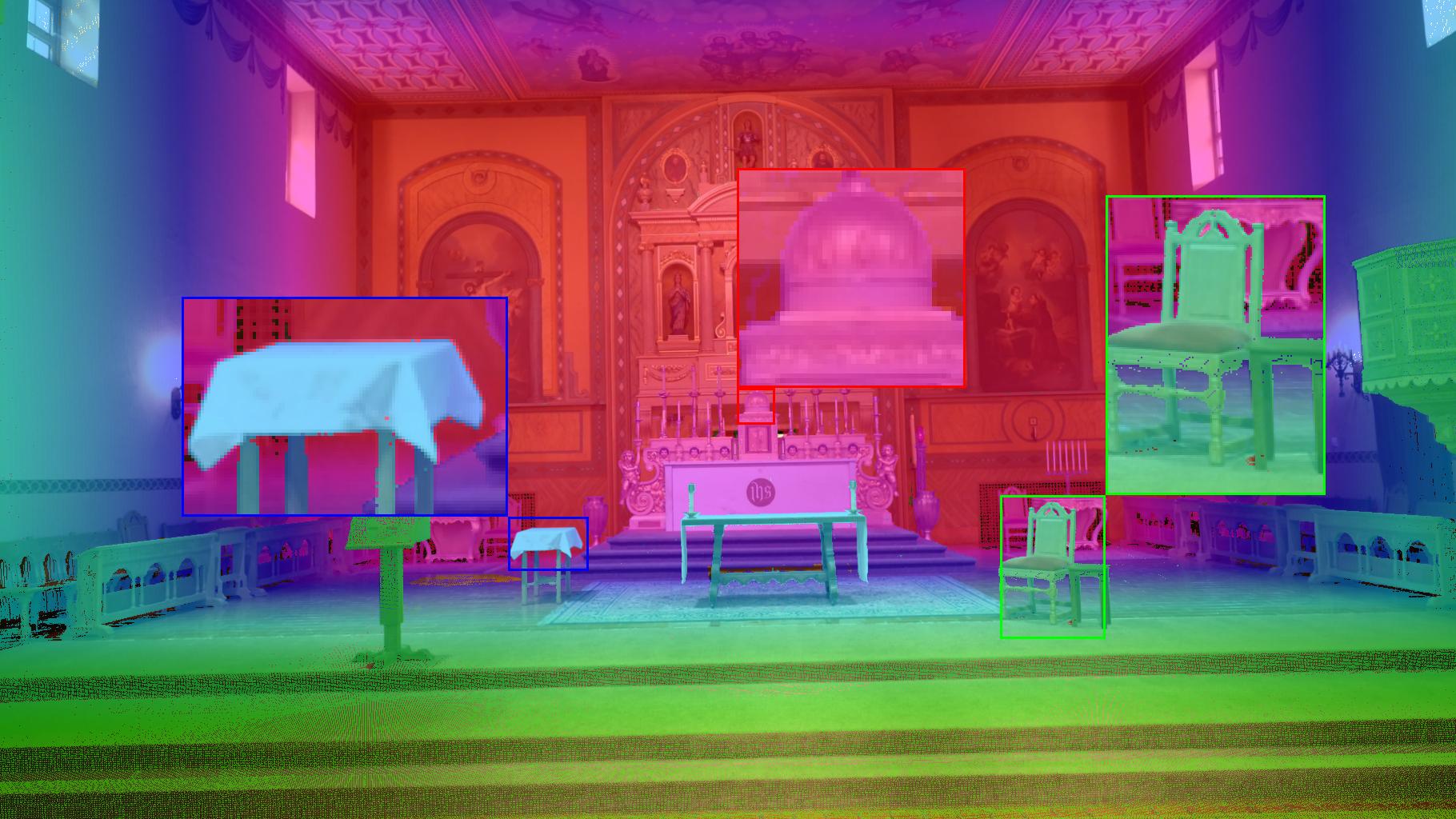}
    \end{subfigure}

    \begin{subfigure}{0.45\textwidth}
        \includegraphics[width=\textwidth]{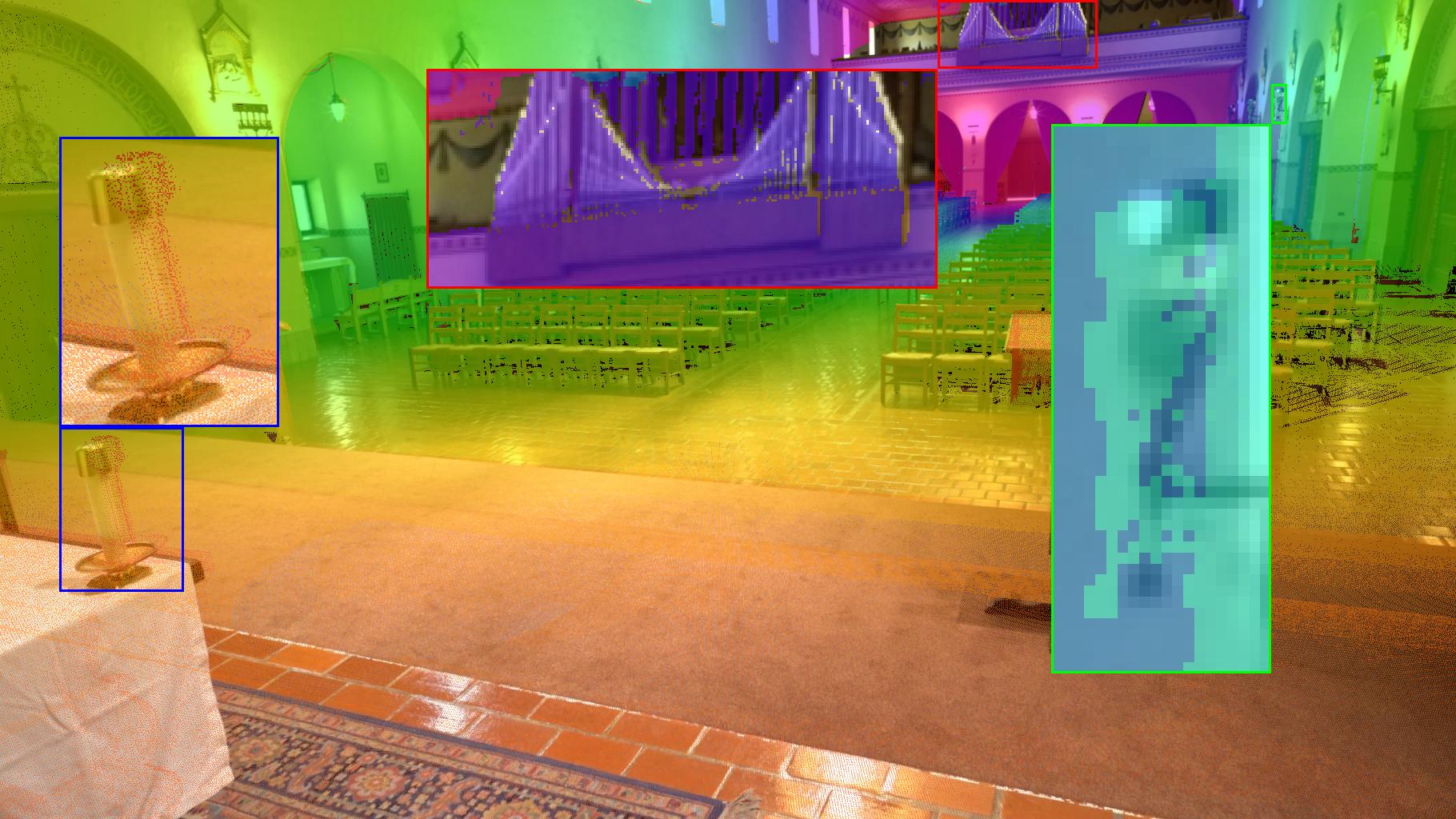}
    \end{subfigure}
    \begin{subfigure}{0.45\textwidth}
        \includegraphics[width=\textwidth]{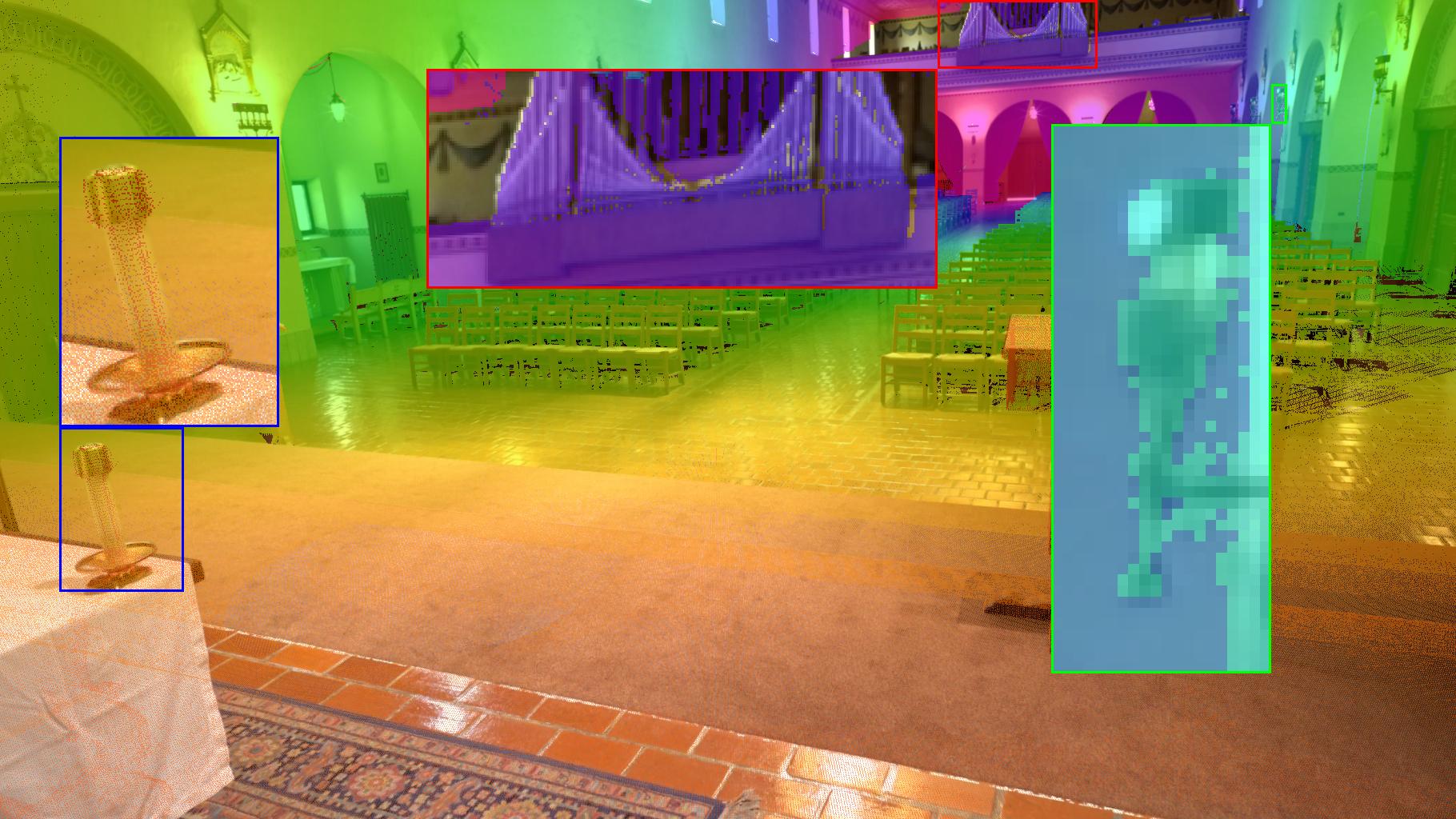}
    \end{subfigure}

    \begin{subfigure}{0.45\textwidth}
        \includegraphics[width=\textwidth]{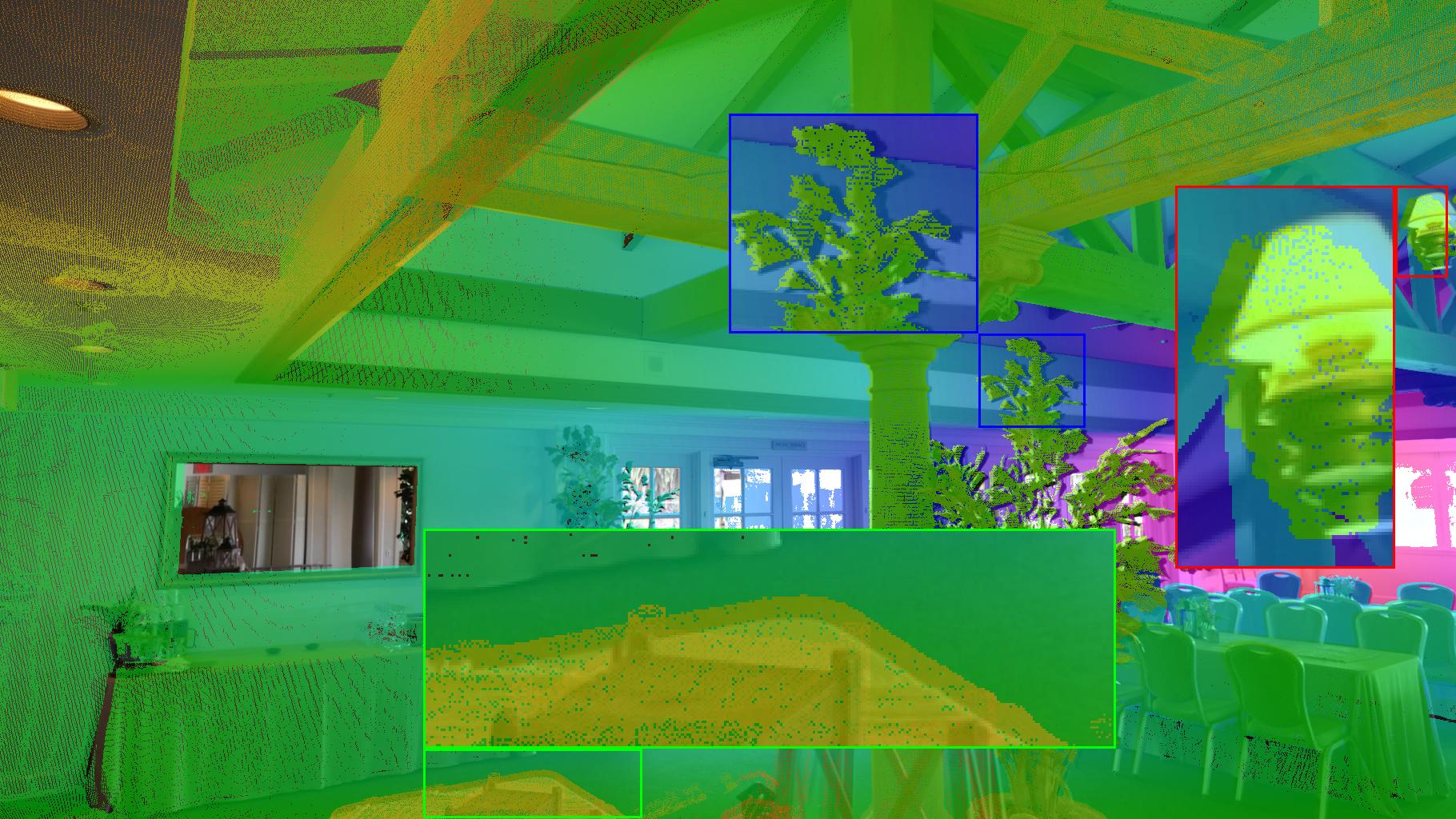}
    \end{subfigure}
    \begin{subfigure}{0.45\textwidth}
        \includegraphics[width=\textwidth]{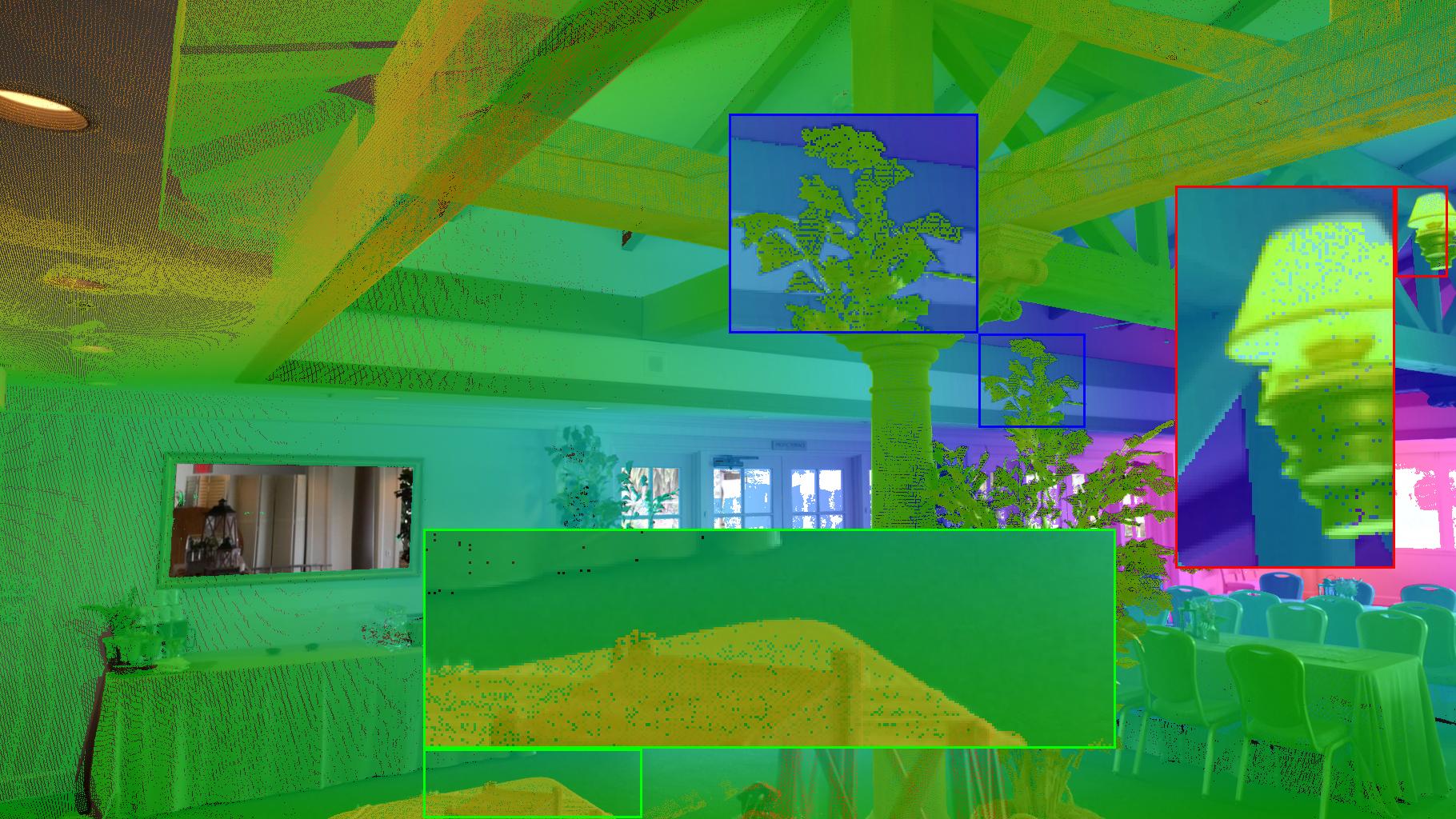}
    \end{subfigure}
    
    \begin{subfigure}{0.45\textwidth}
        \includegraphics[width=\textwidth]{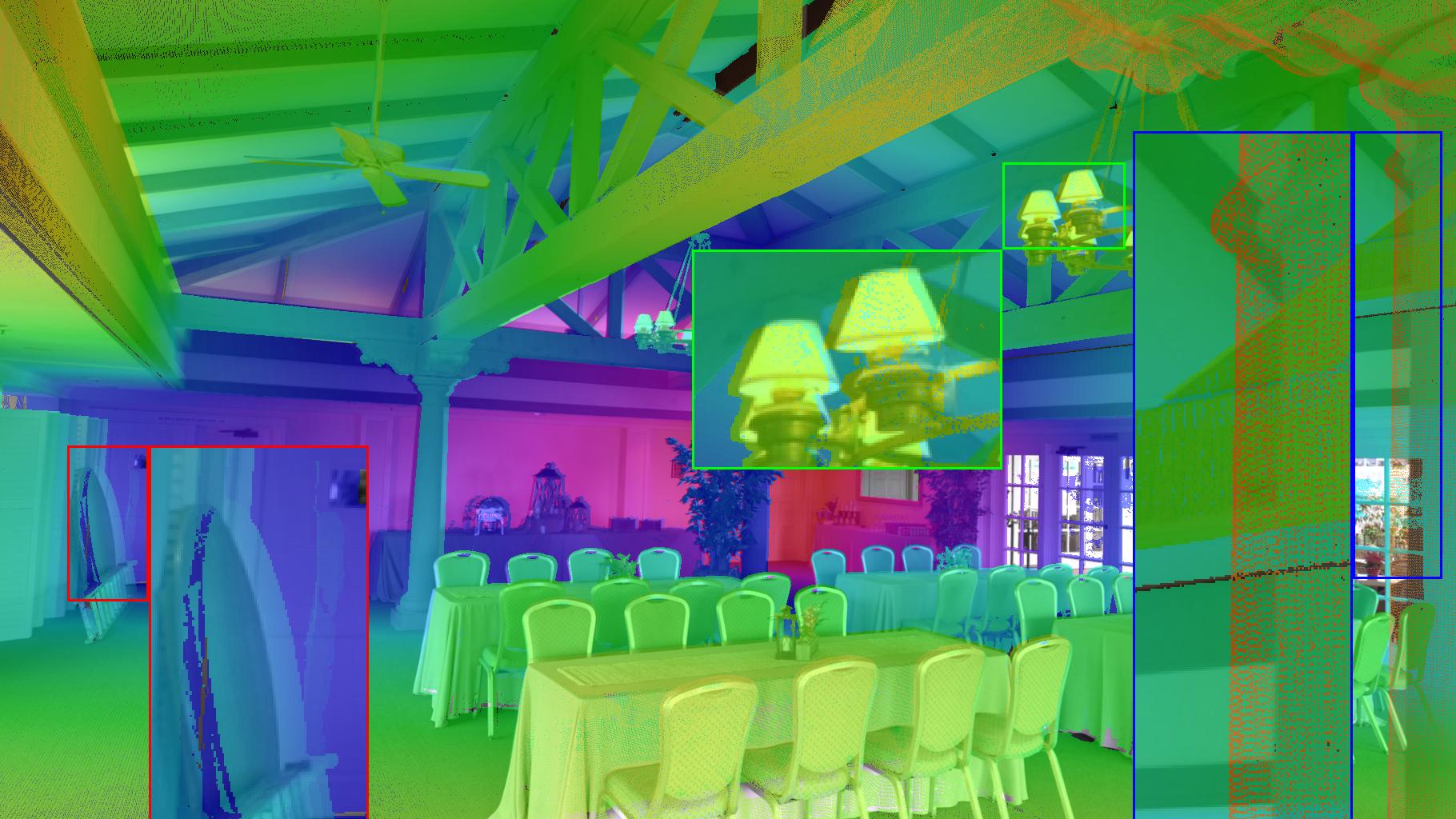}
    \end{subfigure}
    \begin{subfigure}{0.45\textwidth}
        \includegraphics[width=\textwidth]{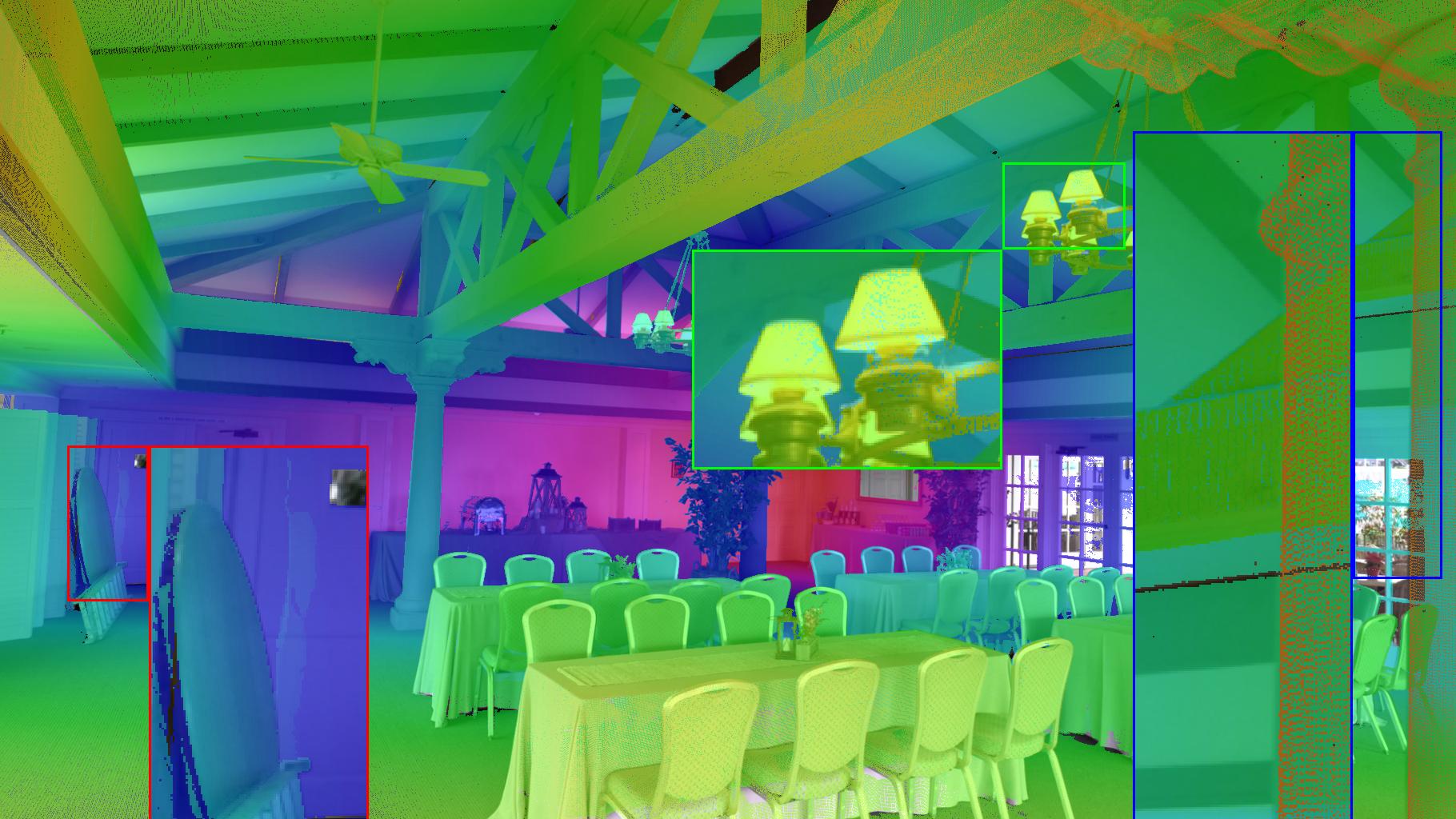}
    \end{subfigure}

    \caption{Manual registration results on Tanks and Temples. Left: before manual registration, right: after.}
    \label{fig:manual_register_tt}
\end{figure}

This manual process may itself present biases as the final given errors are computed with respect to the selected matches, which may themselves be inaccurate. This was especially the case on the nuScenes set, for which the image edges were often blurry because of the video acquisition. RGBDScenes samples were kept with their official ground truth, as the point cloud noise prevented us from selecting accurate enough correspondences. For most images, we were able to significantly reduce the reprojection error down to a few pixels compared to the official registration for each dataset. Our approach is all the more motivated by finding a method to avoid this manual labor. We also observe from \cref{tab:reproj} that the initial error given by the official registration is on the order of 7-10 pixels, based on our manual correspondences on Tanks and Temples and nuScenes. This supports our assumption that fine registration should be studied from an initialization with a reprojection error of about 10 pixels. 
Qualitative results of our registration procedure are shown for nuScenes and Tanks and Temples in \cref{fig:manual_register_nu} and \cref{fig:manual_register_tt}.

    \begin{table}[tb]
        \centering
        \resizebox*{!}{0.9\textheight}{%
        \begin{tabular}{@{}lcc}
            \toprule
        Set / Scene / Image &| Official registration (px) &| Manual registration (px) \\
\midrule
Tanks and Temples\\\quad Church\\\quad \quad 21.png & 5.13& 1.37\\
\quad \quad 40.png & 4.35& 1.89\\
\quad \quad 83.png & 7.27& 1.99\\
\quad \quad 110.png & 4.4& 1.28\\
\quad \quad 190.png & 4.0& 1.68\\
\quad \quad 281.png & 7.6& 1.24\\
\quad \quad 332.png & 6.44& 1.07\\
\quad \quad 358.png & 7.66& 1.37\\
\quad \quad 408.png & 5.11& 1.19\\
\quad \quad 438.png & 364.66& 1.31\\
\quad Meetingroom\\\quad \quad 2.png & 7.24& 1.83\\
\quad \quad 28.png & 8.1& 1.37\\
\quad \quad 43.png & 7.88& 1.31\\
\quad \quad 75.png & 7.47& 1.55\\
\quad \quad 117.png & 7.01& 1.49\\
\quad \quad 171.png & 5.09& 2.15\\
\quad \quad 183.png & 10.04& 2.16\\
\quad \quad 281.png & 7.48& 1.15\\
\quad \quad 301.png & 11.34& 1.55\\
\quad \quad 338.png & 9.21& 1.44\\
\midrule
nuScenes\\\quad scene 1\\\quad \quad 12.png & 11.26& 3.67\\
\quad \quad 18.png & 10.98& 4.69\\
\quad \quad 23.png & 8.3& 2.36\\
\quad \quad 32.png & 8.35& 4.6\\
\quad \quad 38.png & 7.53& 4.7\\
\quad scene 12\\\quad \quad 7.png & 9.03& 3.17\\
\quad \quad 26.png & 8.54& 2.55\\
\quad \quad 31.png & 7.53& 2.62\\
\quad \quad 34.png & 8.37& 2.97\\
\quad \quad 45.png & 7.89& 3.19\\
\quad scene 3\\\quad \quad 6.png & 8.76& 2.77\\
\quad \quad 11.png & 9.32& 3.1\\
\quad \quad 18.png & 8.55& 2.47\\
\quad \quad 32.png & 9.62& 2.91\\
\quad \quad 39.png & 9.2& 2.17\\
\quad scene 7\\\quad \quad 1.png & 8.46& 4.74\\
\quad \quad 9.png & 8.24& 2.72\\
\quad \quad 23.png & 9.05& 2.54\\
\quad \quad 27.png & 9.58& 2.71\\
\quad \quad 33.png & 8.18& 5.19\\
\midrule
Livox\\\quad \quad 0.png & N/A & 3.6\\
\quad \quad 1.png & N/A & 3.84\\
\quad \quad 2.png & N/A & 3.56\\
\quad \quad 3.png & N/A & 3.65\\

    \bottomrule
    \end{tabular}
    }
    \vspace{3pt}
    \caption{Reprojection error before and after manual registration of our images from Tanks and Temples, nuScenes, and Livox. Initial registration was obtained from the official dataset's ground truth for Tanks and Temples and using a state-of-the-art odometry algorithm for nuScenes.}
    \label{tab:reproj}
    \end{table}


\section{Benchmark information and additional experiments}
\subsection{Benchmark methods}
\label{app:benchmark}
We compared ReFineR with 4 different state-of-the-art methods for fine-level registration. 
\par MatchAnything~\cite{He2025MatchAnythingUC} is a transformer-based feature-matching model trained on multimodal images. We used their code and pretrained models available here\footnote{\url{https://huggingface.co/spaces/LittleFrog/MatchAnything/tree/main/imcui/third_party/MatchAnything}}. For our benchmark, we used the RoMa model as it yielded better matches for our task.
\par We ran Yuan \etal's method~\cite{yuanPixelLevelExtrinsicSelf2021} from their repository code\footnote{\url{https://github.com/hku-mars/livox_camera_calib}}. Depending on the scene modality, we used their predefined configurations in order to adapt the edge extraction to the point cloud scale. For instance, the voxel size parameter is set to 0.5 for indoor scenes and 1.0 for outdoor scenarios. All parameters can be found in their configuration file. 
\par For Koide \etal's method~\cite{koideGeneralSingleshotTargetless2023}\footnote{\url{https://github.com/koide3/direct\_visual\_lidar\_calibration}}, we only changed the "$\mathit{delta\_trans\_thresh}$" parameter, which defines the translation threshold used to claim algorithm convergence. For indoor and outdoor scenes, we respectively set it to $1$ and $10$ cm.
\par GS-SLAM~\cite{Matsuki2023GaussianSS} is the least robust method for our benchmark as it was designed to jointly optimize camera poses and Gaussians' parameters in a Gaussian Splatting model, which only leverages photometric information between the 3D and 2D modalities, and is difficult to apply to broader ranges of point clouds (LiDAR, Livox, RGBD).
\par We reimplemented the method from Taylor \etal~ \cite{Taylor2015MultiModalSC} with SciPy's version of the Nelder-Mead~\cite{nelderSimplexMethodFunction1965} optimizer.
\par 
Contrary to Yuan~\etal and Koide \etal's methods, ReFineR does not need any adjustment to the point cloud scale and scene modality. 



\subsection{Rotation and translation errors}
\label{app:rot_results}
Results from \cref{subsec:exp_geom} and \cref{subsec:exp_intens} are expressed in terms of rotation and translation errors in \cref{tab:normals_rot} and \cref{tab:intens_rot}, respectively. 

\begin{table}[h] 
    \centering \scriptsize \resizebox{\linewidth}{!}{%
\addtolength{\tabcolsep}{-0.4em}
\begin{tabular}{@{}lrrrrrrrrrrrr}
\toprule
& \multicolumn{2}{c}{ETH3D}
& \multicolumn{2}{c}{Tanks and Temples}
& \multicolumn{2}{c}{CARLA}
& \multicolumn{2}{c}{Livox}
& \multicolumn{2}{c}{nuScenes}
& \multicolumn{2}{c}{RGBDScenes}
\\
\cmidrule(lr){2-3}
\cmidrule(lr){4-5}
\cmidrule(lr){6-7}
\cmidrule(lr){8-9}
\cmidrule(lr){10-11}
\cmidrule(lr){12-13}
 & rot (°$\downarrow$) & transl (cm$\downarrow$)
 & rot (°$\downarrow$) & transl (cm$\downarrow$)
 & rot (°$\downarrow$) & transl (cm$\downarrow$)
 & rot (°$\downarrow$) & transl (cm$\downarrow$)
 & rot (°$\downarrow$) & transl (cm$\downarrow$)
 & rot (°$\downarrow$) & transl (cm$\downarrow$)
\\
\midrule 
\textbf{One stage} \\
GS-SLAM~\cite{Matsuki2023GaussianSS}
	 & {\smallpm{8.19}{2.30}} & {\smallpm{57.08}{37.98}} & {\smallpm{8.05}{1.32}} & {\smallpm{101.69}{69.42}} & {\smallpm{9.00}{1.23}} & {\smallpm{333.86}{101.20}} & {\smallpm{7.58}{1.15}} & {\smallpm{71.95}{38.52}} & {\smallpm{6.92}{2.08}} & {\smallpm{180.34}{37.02}} & {\smallpm{7.93}{2.57}} & {\smallpm{22.35}{5.49}}\\
Yuan \etal~\cite{yuanPixelLevelExtrinsicSelf2021}
	 & {\smallpm{0.75}{0.92}} & {\smallpm{7.99}{22.47}} & {\smallpm{0.30}{0.47}} & {\smallpm{6.12}{11.04}} & {\smallpm{0.79}{1.51}} & {\smallpm{199.68}{497.18}} & {\smallpm{0.46}{0.12}} & \cellcolor{orange!40} {\smallpm{6.79}{6.46}} & {\smallpm{1.17}{2.76}} & {\smallpm{90.93}{179.32}} & {\smallpm{4.92}{4.11}} & {\smallpm{12.78}{20.35}}\\
Koide \etal~\cite{koideGeneralSingleshotTargetless2023}
	 & {\smallpm{0.79}{2.53}} & {\smallpm{11.52}{17.04}} & {\smallpm{1.33}{3.68}} & {\smallpm{32.38}{67.31}} & {\smallpm{0.48}{0.79}} & {\smallpm{94.46}{351.19}} & {\smallpm{0.35}{0.47}} & {\smallpm{9.78}{8.92}} & {\smallpm{0.49}{0.60}} & {\smallpm{46.30}{67.05}} & {\smallpm{6.60}{3.71}} & {\smallpm{23.38}{14.46}}\\
Taylor \etal~\cite{Taylor2015MultiModalSC}
	 & {\smallpm{0.55}{0.48}} & {\smallpm{6.28}{10.30}} & {\smallpm{0.38}{0.35}} & {\smallpm{7.19}{12.59}} & {\smallpm{0.37}{0.41}} & {\smallpm{84.86}{172.55}} & {\smallpm{0.57}{0.25}} & {\smallpm{8.17}{7.14}} & {\smallpm{0.76}{1.09}} & {\smallpm{65.97}{165.04}} & {\smallpm{2.26}{1.02}} & \cellcolor{orange!40} {\smallpm{5.64}{2.86}}\\
MatchAnything~\cite{He2025MatchAnythingUC}
	 & \cellcolor{orange!40} {\smallpm{0.13}{0.11}} & \cellcolor{orange!40} {\smallpm{1.32}{1.62}} & \cellcolor{orange!40} {\smallpm{0.12}{0.08}} & \cellcolor{orange!40} {\smallpm{2.78}{3.75}} & \cellcolor{orange!40} {\smallpm{0.15}{0.17}} & \cellcolor{orange!40} {\smallpm{37.95}{53.56}} & \cellcolor{red!40} {\smallpm{0.28}{0.11}} & \cellcolor{red!40} {\smallpm{4.84}{4.99}} & \cellcolor{red!40} {\smallpm{0.15}{0.11}} & \cellcolor{orange!40} {\smallpm{14.93}{20.84}} & \cellcolor{red!40} {\smallpm{1.13}{0.55}} & \cellcolor{red!40} {\smallpm{2.48}{1.28}}\\
ReFineR
	 & \cellcolor{red!40} {\smallpm{0.10}{0.36}} & \cellcolor{red!40} {\smallpm{1.02}{4.69}} & \cellcolor{red!40} {\smallpm{0.08}{0.22}} & \cellcolor{red!40} {\smallpm{1.24}{4.97}} & \cellcolor{red!40} {\smallpm{0.08}{0.25}} & \cellcolor{red!40} {\smallpm{4.44}{7.52}} & \cellcolor{orange!40} {\smallpm{0.33}{0.13}} & {\smallpm{6.86}{5.93}} & \cellcolor{orange!40} {\smallpm{0.17}{0.32}} & \cellcolor{red!40} {\smallpm{5.63}{9.76}} & \cellcolor{orange!40} {\smallpm{2.20}{1.00}} & {\smallpm{6.16}{2.58}}\\
\midrule 
\textbf{Two stages} \\
MatchAnything + GS-SLAM~\cite{Matsuki2023GaussianSS}
	 & {\smallpm{7.70}{2.22}} & {\smallpm{58.45}{33.06}} & {\smallpm{7.98}{0.81}} & {\smallpm{142.22}{67.17}} & {\smallpm{8.75}{1.32}} & {\smallpm{331.93}{105.16}} & {\smallpm{7.75}{0.94}} & {\smallpm{83.07}{33.79}} & {\smallpm{7.41}{1.83}} & {\smallpm{184.24}{36.25}} & {\smallpm{nan}{nan}} & {\smallpm{22.11}{11.17}}\\
MatchAnything + Yuan \etal~\cite{yuanPixelLevelExtrinsicSelf2021}
	 & {\smallpm{0.77}{4.87}} & {\smallpm{7.45}{16.26}} & {\smallpm{0.26}{0.29}} & {\smallpm{6.79}{8.19}} & {\smallpm{0.58}{1.31}} & {\smallpm{117.48}{455.80}} & {\smallpm{0.34}{0.20}} & \cellcolor{red!40} {\smallpm{2.77}{6.56}} & {\smallpm{1.23}{3.20}} & {\smallpm{94.80}{582.17}} & {\smallpm{5.08}{3.94}} & {\smallpm{17.73}{16.64}}\\
MatchAnything + Koide \etal~\cite{koideGeneralSingleshotTargetless2023}
	 & {\smallpm{0.39}{1.33}} & {\smallpm{4.40}{12.41}} & {\smallpm{1.21}{3.30}} & {\smallpm{25.68}{81.56}} & {\smallpm{0.40}{1.52}} & {\smallpm{87.48}{258.70}} & {\smallpm{0.39}{0.14}} & {\smallpm{10.80}{6.77}} & {\smallpm{0.34}{1.65}} & {\smallpm{33.79}{186.61}} & {\smallpm{5.30}{4.23}} & {\smallpm{26.55}{19.88}}\\
MatchAnything + Taylor \etal~\cite{Taylor2015MultiModalSC}
	 & {\smallpm{0.13}{0.15}} & {\smallpm{1.38}{1.46}} & {\smallpm{0.12}{0.08}} & \cellcolor{orange!40} {\smallpm{2.22}{2.76}} & \cellcolor{orange!40} {\smallpm{0.18}{0.45}} & {\smallpm{44.04}{72.88}} & \cellcolor{orange!40} {\smallpm{0.31}{0.17}} & \cellcolor{orange!40} {\smallpm{4.02}{4.82}} & {\smallpm{0.24}{1.66}} & \cellcolor{orange!40} {\smallpm{17.31}{241.26}} & {\smallpm{1.39}{0.72}} & {\smallpm{3.39}{1.63}}\\
MatchAnything + MatchAnything~\cite{He2025MatchAnythingUC}
	 & \cellcolor{orange!40} {\smallpm{0.11}{0.08}} & \cellcolor{orange!40} {\smallpm{1.03}{1.65}} & \cellcolor{orange!40} {\smallpm{0.10}{0.08}} & {\smallpm{2.80}{2.38}} & {\smallpm{0.19}{0.17}} & \cellcolor{orange!40} {\smallpm{36.51}{56.88}} & \cellcolor{red!40} {\smallpm{0.31}{0.09}} & {\smallpm{4.22}{4.59}} & \cellcolor{orange!40} {\smallpm{0.16}{0.13}} & {\smallpm{19.34}{16.04}} & {\smallpm{0.99}{0.54}} & \cellcolor{red!40} {\smallpm{2.74}{1.49}}\\
MatchAnything* + ReFineR
	 & \cellcolor{red!40} {\smallpm{0.06}{0.05}} & \cellcolor{red!40} {\smallpm{0.70}{0.62}} & \cellcolor{red!40} {\smallpm{0.06}{0.04}} & \cellcolor{red!40} {\smallpm{1.24}{0.65}} & \cellcolor{red!40} {\smallpm{0.06}{0.09}} & \cellcolor{red!40} {\smallpm{3.05}{6.39}} & {\smallpm{0.38}{0.09}} & {\smallpm{5.81}{3.25}} & \cellcolor{red!40} {\smallpm{0.13}{0.14}} & \cellcolor{red!40} {\smallpm{4.92}{4.80}} & {\smallpm{1.09}{0.89}} & \cellcolor{orange!40} {\smallpm{2.78}{2.14}}\\

    \bottomrule
        \end{tabular}} 
 \vspace{1pt}\caption{Fine image to raw point cloud registration results (\cref{subsec:exp_geom}). Each image was registered using rendered point cloud normals.}\label{tab:normals_rot}
    \end{table}
    
\begin{table}[h] 
    \centering \scriptsize \resizebox{\linewidth}{!}{%
\addtolength{\tabcolsep}{-0.4em}
\begin{tabular}{@{}lrrrrrrrrrr}
\toprule
& \multicolumn{2}{c}{ETH3D}
& \multicolumn{2}{c}{Tanks and Temples}
& \multicolumn{2}{c}{Livox}
& \multicolumn{2}{c}{nuScenes}
& \multicolumn{2}{c}{RGBDScenes}
\\
\cmidrule(lr){2-3}
\cmidrule(lr){4-5}
\cmidrule(lr){6-7}
\cmidrule(lr){8-9}
\cmidrule(lr){10-11}
 & rot (°$\downarrow$) & transl (cm$\downarrow$)
 & rot (°$\downarrow$) & transl (cm$\downarrow$)
 & rot (°$\downarrow$) & transl (cm$\downarrow$)
 & rot (°$\downarrow$) & transl (cm$\downarrow$)
 & rot (°$\downarrow$) & transl (cm$\downarrow$)
\\
\midrule 
\textbf{One stage} \\
GS-SLAM~\cite{Matsuki2023GaussianSS}
	 & {\smallpm{0.47}{1.87}} & {\smallpm{4.31}{20.80}} & {\smallpm{0.10}{1.36}} & \cellcolor{orange!40} {\smallpm{1.62}{67.28}} & {\smallpm{2.57}{2.03}} & {\smallpm{74.66}{39.92}} & {\smallpm{4.37}{2.66}} & {\smallpm{131.13}{57.91}} & {\smallpm{2.81}{2.22}} & {\smallpm{17.61}{7.13}}\\
Koide \etal~\cite{koideGeneralSingleshotTargetless2023}
	 & \cellcolor{red!40} {\smallpm{0.05}{0.03}} & \cellcolor{red!40} {\smallpm{0.38}{0.56}} & \cellcolor{red!40} {\smallpm{0.06}{0.04}} & {\smallpm{1.82}{0.73}} & \cellcolor{orange!40} {\smallpm{0.26}{0.09}} & {\smallpm{8.13}{3.55}} & {\smallpm{0.48}{0.53}} & {\smallpm{30.81}{61.62}} & \cellcolor{orange!40} {\smallpm{0.77}{3.38}} & \cellcolor{orange!40} {\smallpm{2.06}{11.74}}\\
Taylor \etal~\cite{Taylor2015MultiModalSC}
	 & {\smallpm{0.55}{0.53}} & {\smallpm{6.03}{10.20}} & {\smallpm{0.33}{0.43}} & {\smallpm{6.30}{14.72}} & {\smallpm{0.93}{0.29}} & {\smallpm{9.58}{7.64}} & {\smallpm{0.92}{1.74}} & {\smallpm{50.40}{183.95}} & {\smallpm{2.13}{0.82}} & {\smallpm{5.92}{2.40}}\\
MatchAnything~\cite{He2025MatchAnythingUC}
	 & \cellcolor{orange!40} {\smallpm{0.07}{0.10}} & {\smallpm{0.66}{0.79}} & {\smallpm{0.07}{0.05}} & {\smallpm{1.64}{1.52}} & \cellcolor{red!40} {\smallpm{0.26}{0.15}} & \cellcolor{red!40} {\smallpm{3.14}{5.77}} & \cellcolor{red!40} {\smallpm{0.11}{0.06}} & \cellcolor{orange!40} {\smallpm{11.24}{10.67}} & \cellcolor{red!40} {\smallpm{0.40}{0.24}} & \cellcolor{red!40} {\smallpm{0.85}{0.64}}\\
ReFineR
	 & {\smallpm{0.08}{0.50}} & \cellcolor{orange!40} {\smallpm{0.57}{7.09}} & \cellcolor{orange!40} {\smallpm{0.07}{0.10}} & \cellcolor{red!40} {\smallpm{1.20}{1.56}} & {\smallpm{0.28}{0.16}} & \cellcolor{orange!40} {\smallpm{4.83}{6.31}} & \cellcolor{orange!40} {\smallpm{0.23}{0.28}} & \cellcolor{red!40} {\smallpm{7.43}{9.85}} & {\smallpm{2.12}{1.19}} & {\smallpm{5.77}{2.96}}\\
\midrule 
\textbf{Two stages} \\
MatchAnything + GS-SLAM~\cite{Matsuki2023GaussianSS}
	 & {\smallpm{0.13}{2.00}} & {\smallpm{1.15}{18.81}} & {\smallpm{0.08}{1.74}} & \cellcolor{orange!40} {\smallpm{1.51}{24.68}} & {\smallpm{3.00}{2.09}} & {\smallpm{82.11}{35.87}} & {\smallpm{4.91}{3.09}} & {\smallpm{141.23}{68.28}} & {\smallpm{2.73}{3.67}} & {\smallpm{16.21}{8.91}}\\
MatchAnything + Koide \etal~\cite{koideGeneralSingleshotTargetless2023}
	 & \cellcolor{orange!40} {\smallpm{0.05}{0.03}} & \cellcolor{red!40} {\smallpm{0.37}{0.54}} & \cellcolor{orange!40} {\smallpm{0.06}{0.04}} & {\smallpm{1.76}{0.75}} & {\smallpm{0.25}{0.09}} & {\smallpm{10.38}{4.60}} & {\smallpm{0.34}{0.52}} & {\smallpm{32.70}{96.20}} & {\smallpm{0.68}{3.56}} & {\smallpm{1.75}{15.71}}\\
MatchAnything + Taylor \etal~\cite{Taylor2015MultiModalSC}
	 & {\smallpm{0.07}{0.11}} & {\smallpm{0.66}{0.85}} & {\smallpm{0.08}{0.06}} & {\smallpm{1.63}{1.37}} & {\smallpm{0.28}{0.14}} & \cellcolor{red!40} {\smallpm{2.52}{6.02}} & {\smallpm{0.16}{1.00}} & {\smallpm{18.14}{55.96}} & {\smallpm{0.45}{0.37}} & {\smallpm{1.38}{1.06}}\\
MatchAnything + MatchAnything~\cite{He2025MatchAnythingUC}
	 & \cellcolor{orange!40} {\smallpm{0.05}{0.05}} & {\smallpm{0.49}{0.66}} & {\smallpm{0.07}{0.05}} & {\smallpm{1.51}{1.03}} & \cellcolor{orange!40} {\smallpm{0.24}{0.16}} & {\smallpm{3.85}{5.85}} & \cellcolor{orange!40} {\smallpm{0.12}{0.06}} & \cellcolor{orange!40} {\smallpm{10.63}{9.80}} & \cellcolor{orange!40} {\smallpm{0.35}{0.16}} & \cellcolor{orange!40} {\smallpm{0.88}{0.46}}\\
MatchAnything* + ReFineR
	 & \cellcolor{red!40} {\smallpm{0.04}{0.04}} & \cellcolor{orange!40} {\smallpm{0.40}{0.60}} & \cellcolor{red!40} {\smallpm{0.06}{0.04}} & \cellcolor{red!40} {\smallpm{1.17}{0.65}} & \cellcolor{red!40} {\smallpm{0.22}{0.16}} & \cellcolor{orange!40} {\smallpm{3.55}{6.21}} & \cellcolor{red!40} {\smallpm{0.11}{0.08}} & \cellcolor{red!40} {\smallpm{3.88}{3.02}} & \cellcolor{red!40} {\smallpm{0.24}{0.13}} & \cellcolor{red!40} {\smallpm{0.50}{0.62}}\\

    \bottomrule
        \end{tabular}} 
 \vspace{1pt}\caption{Fine image to intensity point cloud registration results (\cref{subsec:exp_intens}). Each image was registered using rendered point cloud reflectivity or grayscale value.}\label{tab:intens_rot}
    \end{table}

\subsection{Qualitative results}
\cref{fig:qualit_geom} and \cref{fig:qualit_intens} show qualitative results for registration on the raw point cloud and using point cloud intensity (grayscale value or LiDAR reflectivity).
\begin{figure}[tb]
    \centering
    \includegraphics[width=0.9\linewidth]{images/normals.jpg}
    \caption{Qualitative results from experiment \cref{subsec:exp_geom}.}
    \label{fig:qualit_geom}
\end{figure}
\begin{figure}[tb]
    \centering
    \includegraphics[width=0.9\linewidth]{images/intensity.jpg}
    \caption{Qualitative results from experiment \cref{subsec:exp_intens}.}
    \label{fig:qualit_intens}
\end{figure}

\subsection{Supervision channel}
\label{app:ablation_channel}
Apart from point cloud normals and intensity, as well as image RGB color, we experimented with different comparison modalities. Supervising the registration with the rendered depth (${\mathcal{C}_\leftrightarrow = \{(p^d, i^g)\})}$, we generally observe a decrease in registration performance in all benchmarking datasets. We suspect that this is the case because the main source of gradients in depth images arises from discontinuities between objects and gradients converging towards vanishing points, whereas rendered normal images from point clouds provide a richer set of gradients, including regions where the surface orientation changes (\eg, wall edges).

Another source of supervision could come from the monocular estimation of depth and normals with the Marigold diffusion model~\cite{Ke2025MarigoldAA}. From a single RGB image, Marigold uses its trained geometric priors to estimate the normal orientation and normalized depth from a single image. For the normals, we kept the point splatting rendering of normals and reoriented them using the rendered view such that both the normal frames from the render and the generation coincide, and the rendered depth was normalized. We compared these true rendered channels with the monocular estimated normals and depth ($\mathcal{C}_{\leftrightarrow}=\{(p^\mathit{nx}, i^\mathit{nx}), (p^\mathit{ny}, i^\mathit{ny}), (p^\mathit{nz}, i^\mathit{nz})\}$ and ${\mathcal{C}_\leftrightarrow = \{(p^d, i^d)\}}$). 
\cref{tab:ablation_channel} shows that using the monocular normals produces better registration than the monocular depth, and sometimes matches the performance of the ReFineR baseline. Both synthetic modalities, however, generally struggle at finer registration levels. We may explain this because of the coarse resemblance between the image's gradient and its diffusion-generated counterpart, which allows for some edge supervision, but this supervision is limited by the current performance of diffusion models, which still create blurry edges, removing any pixel-level registration ability. \\

\begin{table}[h] 
    \centering \scriptsize \resizebox{\linewidth}{!}{%
\addtolength{\tabcolsep}{-0.4em}
\begin{tabular}{@{}lrrrrrrrrrrrrrrrrrrrrrrrr}
\toprule
& \multicolumn{4}{c}{ETH3D}
& \multicolumn{5}{c}{Tanks and Temples}
& \multicolumn{4}{c}{CARLA}
& \multicolumn{3}{c}{Livox}
& \multicolumn{4}{c}{nuScenes}
& \multicolumn{4}{c}{RGBDScenes}
\\
\cmidrule(lr){2-5}
\cmidrule(lr){6-10}
\cmidrule(lr){11-14}
\cmidrule(lr){15-17}
\cmidrule(lr){18-21}
\cmidrule(lr){22-25}
& \multicolumn{3}{c}{pix@ (\%$\uparrow$)}  & $p_e$(px$\downarrow$)
& \multicolumn{4}{c}{pix@ (\%$\uparrow$)}  & $p_e$(px$\downarrow$)
& \multicolumn{3}{c}{pix@ (\%$\uparrow$)}  & $p_e$(px$\downarrow$)
& \multicolumn{2}{c}{pix@ (\%$\uparrow$)}  & $p_e$(px$\downarrow$)
& \multicolumn{3}{c}{pix@ (\%$\uparrow$)}  & $p_e$(px$\downarrow$)
& \multicolumn{3}{c}{pix@ (\%$\uparrow$)}  & $p_e$(px$\downarrow$)
\\
& 1 & 2 & 5 & 
& 1 & 2 & 3 & 5 & 
& 1 & 2 & 5 & 
& 5 & 10 & 
& 3 & 5 & 10 & 
& 3 & 5 & 10 &\\
\midrule 
\textbf{(VII)} \\
Rendered depth
	 & \cellcolor{orange!40} {39} & \cellcolor{orange!40} {93} & \cellcolor{red!40} {100} & \cellcolor{orange!40} {\smallpm{1.04}{0.55}} & \cellcolor{red!40} {0} & \cellcolor{orange!40} {58} & \cellcolor{orange!40} {92} & \cellcolor{red!40} {100} & \cellcolor{orange!40} {\smallpm{1.95}{0.76}} & \cellcolor{orange!40} {8} & \cellcolor{orange!40} {36} & \cellcolor{orange!40} {78} & \cellcolor{orange!40} {\smallpm{2.45}{7.23}} & {25} & \cellcolor{orange!40} {95} & {\smallpm{6.48}{2.32}} & \cellcolor{orange!40} {3} & \cellcolor{orange!40} {36} & \cellcolor{orange!40} {92} & \cellcolor{orange!40} {\smallpm{5.66}{8.57}} & {6} & {22} & {80} & {\smallpm{6.61}{4.58}}\\
Monocular normals
	 & {0} & {2} & {9} & {\smallpm{8.34}{2.26}} & \cellcolor{red!40} {0} & {28} & {90} & \cellcolor{red!40} {100} & {\smallpm{2.43}{0.58}} & {2} & {23} & {71} & {\smallpm{3.29}{6.49}} & \cellcolor{red!40} {65} & \cellcolor{red!40} {100} & \cellcolor{red!40} {\smallpm{4.94}{0.27}} & {0} & {12} & {71} & {\smallpm{7.15}{13.75}} & \cellcolor{red!40} {18} & \cellcolor{orange!40} {57} & \cellcolor{red!40} {94} & \cellcolor{red!40} {\smallpm{4.47}{2.83}}\\
Monocular depth
	 & {0} & {0} & \cellcolor{orange!40} {24} & {\smallpm{7.20}{5.17}} & \cellcolor{red!40} {0} & {5} & {32} & \cellcolor{orange!40} {83} & {\smallpm{3.62}{2.67}} & {2} & {7} & {41} & {\smallpm{5.76}{559.08}} & {5} & {60} & {\smallpm{9.77}{2.87}} & {0} & {2} & {37} & {\smallpm{11.88}{34.59}} & {5} & \cellcolor{red!40} {61} & {90} & \cellcolor{orange!40} {\smallpm{4.59}{3.03}}\\
MatchAnything* + ReFineR (RGB image)
	 & \cellcolor{red!40} {58} & \cellcolor{red!40} {96} & \cellcolor{red!40} {100} & \cellcolor{red!40} {\smallpm{0.90}{0.40}} & \cellcolor{red!40} {0} & \cellcolor{red!40} {70} & \cellcolor{red!40} {95} & \cellcolor{red!40} {100} & \cellcolor{red!40} {\smallpm{1.76}{0.51}} & \cellcolor{red!40} {16} & \cellcolor{red!40} {68} & \cellcolor{red!40} {91} & \cellcolor{red!40} {\smallpm{1.54}{3.04}} & \cellcolor{orange!40} {55} & \cellcolor{red!40} {100} & \cellcolor{orange!40} {\smallpm{4.94}{0.94}} & \cellcolor{red!40} {6} & \cellcolor{red!40} {54} & \cellcolor{red!40} {97} & \cellcolor{red!40} {\smallpm{4.90}{2.75}} & \cellcolor{orange!40} {13} & {43} & \cellcolor{orange!40} {92} & {\smallpm{5.36}{4.25}}\\

    \bottomrule
        \end{tabular}} 
 \vspace{1pt}\caption{Ablation study on ReFineR's supervision channel (\cref{app:ablation_channel}).}\label{tab:ablation_channel}
    \end{table}

\subsection{Registration at scale on ETH3D}
\label{app:exp_scale2}
We ran a similar experiment to the one in \cref{subsec:exp_scale} on the ETH3D dataset. Results in \cref{tab:scale2} show strong performance of the MatchAnything* + ReFineR pipeline even at large perturbations such as 100px.
\begin{table}[h] 
    \centering \scriptsize \resizebox{\linewidth}{!}{%
\begin{tabular}{@{}lrrrrrrrrrrr}
\toprule
& \multicolumn{11}{c}{ETH3D}
\\
\cmidrule(lr){2-12}
& \multicolumn{8}{c}{pix@ (\%$\uparrow$)}  & $p_e$(px$\downarrow$) & rot (°$\downarrow$) & transl (cm$\downarrow$)
\\
& 1 & 2 & 3 & 5 & 10 & 20 & 50 & 100 &  &  &\\
\midrule 
\textbf{0px initialization} \\
MatchAnything
	 & 14 & 67 & 98 & 100 & 100 & 100 & 100 & 100 & {\smallpm{1.66}{0.66}} & {\smallpm{0.10}{0.05}} & {\smallpm{0.94}{1.55}}\\
ReFineR
	 & 64 & 100 & 100 & 100 & 100 & 100 & 100 & 100 & {\smallpm{0.89}{0.35}} & {\smallpm{0.06}{0.04}} & {\smallpm{0.70}{0.62}}\\
MatchAnything* + ReFineR
	 & 65 & 97 & 100 & 100 & 100 & 100 & 100 & 100 & {\smallpm{0.87}{0.38}} & {\smallpm{0.06}{0.05}} & {\smallpm{0.68}{0.74}}\\
\midrule 
\textbf{5px initialization} \\
MatchAnything
	 & 8 & 62 & 91 & 100 & 100 & 100 & 100 & 100 & {\smallpm{1.72}{0.84}} & {\smallpm{0.13}{0.09}} & {\smallpm{1.20}{1.72}}\\
ReFineR
	 & 52 & 92 & 96 & 98 & 100 & 100 & 100 & 100 & {\smallpm{0.95}{0.86}} & {\smallpm{0.07}{0.12}} & {\smallpm{0.75}{1.95}}\\
MatchAnything* + ReFineR
	 & 58 & 97 & 100 & 100 & 100 & 100 & 100 & 100 & {\smallpm{0.87}{0.38}} & {\smallpm{0.06}{0.05}} & {\smallpm{0.69}{0.66}}\\
\midrule 
\textbf{10px initialization} \\
MatchAnything
	 & 8 & 65 & 90 & 100 & 100 & 100 & 100 & 100 & {\smallpm{1.80}{0.83}} & {\smallpm{0.13}{0.11}} & {\smallpm{1.32}{1.62}}\\
ReFineR
	 & 41 & 73 & 78 & 79 & 96 & 100 & 100 & 100 & {\smallpm{1.10}{3.17}} & {\smallpm{0.10}{0.36}} & {\smallpm{1.02}{4.69}}\\
MatchAnything* + ReFineR
	 & 58 & 96 & 100 & 100 & 100 & 100 & 100 & 100 & {\smallpm{0.90}{0.40}} & {\smallpm{0.06}{0.05}} & {\smallpm{0.70}{0.62}}\\
\midrule 
\textbf{20px initialization} \\
MatchAnything
	 & 7 & 52 & 87 & 96 & 100 & 100 & 100 & 100 & {\smallpm{1.92}{1.16}} & {\smallpm{0.15}{0.11}} & {\smallpm{1.28}{1.82}}\\
ReFineR
	 & 20 & 35 & 39 & 41 & 48 & 87 & 100 & 100 & {\smallpm{10.74}{8.11}} & {\smallpm{0.79}{1.10}} & {\smallpm{10.21}{14.01}}\\
MatchAnything* + ReFineR
	 & 60 & 96 & 100 & 100 & 100 & 100 & 100 & 100 & {\smallpm{0.89}{0.39}} & {\smallpm{0.06}{0.05}} & {\smallpm{0.70}{0.67}}\\
\midrule 
\textbf{50px initialization} \\
MatchAnything
	 & 10 & 54 & 77 & 96 & 97 & 100 & 100 & 100 & {\smallpm{1.96}{2.17}} & {\smallpm{0.13}{0.15}} & {\smallpm{1.43}{1.85}}\\
ReFineR
	 & 1 & 1 & 1 & 1 & 1 & 2 & 60 & 100 & {\smallpm{47.85}{9.06}} & {\smallpm{4.90}{2.23}} & {\smallpm{41.26}{45.44}}\\
MatchAnything* + ReFineR
	 & 58 & 97 & 100 & 100 & 100 & 100 & 100 & 100 & {\smallpm{0.90}{0.40}} & {\smallpm{0.06}{0.05}} & {\smallpm{0.70}{0.65}}\\
\midrule 
\textbf{100px initialization} \\
MatchAnything
	 & 3 & 39 & 71 & 88 & 97 & 100 & 100 & 100 & {\smallpm{2.14}{2.37}} & {\smallpm{0.14}{0.25}} & {\smallpm{1.94}{2.47}}\\
ReFineR
	 & 0 & 0 & 0 & 0 & 0 & 0 & 0 & 36 & {\smallpm{107.20}{17.05}} & {\smallpm{10.17}{4.36}} & {\smallpm{84.29}{92.17}}\\
MatchAnything* + ReFineR
	 & 60 & 97 & 100 & 100 & 100 & 100 & 100 & 100 & {\smallpm{0.87}{0.41}} & {\smallpm{0.06}{0.05}} & {\smallpm{0.70}{0.70}}\\

    \bottomrule
        \end{tabular}} 
 \vspace{1pt}
        \caption{Registration at scale experiment, each camera pose was initialized at increasing pixel errors (\cref{app:exp_scale2}).}
        \label{tab:scale2}
    \end{table}

\subsection{Computation times}
\label{app:times}
The experiments were carried out on Ubuntu 24.04.2 LTS with an Intel Core Ultra 9 285K CPU and NVIDIA GeForce RTX 5090 GPU. We provide in \cref{tab:times} average computation times on CARLA with normals and ETH3D with RGB color. ReFineR on its own is on average as fast as Koide \etal's method on CARLA, but a little slower when adding MatchAnything* as a first stage. Two main factors impact ReFineR's runtime: first, the quality of the point cloud: a more precise point cloud will require fewer iterations to be registered; second, the number of points, which directly correlates with the Gaussian splatting forward and backward passes as shown in \cref{tab:times_set2}. However, this second point is mitigated by the tile-based Gaussian Splatting rendering CUDA engine, which parallelizes computations. By contrast, Koide \etal's method does not have such an implementation and scales more slowly with the number of points. Large computation times for GS-SLAM on CARLA are caused by its complete divergence, forcing the algorithm to run until the maximum number of iterations (arbitrarily fixed) is reached. On the contrary, on the ETH3D set with color registration, GS-SLAM is able to converge in roughly the same amount of time as ReFineR. From our experiments with ReFineR alone on the ETH3D set, our method converges on average after 200 iterations, while using GS-SLAM's photometric loss, the algorithm settled after 330 iterations. This result explains the time gap between GS-SLAM (10s / image) and ReFineR (7s / image), and tends to prove that GOM is also a more efficient supervision metric for this registration task than photometric loss. Adding MatchAnything as a first stage provides more accurate initialization, which consequently reduces the number of required iterations. For ETH3D, the algorithm looped only 80 times on average, therefore reducing the total amount of time compared to ReFineR alone.
\begin{table}[h] 
    \centering \scriptsize %
\begin{tabular}{@{}lcc}
\toprule
& CARLA (normals)
& ETH3D (intensity)
\\
\cmidrule(lr){2-2}
\cmidrule(lr){3-3}
&Time (s)
&Time (s) \\
\midrule 
\textbf{GS-SLAM} \cite{Matsuki2023GaussianSS} & 36.13 & 9.68 \\
\quad Render  & 11.31 & 3.48 \\
\quad Loss & 0.12 & 0.62 \\
\quad Backward & 22.24 & 4.62 \\

\textbf{Yuan \etal} \cite{yuanPixelLevelExtrinsicSelf2021} & 47.03 & N/A \\

\textbf{Koide \etal} \cite{koideGeneralSingleshotTargetless2023} & 10.89 & 26.80 \\

\textbf{Taylor} \etal \cite{Taylor2015MultiModalSC} & 80.01 & 87.79 \\

\textbf{MatchAnything} \cite{He2025MatchAnythingUC} & 4.84 & 4.90 \\
\quad Weight Loading & 3.12 & 3.11 \\
\quad Inference & 1.69 & 1.75 \\
\quad RANSAC & 0.02 & 0.03 \\

\textbf{ReFineR} & 11.62 & 6.83 \\
\quad Render & 3.69 & 2.89 \\
\quad Loss & 0.08 & 0.10 \\
\quad Backward & 7.31 & 3.14 \\
\textbf{MatchAnything* + ReFineR} & 13.81 & 6.97 \\
    \bottomrule
        \end{tabular}
    \vspace{5pt}
\caption{Detailed computation times for the benchmarking methods on the CARLA and ETH3D sets.}\label{tab:times}
    \end{table}
\begin{table}[h] 
    \centering \scriptsize{%
\begin{tabular}{@{}lcc}
\toprule
 & time / iter (s) & Num Points (M)
\\
\midrule 
ETH3D & {\smallpm{0.03}{0.02}} & {\smallpm{3.21}{2.82}} \\
Tanks and Temples & {\smallpm{0.14}{0.07}} & {\smallpm{19.92}{11.40}} \\
CARLA & {\smallpm{0.08}{0.05}} & {\smallpm{3.74}{2.72}} \\
Livox & {\smallpm{0.07}{0.02}} & {\smallpm{5.01}{1.68}} \\
nuScenes & {\smallpm{0.02}{0.00}} & {\smallpm{1.02}{0.35}} \\
RGBDScenes & {\smallpm{0.03}{0.01}} & {\smallpm{0.65}{0.17}}\\

    \bottomrule
        \end{tabular}} 
 \vspace{5pt}\caption{Average time per ReFineR iteration and average number of rendered points for each set.}\label{tab:times_set2}
    \end{table}

\subsection{Point downsampling}
In \cref{tab:subsample} we show a controlled robustness experiment on CARLA and ETH3D by randomly downsizing the point cloud (from 75\% to 5\%) to better characterize the density boundary of the method. We observe that starting from a 10-pixel perturbation, the final error monotonically decreases when the density increases, proving that ReFineR performs better with denser point clouds.
\begin{table}[h] 
    \centering \scriptsize \resizebox{\linewidth}{!}{%
\begin{tabular}{@{}lrrrrrrrrrrrrrr}
\toprule
& \multicolumn{7}{c}{ETH3D}
& \multicolumn{7}{c}{CARLA}
\\
\cmidrule(lr){2-8}
\cmidrule(lr){9-15}
& \multicolumn{5}{c}{pix@ (\%$\uparrow$)}  & $p_e$(px$\downarrow$) & time (s)
& \multicolumn{5}{c}{pix@ (\%$\uparrow$)}  & $p_e$(px$\downarrow$) & time (s)
\\
& 1 & 2 & 3 & 5 & 10 &  & 
& 1 & 2 & 3 & 5 & 10 &  &\\
\midrule 
ReFineR (s=100\%)
	 & 41 & 73 & 78 & 79 & 96 & {\smallpm{1.10}{3.17}} & {\smallpm{6.83}{5.03}} & 12 & 44 & 57 & 74 & 92 & {\smallpm{2.36}{3.45}} & {\smallpm{11.62}{7.98}}\\
ReFineR (s=75\%)
	 & 31 & 64 & 68 & 73 & 96 & {\smallpm{1.35}{3.26}} & {\smallpm{6.73}{4.15}} & 2 & 25 & 48 & 60 & 92 & {\smallpm{3.58}{3.36}} & {\smallpm{12.38}{10.85}}\\
ReFineR (s=50\%)
	 & 29 & 58 & 61 & 69 & 92 & {\smallpm{1.45}{3.52}} & {\smallpm{5.35}{3.09}} & 2 & 22 & 36 & 54 & 88 & {\smallpm{4.27}{4.01}} & {\smallpm{9.63}{8.06}}\\
ReFineR (s=25\%)
	 & 19 & 43 & 46 & 51 & 92 & {\smallpm{4.20}{3.71}} & {\smallpm{4.00}{2.17}} & 0 & 8 & 29 & 48 & 80 & {\smallpm{5.17}{3.92}} & {\smallpm{5.97}{4.05}}\\
ReFineR (s=10\%)
	 & 8 & 34 & 42 & 50 & 90 & {\smallpm{4.94}{3.75}} & {\smallpm{3.18}{1.57}} & 0 & 0 & 9 & 26 & 71 & {\smallpm{7.52}{3.66}} & {\smallpm{3.77}{1.98}}\\
ReFineR (s=5\%)
	 & 4 & 14 & 20 & 24 & 78 & {\smallpm{8.05}{3.60}} & {\smallpm{2.63}{1.52}} & 0 & 0 & 0 & 15 & 69 & {\smallpm{9.23}{3.06}} & {\smallpm{2.86}{1.56}}\\

    \bottomrule
        \end{tabular}} 
\vspace{3pt}
        
\caption{Downsampling experiment with raw point clouds. Each registration was initially perturbed by 10px and carried out with random downsampling of the point cloud by a varying amount.}
\label{tab:subsample}
    \end{table}

\subsection{Impact of pixel-level misalignment on 3D semantic label transfer}
\label{app:seg_transfer}
To assess the practical importance of fine image-LiDAR registration, we conducted an additional experiment measuring how semantic labels transferred from a single segmented image to a dense point cloud degrade as the camera pose is increasingly perturbed. Starting from the reference pose, we introduced controlled perturbations on 100 cameras of our CARLA dataset corresponding to reprojection errors ranging from 0 to 20 pixels, projected the point cloud into the image, and assigned semantic labels only to geometrically visible points using occlusion estimation from a splatting rendering. We then evaluated the consistency of the resulting 3D labels with the reference assignment using mIoU on 20 semantic classes.

\begin{figure}
    \centering
    \includegraphics[width=0.9\linewidth]{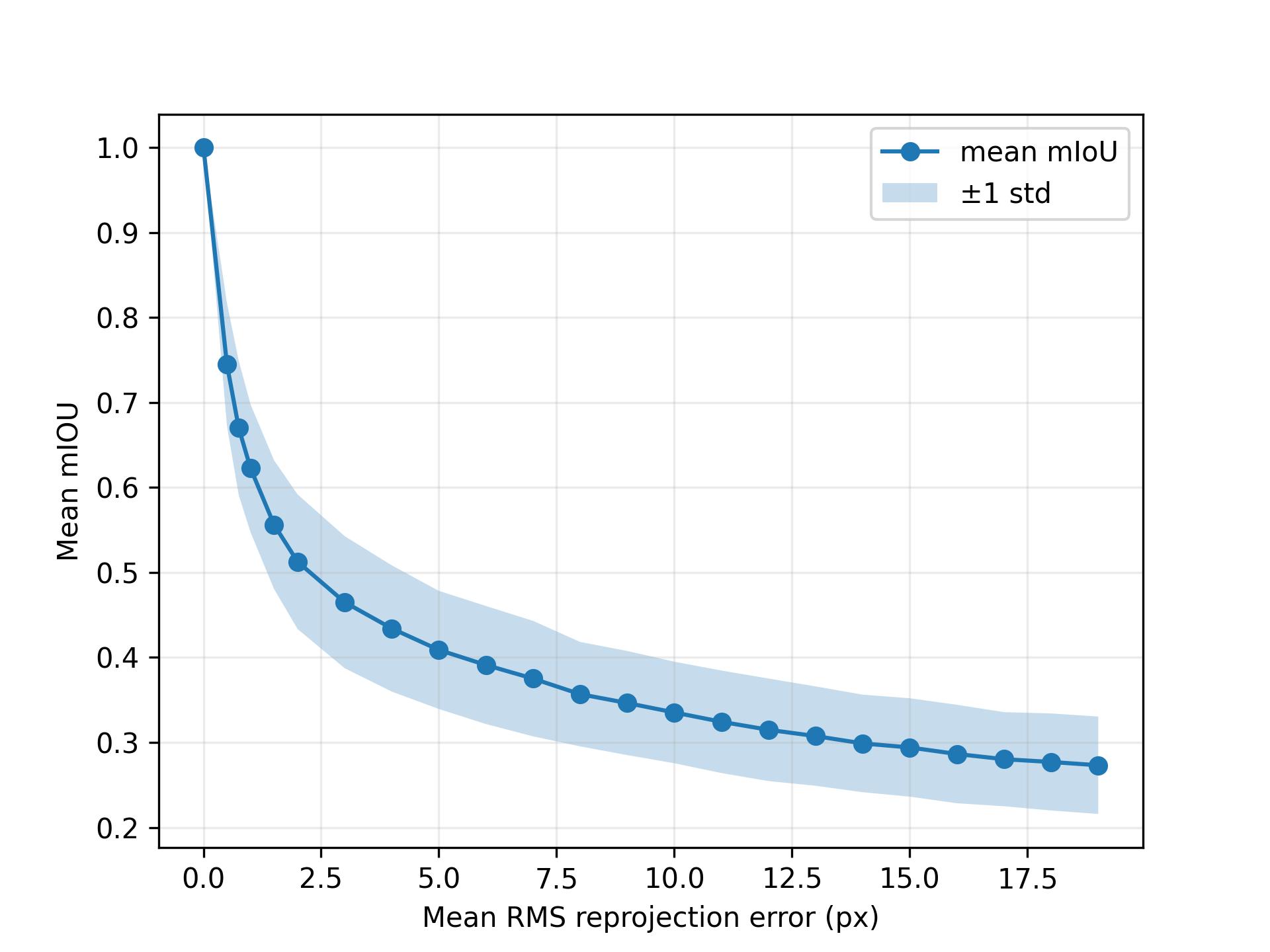}
    \caption{Average impact of camera pose error on projected 3D semantic labels.}
    \label{fig:seg}
\end{figure}

The results in \cref{fig:seg} show that semantic transfer is highly sensitive to small pose errors. In particular, the average mIoU drops sharply from 1.0 at 0 pixels to about 0.4 at 5 pixels, and to about 0.33 at 10 pixels. This strong degradation is mainly caused by thin structures and object boundaries (such as poles, barriers, lamp posts, and contour regions), which are the first to be incorrectly reprojected under small pose perturbations. In contrast, the degradation becomes much slower beyond 10 pixels, with mIoU decreasing only marginally from about 0.33 to 0.27 between 10 and 20 pixels. This observation supports the importance of pixel-level image-LiDAR registration.

\subsection{Qualitative registration experiment using UniCorrn}
\label{app:unicorrn}
\begin{figure}
    \centering
    
    \begin{subfigure}{0.3\textwidth}
        \includegraphics[width=\textwidth]{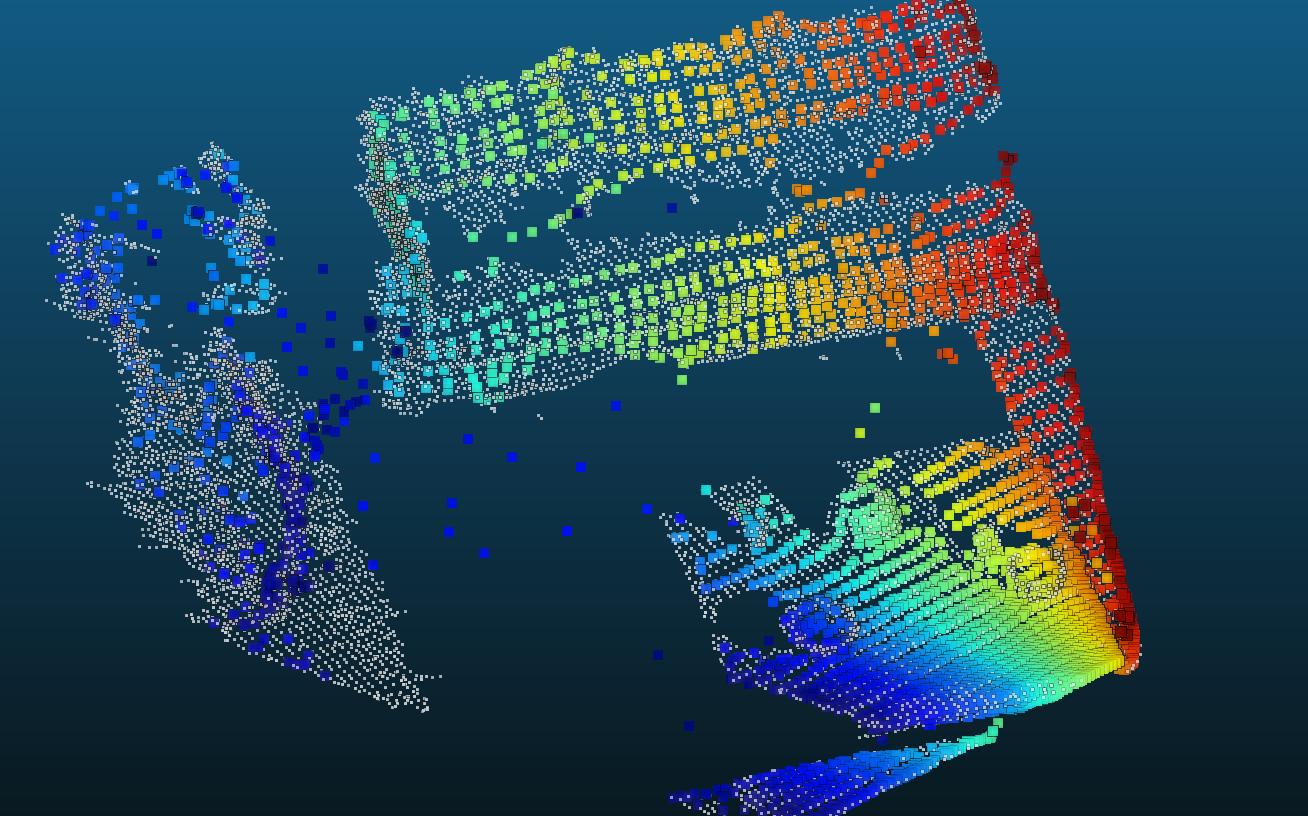}
        \label{fig:unicorrn_frame}
    \end{subfigure}
    \begin{subfigure}{0.3\textwidth}
        \includegraphics[width=\textwidth]{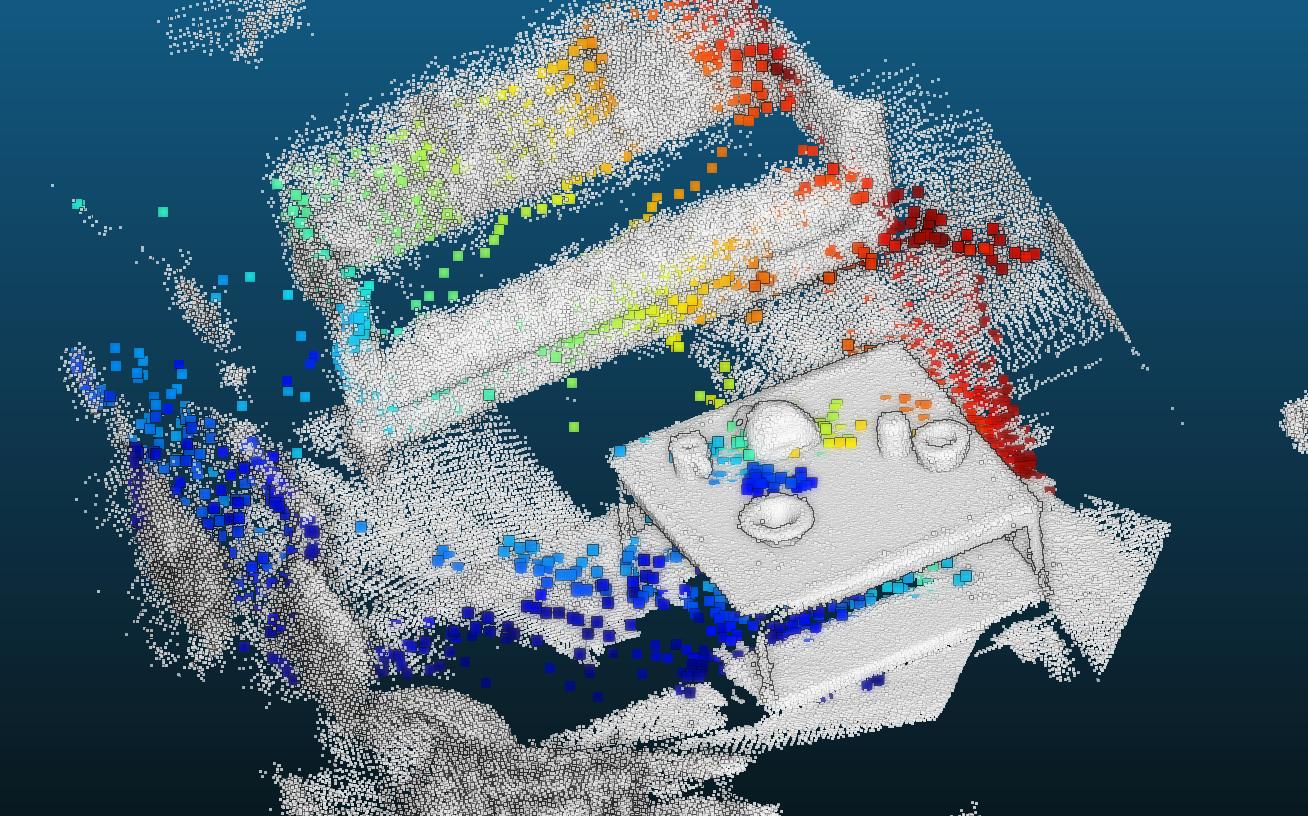}
        \label{fig:unicorrn_dense}
    \end{subfigure} 
    \begin{subfigure}{0.3\textwidth}
        \includegraphics[width=\textwidth]{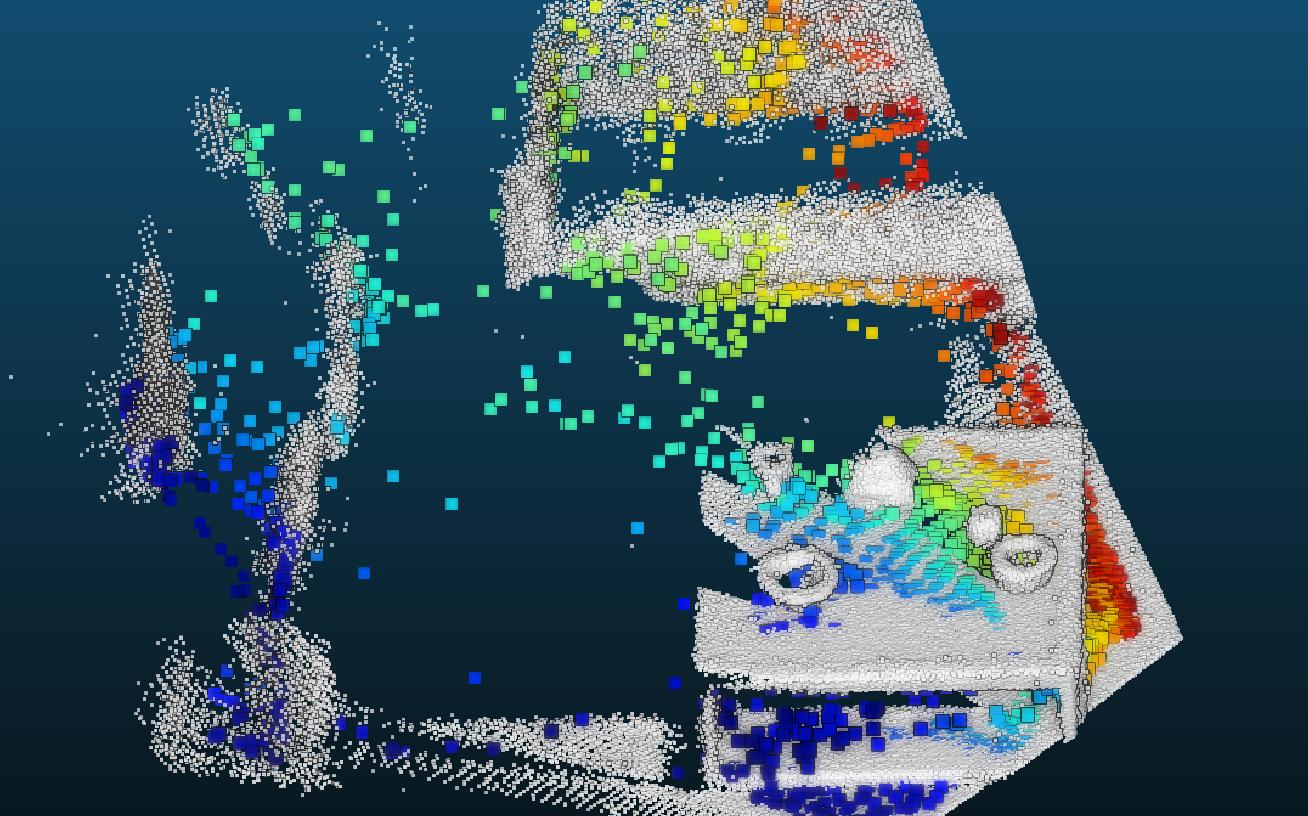}
        \label{fig:unicorrn_depth}
    \end{subfigure}
    \caption{Left to right: UniCorrn inference on a single RGBD frame (accurate predictions), on densely aggregated scans, and on synthetic depth from the dense point cloud (large errors). }
    \label{fig:unicorrn2}
\end{figure}

In preliminary experiments, we ran different inferences of the state-of-the-art image-to-point cloud registration model UniCorrn~\cite{goswami2026unicorrn}. This model trains several layers of self- and cross-attention, embedding 2D and 3D features in a common space in order to generate dimensionality-agnostic descriptors and correspondences. In the 2D-3D case, the model was trained using the 7scenes~\cite{glocker2013real-time} and RGB-D Scenes v2~\cite{Lai2014UnsupervisedFL} datasets, supplemented with synthetic depth maps from ScanNet++~\cite{Yeshwanth2023ScanNetAH} and ARKitScenes~\cite{dehghan2021arkitscenes}. From \cref{fig:unicorrn2}, we suspect that the training data specialized the model for indoor depth images, which have particular geometry and density. Running the same model on general LiDAR point clouds may then represent a distribution shift for the model, which explains the weak results. We believe that this claim should be studied in more depth as it may represent a limitation for other similar methods~\cite{wangFreeRegImagetoPointCloud2023, li2D3DMATR2D3DMatching2023}.

\end{document}